\documentclass[10pt,twocolumn,letterpaper]{article}

\usepackage{iccv}
\usepackage{times}
\usepackage{epsfig}
\usepackage{graphicx}
\usepackage{amsmath}
\usepackage{amssymb}
\usepackage{subfigure}
\usepackage{url}
\usepackage{bm}
\usepackage{makecell}
\usepackage[accsupp]{axessibility}  

\usepackage[breaklinks=true,bookmarks=false]{hyperref}

\iccvfinalcopy 

\def\iccvPaperID{5} 

\begin{document}

\title{LLVIP: A Visible-infrared Paired Dataset for Low-light Vision}

\author{Xinyu Jia, Chuang Zhu\thanks{the corresponding author: Chuang Zhu (czhu@bupt.edu.cn)}, Minzhen Li, Wenqi Tang, Shengjie Liu, Wenli Zhou \\
Beijing University of Posts and Telecommunications Beijing 100876, China \\
{\tt\small \{jiaxinyubupt, czhu, 2020140111lmz, tangwenqi, shengjie.Liu, zwl\}@bupt.edu.cn}\\
\textcolor{red}{\url{https://bupt-ai-cz.github.io/LLVIP}}
}

\maketitle

\begin{abstract}
It is very challenging for various visual tasks such as image fusion, pedestrian detection and image-to-image translation in low light conditions due to the loss of effective target areas. In this case, infrared and visible images can be used together to provide both rich detail information and effective target areas. In this paper, we present LLVIP, a visible-infrared paired dataset for low-light vision. This dataset contains 30976 images, or 15488 pairs, most of which were taken at very dark scenes, and all of the images are strictly aligned in time and space. Pedestrians in the dataset are labeled. We compare the dataset with other visible-infrared datasets and evaluate the performance of some popular visual algorithms including image fusion, pedestrian detection and image-to-image translation on the dataset. The experimental results demonstrate the complementary effect of fusion on image information, and find the deficiency of existing algorithms of the three visual tasks in very low-light conditions. We believe the LLVIP dataset will contribute to the community of computer vision by promoting image fusion, pedestrian detection and image-to-image translation in very low-light applications. The dataset is being released in \url{https://bupt-ai-cz.github.io/LLVIP/}. Raw data is also provided for further research such as image registration.
\end{abstract}


\begin{figure*}[ht]
\begin{center}
\includegraphics[width=0.90\linewidth]{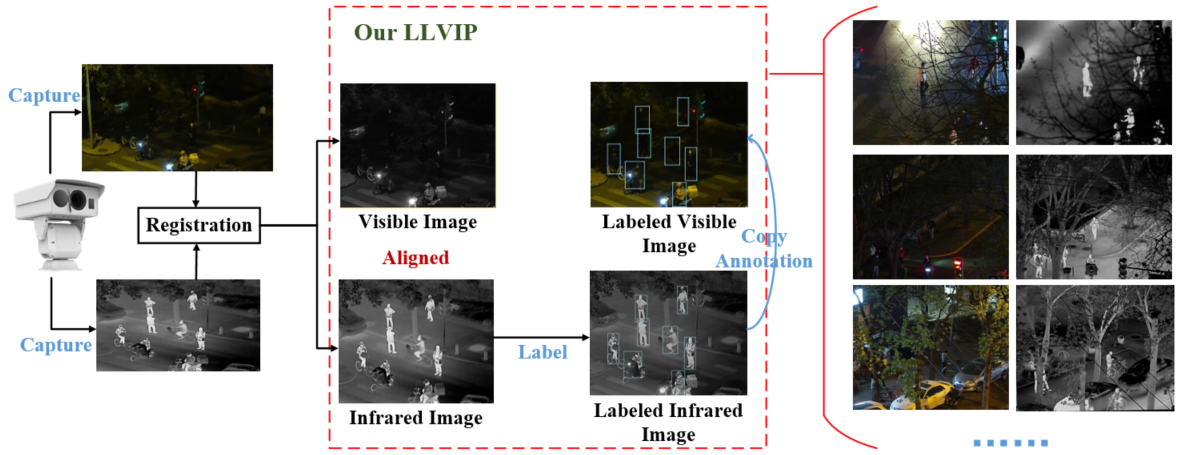}
\end{center}
\caption{The establishment of the LLVIP dataset. We first collect images with a binocular camera, and then perform registration to align the infrared and visible images. Finally we annotate on the infrared images.} 
\label{fig-LLVIP} 
\end{figure*}

\section{Introduction}

It is very challenging for various visual tasks on visible image with limited quality, for example, in low light conditions due to the loss of effective target areas. Infrared images, which are not limited by light conditions, can play a role of supplementary information. Visible images contain a great deal of texture information and details, but it is difficult to distinguish objects under low-light condition. Infrared images are imaged through the temperature field of the object surface, so they can highlight targets such as pedestrians, but the texture information is missing. Visible and infrared image fusion can generate a single complementary image that has both rich detail information and effective target areas. Then the fused image can be applied to human visual perception, object detection and video surveillance.

The target of image fusion is to extract salient features from source images and integrate them into a single image by appropriate fusion method. Image fusion task has been developed with many different methods. Deep learning algorithms~\cite{ma2019fusiongan, li2018densefuse, zhang2020ifcnn} have achieved great success in the image fusion field. Data is an essential part of building an accurate deep learning system, so visible and infrared paired datasets are required. TNO~\cite{toet2014tno}, KAIST Multispectral Dataset~\cite{hwang2015multispectral}, OTCBVS OSU Color-Thermal Database~\cite{davis2007otcbvs}, etc. are all very practical datasets. However, they are not simultaneously aimed at image fusion and low-light pedestrian detection, that is, they cannot simultaneously satisfy the conditions of large scale, image alignment, low-light scene and a lot of pedestrians. Therefore, it is necessary to propose a visible-infrared paired dataset containing many pedestrians under low-light condition.

We build LLVIP, a visible-infrared paired dataset for low-light vision. We collect images with a binocular camera which consists of a visible light camera and an infrared camera. Such a binocular camera can ensure the consistency of image pairs in time and space. Each pair of images are registered and cropped so that they have the same field of view and size. Images are strictly aligned in time and space, which makes the dataset useful in image fusion and image-to-image translation. Different fusion algorithms are evaluated on our LLVIP dataset, and we analyze the results subjectively and objectively. We evaluate the fusion algorithms in many aspects and find that LLVIP is challenging to the existing fusion methods. Fusion algorithms cannot capture details in low-light visible images. We also evaluate the typical image-to-image translation algorithm on the dataset, and it performs very poorly.

The dataset contains a large number of different pedestrians under low-light condition, which makes it useful for low-light pedestrian detection. One of the difficulties in this detection task is image labeling, because human eyes can hardly distinguish pedestrians, let alone mark the bounding boxes accurately. We propose a method to label low-light visible images by aligned infrared images reverse mapping and labeled all the images in the dataset. The low-light pedestrian detection experiment is also carried out on our dataset, which demonstrates that there is still a lot of room for improvement in the performance of the task.

The main contributions of this paper are as follows:
1) We propose LLVIP, the first visible-infrared paired dataset for various low-light visual tasks.
2) We propose a method to label low-light visible images by aligned infrared images, and label pedestrians in LLVIP.
3) We evaluate the experimental results of image fusion, pedestrian detection and image-to-image translation on LLVIP, and find that the dataset is a huge challenge for all the tasks.


\section{Related Datasets}

There are now datasets for visible and infrared pairs images for a variety of visual tasks, such as TNO Image Fusion Dataset~\cite{toet2014tno}, INO Videos Analytics Dataset and OTCBVS OSU Color-Thermal Database~\cite{davis2007otcbvs}, CVC-14~\cite{gonzalez2016pedestrian}, KAIST Multispectral Dataset~\cite{hwang2015multispectral} and FLIR Thermal Dataset.

\begin{figure}[htb]
\begin{center}
\subfigure[TNO: Kaptein 1123]{
    \includegraphics[height=0.17\linewidth]{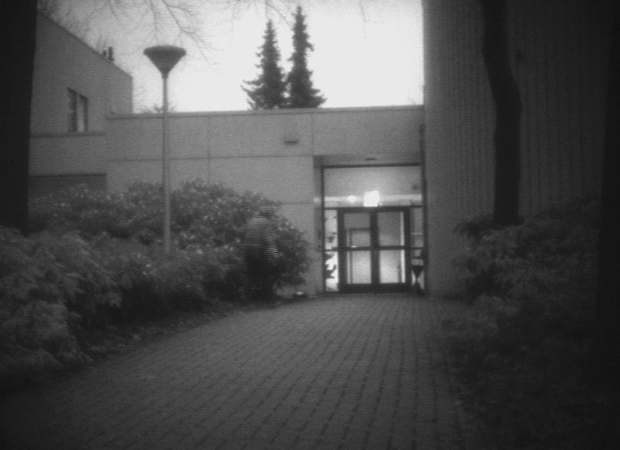}
    \includegraphics[height=0.17\linewidth]{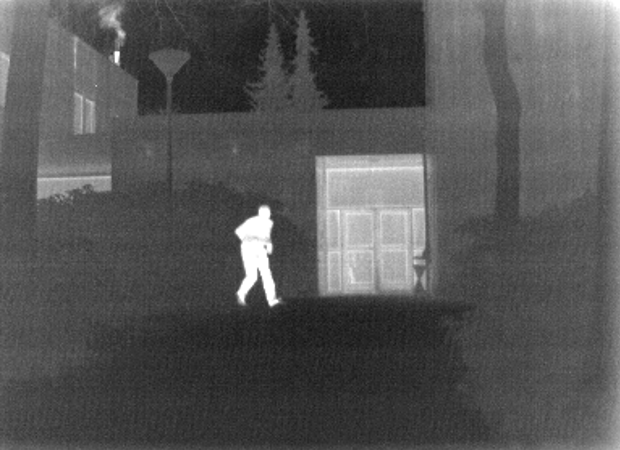}
}
\subfigure[TNO: Nato camp 1819]{
    \includegraphics[height=0.17\linewidth]{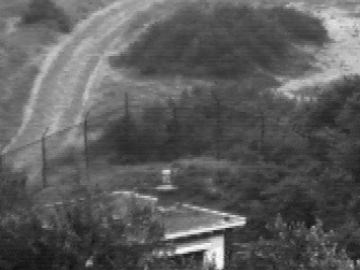}
    \includegraphics[height=0.17\linewidth]{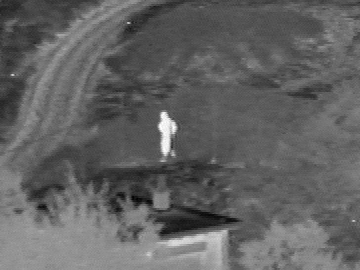}
}
\subfigure[OSU : Sequence 1]{
    \includegraphics[height=0.17\linewidth]{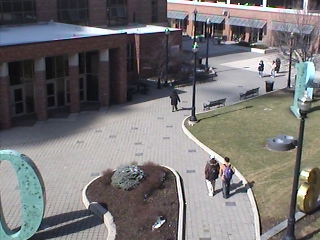}
    \includegraphics[height=0.17\linewidth]{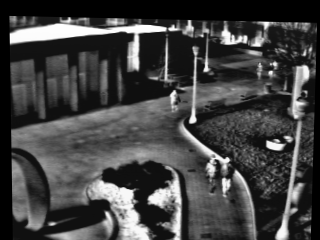}
}
\subfigure[OSU: Sequence 4]{
    \includegraphics[height=0.17\linewidth]{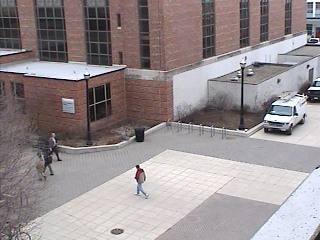}
    \includegraphics[height=0.17\linewidth]{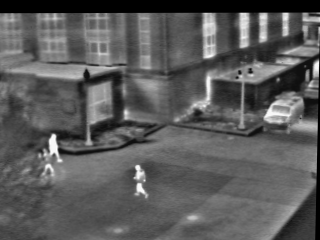}
}
\end{center}
\caption{Some pairs of images in TNO dataset (a)(b) and OTCBVS OSU Color-Thermal Database (c)(d). In each pair of images, the image on the left is visible and the image on the right is infrared.} 
\label{fig-TNO} 
\end{figure}

The TNO Image Fusion Dataset~\cite{toet2014tno} posted on 2014 by Alexander Toet is the most commonly used public dataset for visible and infrared image fusion. TNO contains multi-spectral (enhanced visual, near-infrared, and long-wave infrared or thermal) nighttime imagery of different military scenes, and is recorded in different multi-band camera systems. Fig.~\ref{fig-TNO}(a)(b) shows two pairs of images commonly used in TNO.

TNO plays a huge role in image fusion research. However, it is not suitable for image fusion algorithms based on deep learning, for the following reasons: 1) TNO contains only 261 pairs of images, including many sequences of consecutive similar images. 2) TNO contains few objects such as pedestrian, so it is difficult to be used for object detection after fusion.

INO Videos Analytics Dataset is provided by the National Optics Institute of Canada, and contains several pairs of visible and infrared videos representing different scenarios captured under different weather conditions. Over the years, INO has developed a strong expertise in using multiple sensor types for video analytics applications in uncontrolled environment. INO Videos Analytics Dataset contains very rich scenes and environments, but few pedestrian and few low-light images.

The OTCBVS Benchmark Dataset~\cite{davis2007otcbvs} Collection initiated by Dr. Riad I. Hammoud in 2004 contains very rich infrared datasets, the OSU Color-Thermal Database~\cite{davis2007background} is a visible-infrared paired dataset for fusion of color and thermal imagery and fusion-based object detection. The images were taken at a busy pathway intersection on the Ohio State University Campus, cameras mounted to each other on tripod at two locations approximately 3 stories above ground. The images contain a large number of pedestrians. However, all images are collected in the daytime, 
so the pedestrians in visible images are already very clear. In such cases, the advantages of infrared images are not prominent. Some pairs of images are shown in Fig.~\ref{fig-TNO}(c)(d).

\begin{figure}[htb]
\begin{center}
\subfigure[2014\_05\_04\_23\_14\_48\_447000]{
    \includegraphics[height=0.17\linewidth]{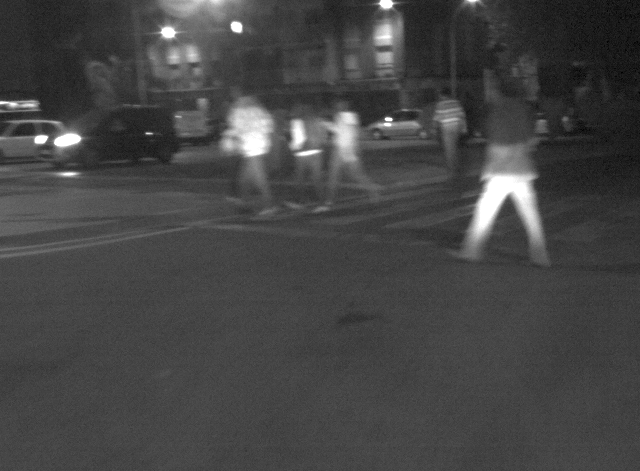}
    \includegraphics[height=0.17\linewidth]{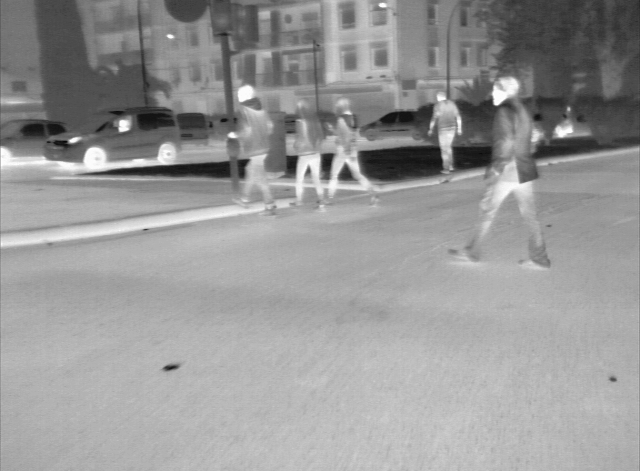}
}
\subfigure[2014\_05\_04\_23\_19\_42\_899000]{
    \includegraphics[height=0.17\linewidth]{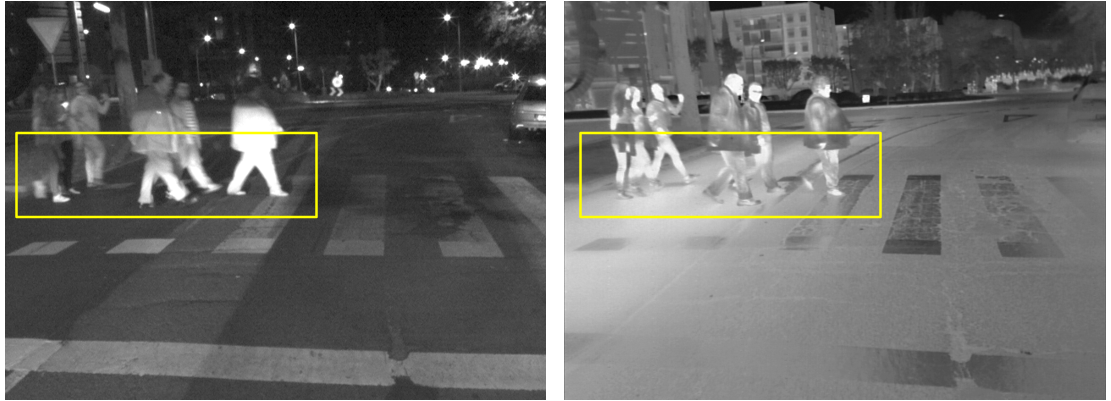}
}
\end{center}
\caption{Some pairs of images in CVC-14 dataset. The name of the image pair is its name in CVC-14 dataset. The images are all from the driving perspective, and the visible light image and infrared image are not strictly corresponding in time (b).} 
\label{fig-CVC14} 
\end{figure}

CVC-14~\cite{gonzalez2016pedestrian} is a visible and infrared images dataset aiming at automatic pedestrian detection task. CVC-14 dataset contains four sequences: day/FIR, night/FIR, day/visible and night/visible. It is used to study automatic driving, so images are not suitable for video surveillance, as shown in Fig.~\ref{fig-CVC14}. Moreover, the images in CVC-14 are not dark enough, and the human eye can easily identify the objects. Note that CVC-14 can not be used for image fusion task because the visible and infrared images are not strictly aligned in time, as shown in the yellow box of Fig.~\ref{fig-CVC14}(b).

KAIST Multispectral Dataset~\cite{hwang2015multispectral} provides well aligned color-thermal image pairs, captured by beam splitter-based special hardware. With this hardware, they captured various regular traffic scenes at day and night time to consider changes in light conditions. KAIST Multispectral Dataset is also a data set for autonomous driving.

The FLIR starter thermal dataset enables developers to start training convolutional neural networks (CNN), empowering the automotive community to create the next generation of safer and more efficient ADAS and driverless vehicle systems using cost-effective thermal cameras from FLIR. However, the visible and infrared images in the dataset are not registered, so they cannot be used for image fusion.


\section{The LLVIP Dataset}

We propose LLVIP, a visible-infrared paired dataset for low-light Vision. In this section, we will talk about how we collect, select, register and annotate images, and then analyze the advantages, disadvantages and application scenarios of the dataset.

\vspace{-1.2em}
\paragraph{Image capture.} The camera equipment we use is HIKVISION DS-2TD8166BJZFY-75H2F/V2, a binocular camera platform that consist of a visible light camera and a infrared camera. The working wavelength for the thermal infrared camera is 8$\sim$14um. We capture images containing many pedestrians and cyclists from different locations on the street between 6 and 10 o'clock in the evening.

After time alignment and manual filtering, time-synchronized and high-quality image pairs containing pedestrians are selected. So far, we have collected 15488 pairs of visible-infrared images from 26 different locations. Each of the 15488 pairs of images contains pedestrians.

\begin{figure}[htb]
\begin{center}
\subfigure[dual-spectrum camera]{
    \includegraphics[height=0.28\linewidth]{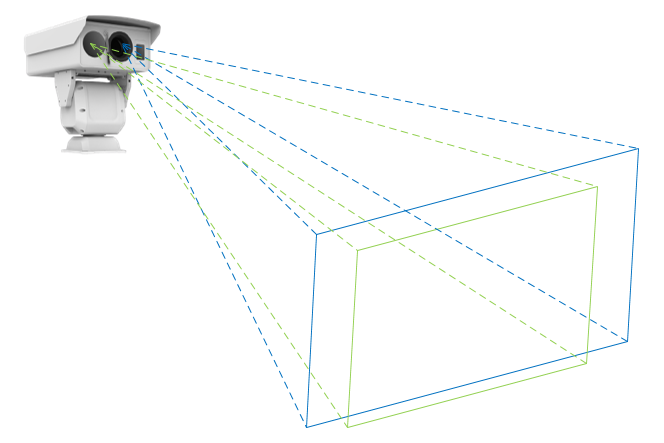}
}
\subfigure[different field of views]{
    \includegraphics[height=0.28\linewidth]{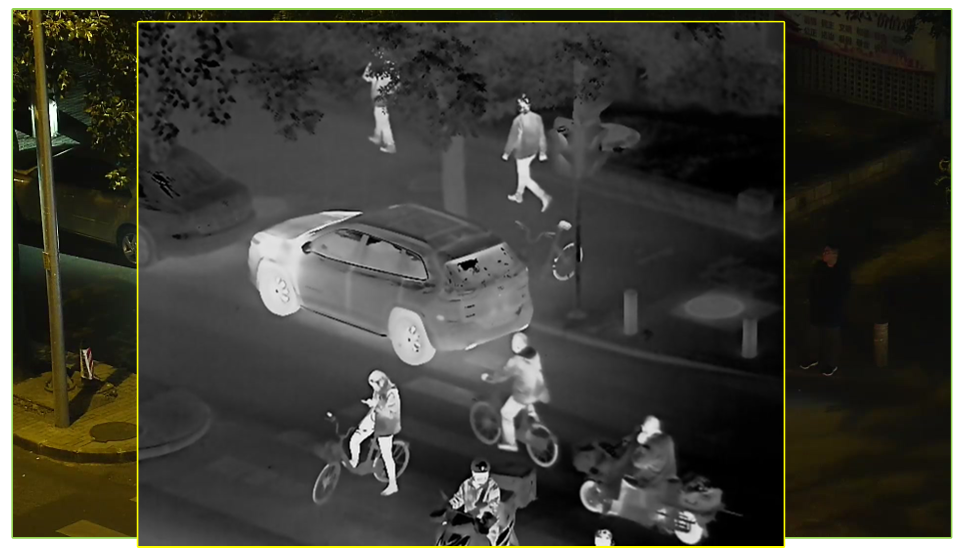}
}
\subfigure[images after registration]{
    \includegraphics[height=0.28\linewidth]{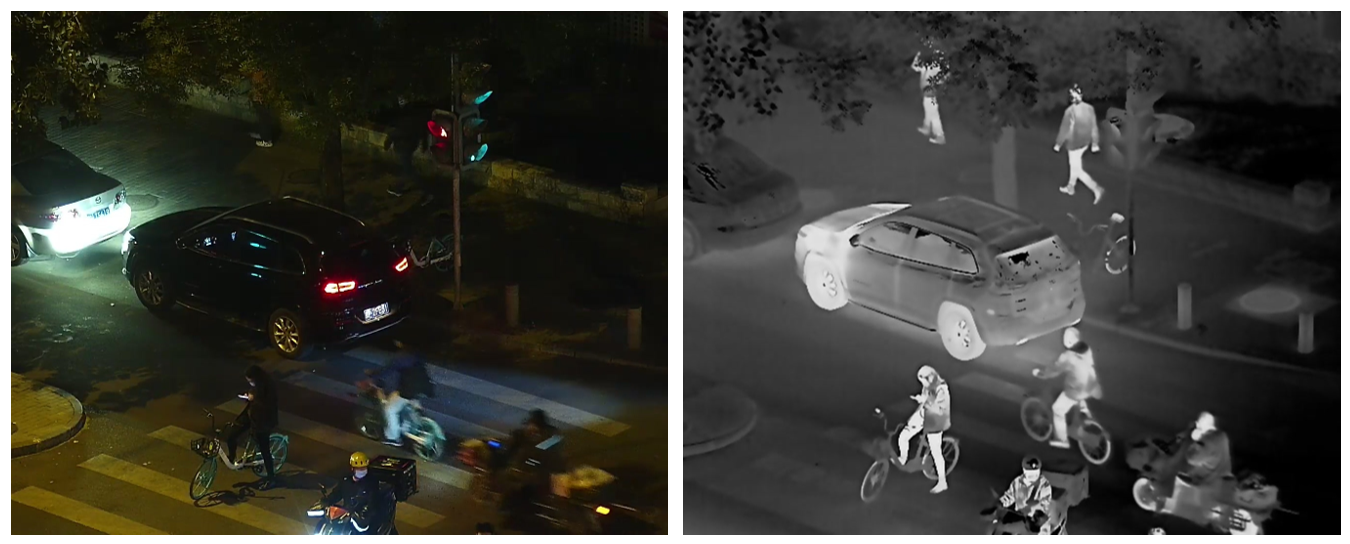}
}
\end{center}
\caption{Image capture and image registration. Dual-spectrum camera captures images of different field of views (a)(b), and the image pairs are aligned after registration (c).} 
\label{fig-VID} 
\end{figure}
\vspace{-1.2em}
\paragraph{Registration.} Although visible light images and infrared images are shot by a binocular camera, they are not aligned due to the different field sizes of different sensor cameras. We clipped and registered visible-infrared image pairs so that they have exactly the same field of vision and the same image size. For this multi-modal image registration task, it is difficult to just apply automatic detection registration methods, so we chose a semi-manual method. We first manually select several pairs of points that need to be aligned between the two images, then calculate the projection transformation to deform the infrared image, and finally cut out to get the registered image pairs. Fig.~\ref{fig-VID}(b)(c) shows the comparison of visible-infrared images before and after registration. We also provide unregistered image pairs for researchers to study visible and infrared image registration.
\vspace{-1.2em}
\paragraph{Annotations.} One of the difficulties in low-light pedestrian detection is image labeling, because human eyes can hardly distinguish human bodies and mark the bounding boxes accurately in images. We propose a method to label low-light visible images by using of infrared images. Firstly, we label pedestrians on infrared images where pedestrians are obvious. Then because the visible image and the infrared image are aligned, the annotations can be copied directly to the visible image. We labeled all the image pairs of our dataset in this way.

\begin{table*}[htb]
\begin{center}
\resizebox{0.98\linewidth}{!}{
\begin{tabular}{ccccccc}
\hline
       & \makecell[c]{Number of image pairs\\(1 frame selected per second)} & Resolution & Aligned & Camera angle & Low-light & Pedestrian \\ \hline
TNO    & 261             & 768 $\times$ 576           & $\surd$      & shot on the ground & few          & few          \\ 
INO    & 2100            & 328 $\times$ 254           & $\surd$      & surveillance       & $\surd$      & few          \\ 
OSU    & 285             & 320 $\times$ 240           & $\surd$      & surveillance       & $\times$            & $\surd$      \\ 
CVC-14 & 8490             & 640 $\times$ 512           & $\times$            & driving            & $\surd$      & $\surd$      \\ 
KAIST  & 4750            & 640 $\times$ 480           & $\surd$      & driving            & $\surd$      & $\surd$      \\ 
FILR   & 5258            & 640 $\times$ 512           & $\times$            & driving            & $\surd$      & $\surd$      \\ 
LLVIP & \textbf{15488}  & \textbf{1080 $\times$ 720} & \bm{$\surd$} & surveillance       & \bm{$\surd$} & \bm{$\surd$} \\ \hline
\end{tabular}}
\end{center}
\caption{Comparison of LLVIP and existing datasets including TNO, INO Videos Analytics Dataset, OSU Color-Thermal Database, CVC-14, KAIST Multispectral Dataset and FLIR Thermal Dataset. Resolution refers to the average when it is different in a dataset. } 
\label{table-datasets}
\end{table*}
\vspace{-1.2em}
\paragraph{Advantages.} Table~\ref{table-datasets} shows comparison of LLVIP and existing datasets mentioned in Section 2. Our LLVIP dataset has the following advantages: 
\begin{itemize}
    \item Visible-infrared images are synchronous in time and space. Thus the image pair can be used for image fusion and supervised image-to-image translation.
    \item The dataset is under low-light conditions. Infrared images bring abundant supplementary information to low-light visible images. Therefore, the dataset is suitable for the study of image fusion and can be used for low-light pedestrian detection.
    \item The dataset contain a large number of pedestrian with annotations. Visible and infrared image fusion has more obvious effect and significance in pedestrian detection.
    \item The quality of the images is very high. The resolution of the original visible images is 1920 $\times$ 1080 and that of the infrared images is 1280 $\times$ 720. The dataset is a high quality visible-infrared paired dataset compared to others.
\end{itemize}
\vspace{-1.2em}
\paragraph{Disadvantages.} Most of the images in the dataset are collected from a medium distance, and the pedestrians in the images are of a medium size. Therefore, this dataset is not suitable for the study of long-distance small-target pedestrian detection.
\vspace{-1.2em}
\paragraph{Applications.} LLVIP dataset can be used to study the following visual task: 1) Visible and infrared image fusion. 2) Low-light pedestrian detection. 3)Visible-to-infrared image-to-image translation. 4) Others, such as multimodel image registration. 

\section{Tasks}

\begin{figure*}[ht]
\begin{center}
\includegraphics[width=0.96\linewidth]{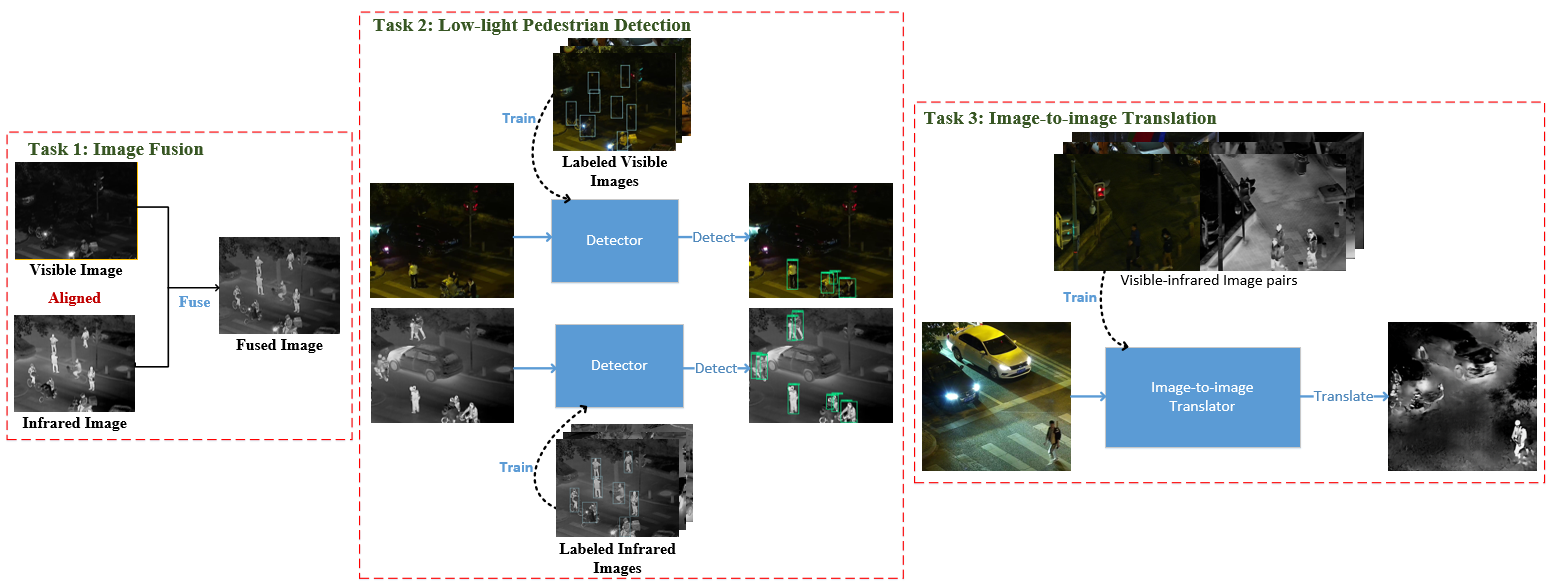}
\end{center}
\caption{Three low-light vision tasks our LLVIP dataset can apply to. Task 1: Image fusion. Task 2: Low-light pedestrian detection. Task 3: Image-to-image translation.} 
\label{fig-tasks} 
\end{figure*}

In this section, we will detail the visual tasks to which the dataset can be applied as mentioned in Section 3. As shown in Fig.~\ref{fig-tasks}, they are image fusion, low-light pedestrian detection and image-to-image translation.

\subsection{Image Fusion and Metrics}

Image Fusion attempts to extract salient features from source images, then these features are integrated into a single image by appropriate fusion method. The fusion of visible and infrared images can obtain both the rich details of visible images and the prominence of heat source targets in infrared images.

\subsubsection{Fusion methods}

In recent years, many fusion methods have been proposed. In contrast to traditional manual methods, we focus on deep learning methods including convolutional neural network and generative adversarial network. Deep learning methods have achieved the best performance of existing methods.

Hui Li and Xiao-Jun Wu proposed DenseFuse~\cite{li2018densefuse}, which incorporated the dense block in the encoder, that is, the outputs of each convoluted layer were connected to each other. In this way, the network can get more features from the source images during the encoding process. Besides, DenseFuse also designed two different fusion strategies, addition and $l_1$-norm.

Jiayi Ma \etal proposed FusionGAN~\cite{ma2019fusiongan}, which is a method to fuse visible and infrared images using a generative adversarial network. The generator makes the fused image contain the pixel intensity of infrared image and the gradient information of visible image. The discriminator is designed to distinguish the fused image from the visible image after extracting the feature, so that the fused image can contain more texture information of the visible image.

\subsubsection{Fusion metrics}

Many fusion metrics have been proposed, but it is hard to say which one is better, so it is necessary to select multiple metrics to evaluate the fusion methods. We objectively evaluate the performances of different fusion methods using entropy (EN), mutual information (MI)~\cite{qu2002information,ramesh2002fusion} series, structural similarity (SSIM)~\cite{wang2004image}, $Q_{abf}$~\cite{piella2003new} and visual information fidelity for fusion (VIFF)~\cite{han2013new}. Detailed definitions and calculation formulas are provided in the supplementary materials.

EN is defined based on information theory, which measures the amount of information the fused image contains. MI~\cite{qu2002information} is the most commonly used objective metric for image fusion. Fusion factor (FF) ~\cite{ramesh2002fusion} is concepts based on MI. Normalize mutual information $Q_MI$ is defined based on entropy and mutual information. SSIM~\cite{wang2004image} is a perceptual metric that quantifies image quality degradation caused by processing such as data compression or by losses in data transmission. $Q_{abf}$~\cite{piella2003new} is a quality index which gives an indication of how much of the salient information contained in each of the input images has been transferred into the fused image without introducing distortions. VIFF~\cite{han2013new} utilizes the models in VIF to capture visual information from the two source fused pairs.

\subsection{Low-light Pedestrian Detection}

Pedestrian detection has made great progress over the past few years due to its multiple applications in automatic drive, video surveillance and people counting. The performance of pedestrian detection methods remains limited in poor light conditions, and there are few methods and datasets for low light conditions. One reason for the lack of low-light visible pedestrian datasets is that it is difficult to label them accurately. We annotate low-light visible images by labeling aligned infrared images, which overcomes this difficulty.

The Yolo~\cite{redmon2016you, redmon2017yolo9000, redmon2018yolov3, bochkovskiy2020yolov4, glenn_jocher_2020_3983579} series are the most commonly used one stage algorithms for object detection. As computer vision technology evolves, the series continues to incorporate new technologies and updates. In Section 5.2, we select Yolov3~\cite{redmon2018yolov3} and Yolov5~\cite{glenn_jocher_2020_3983579} for pedestrian detection experiments on our LLVIP dataset, and the experimental results demonstrate that the existing pedestrian detection algorithms do not perform well in low light conditions. 

\subsection{Image-to-image Translation}

Image-to-image translation is a technique that converts images from one domain to another. It has made great progress with the development of conditional generative adversarial networks (cGANs)~\cite{mirza2014conditional}. And it has been used in many scenarios such as transformation of semantic label map and photo~\cite{isola2017image}, black-and-white picture and color picture, sketch and photo, daytime picture and nighttime picture, etc. Compared with visible images, infrared images are difficult to capture due to the expensive facility and strict shooting conditions. To overcome these restrictions, image-to-image translation methods are used to construct infrared data from easily obtained visible images.

Existing visible-to-infrared translation methods can be mainly divided into two categories, one is the use of physical model and manual image conversion relation design, the other is deep learning method. The situation of thermal imaging is complicated, so it is difficult to manually summarize all the mapping relation between optical images and infrared images. Therefore, the results of physical model methods are often inaccurate and lacking in detail. In recent years, deep learning research has developed rapidly, as for image-to-image translation, it mainly focuses on generative adversarial networks (GANs)~\cite{goodfellow2014generative}. Pix2pix GAN was a general-purpose solution to image-to-image translation problems, which made it possible to apply the same generic approach to problems that traditionally would require very different loss formulations~\cite{isola2017image}.


\section{Experiments}

In this section, we describe in detail the experiments of image fusion, pedestrian detection, and image-to-image translation on our LLVIP dataset, and evaluate the results. The experiments are conducted on NVIDIA Tesla T4 GPU, 16GB.

\subsection{Image Fusion}

The fusion algorithms selected by us include gradient transfer fusion (GTF)~\cite{ma2016infrared}, FusionGAN~\cite{ma2019fusiongan}, Densefuse (addition fusion strategy and $l_1$ fusion strategy)~\cite{li2018densefuse} and IFCNN~\cite{zhang2020ifcnn}. We use the original models and parameters of these algorithms. Then, we evaluate these fusion results subjectively and objectively. Finally, we illustrate the significance of our dataset for the study of image fusion algorithm from the fusion experimental results. All hyperparameters and settings are as given by the author in the papers. The GTF experiments are conducted on Intel Core i7-4720HQ CPU.

\begin{figure*}[htb]
  \begin{center}
  \begin{minipage}[b]{0.99\linewidth}
  \subfigure[visible]{
    \begin{minipage}[b]{0.127\linewidth}
      \centering
      \includegraphics[width=\linewidth]{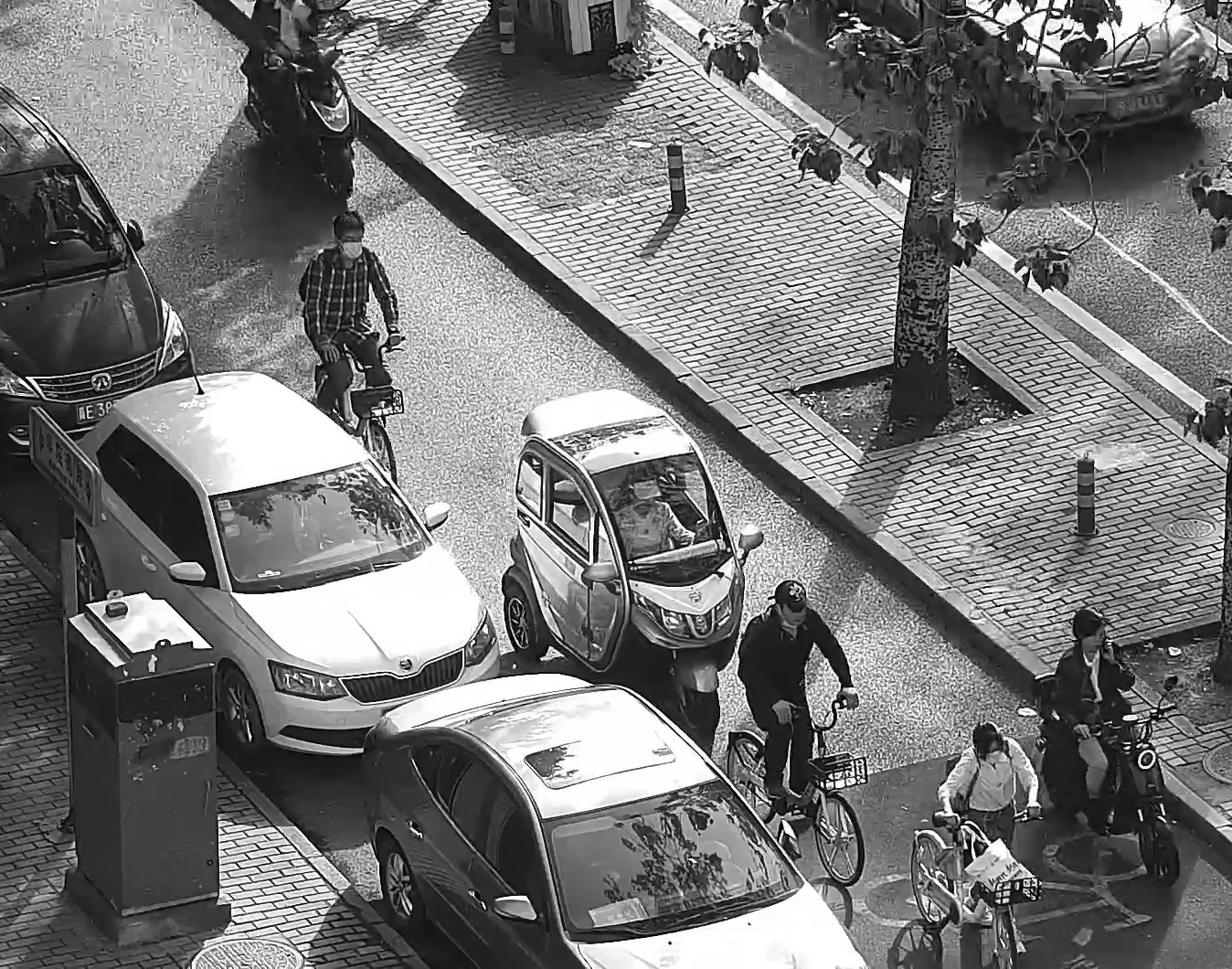}\vspace{2pt}
      \includegraphics[width=\linewidth]{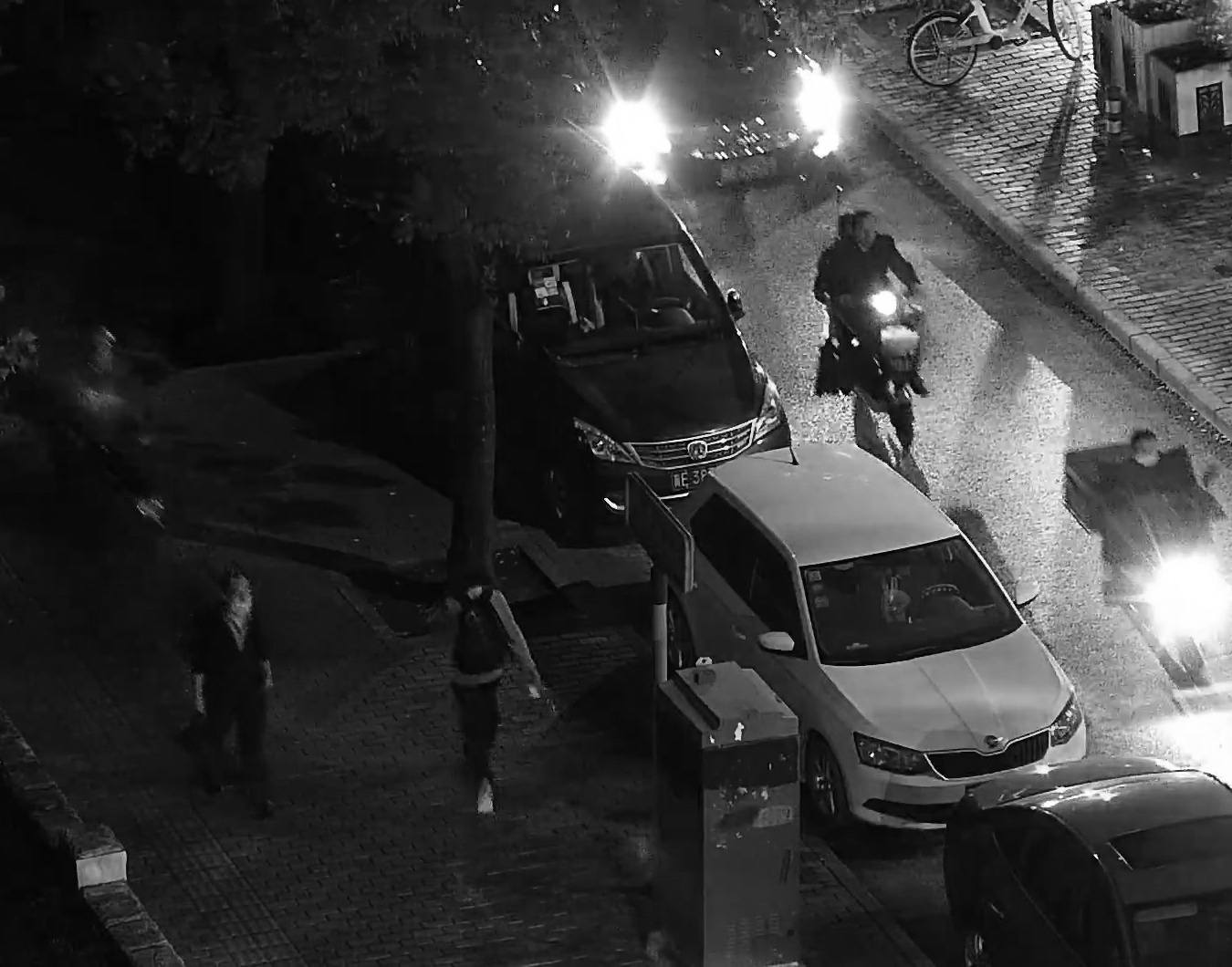}\vspace{2pt}
      \includegraphics[width=\linewidth]{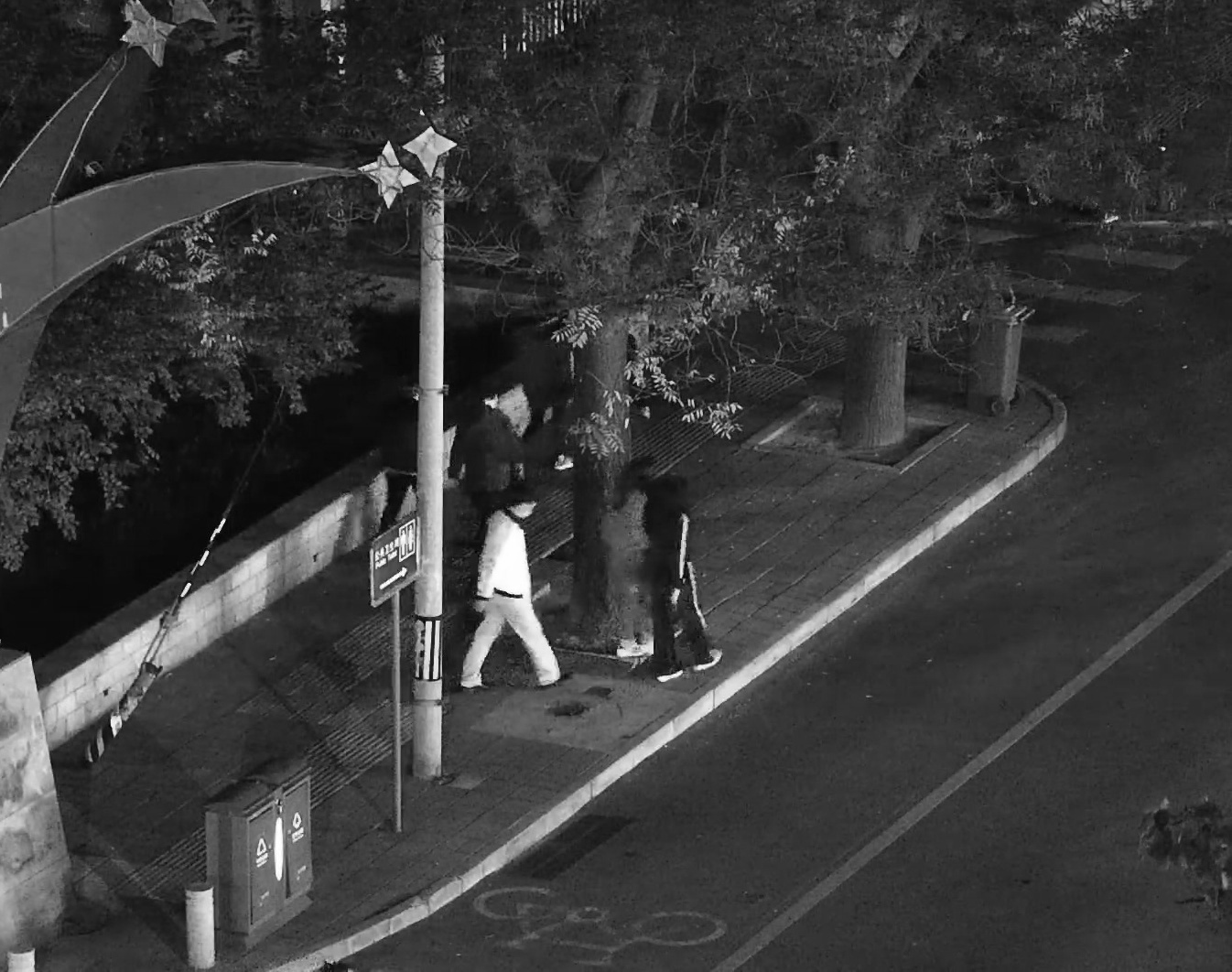}\vspace{2pt}
      \includegraphics[width=\linewidth]{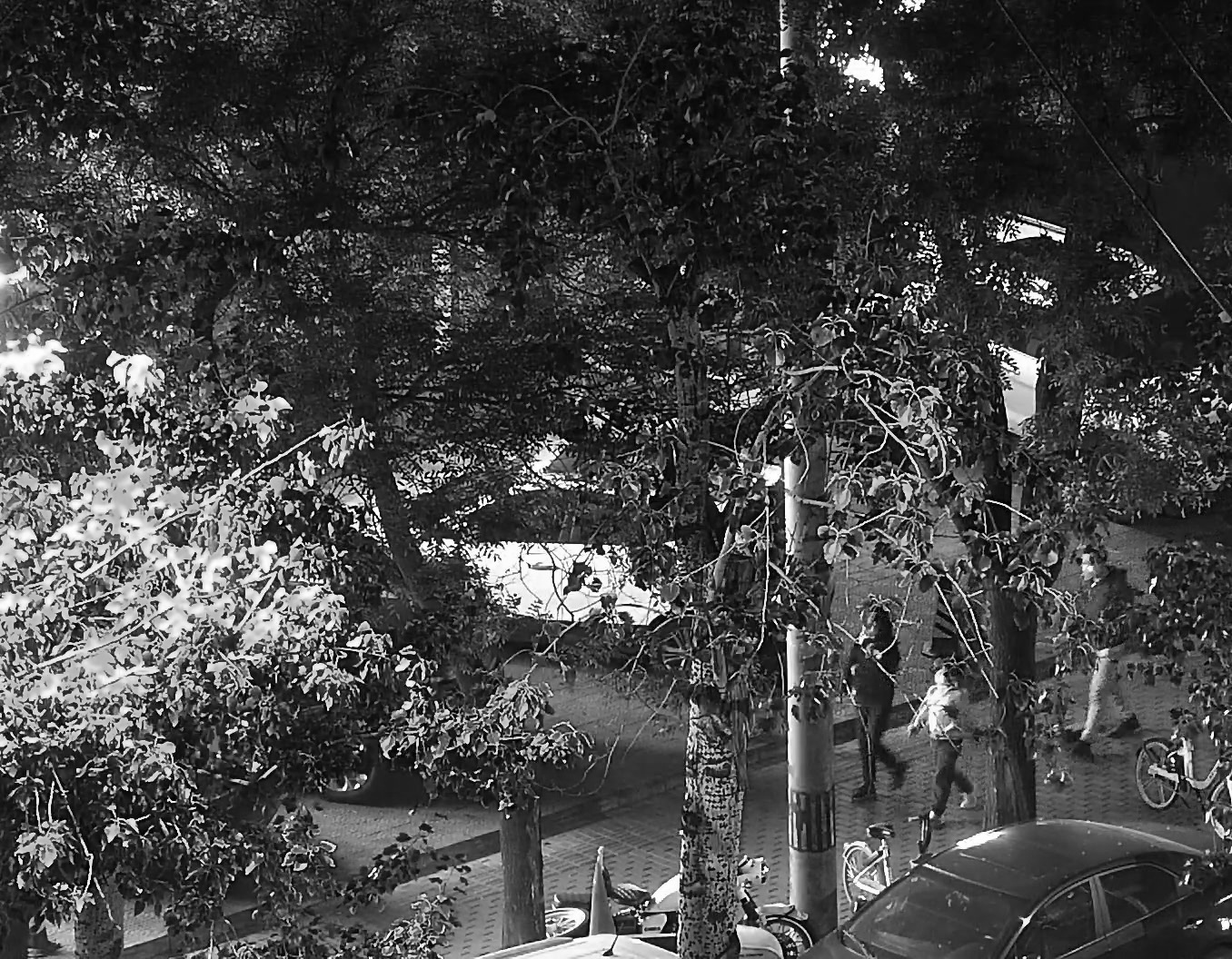}\vspace{2pt}
      \includegraphics[width=\linewidth]{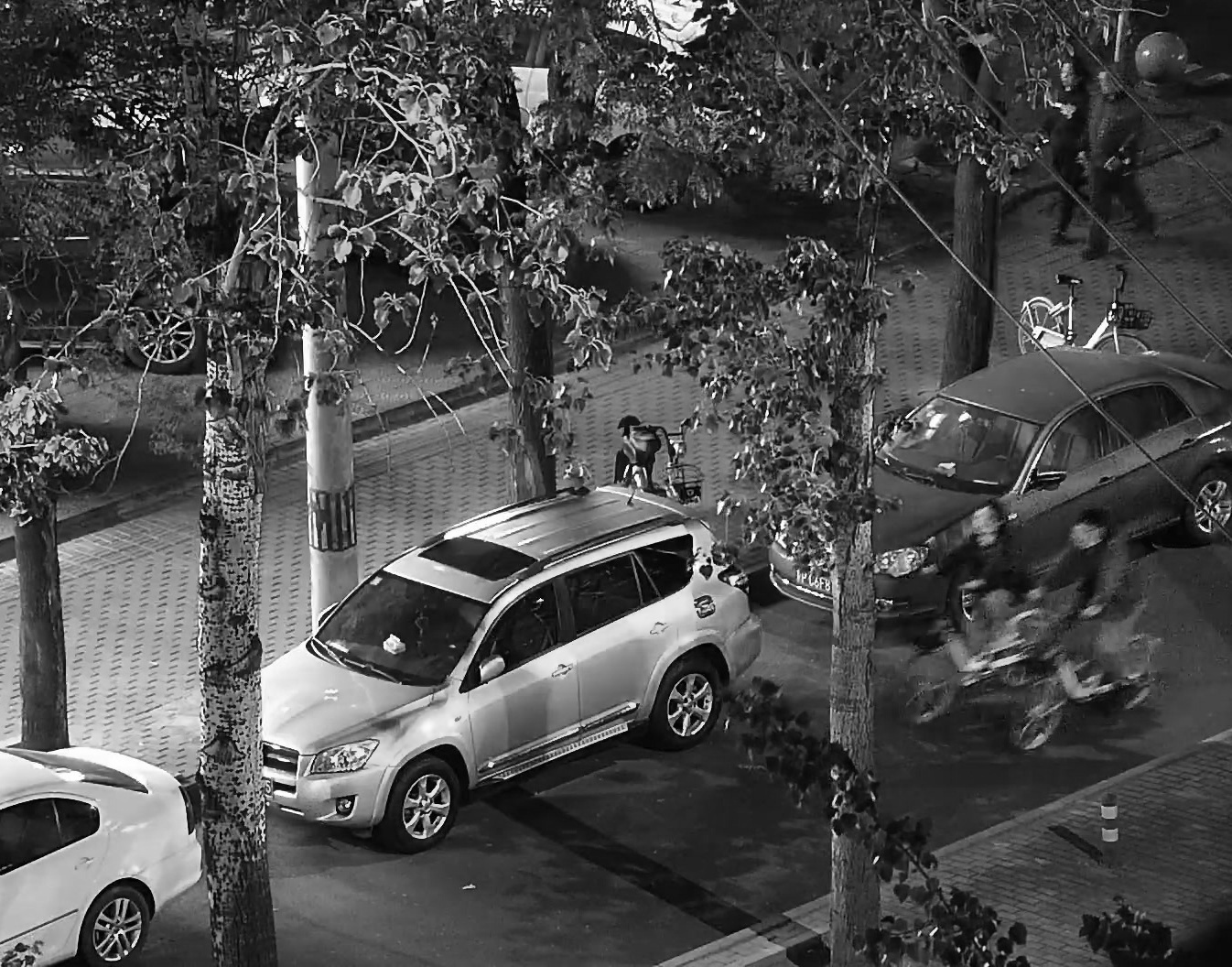}\vspace{2pt}
      \includegraphics[width=\linewidth]{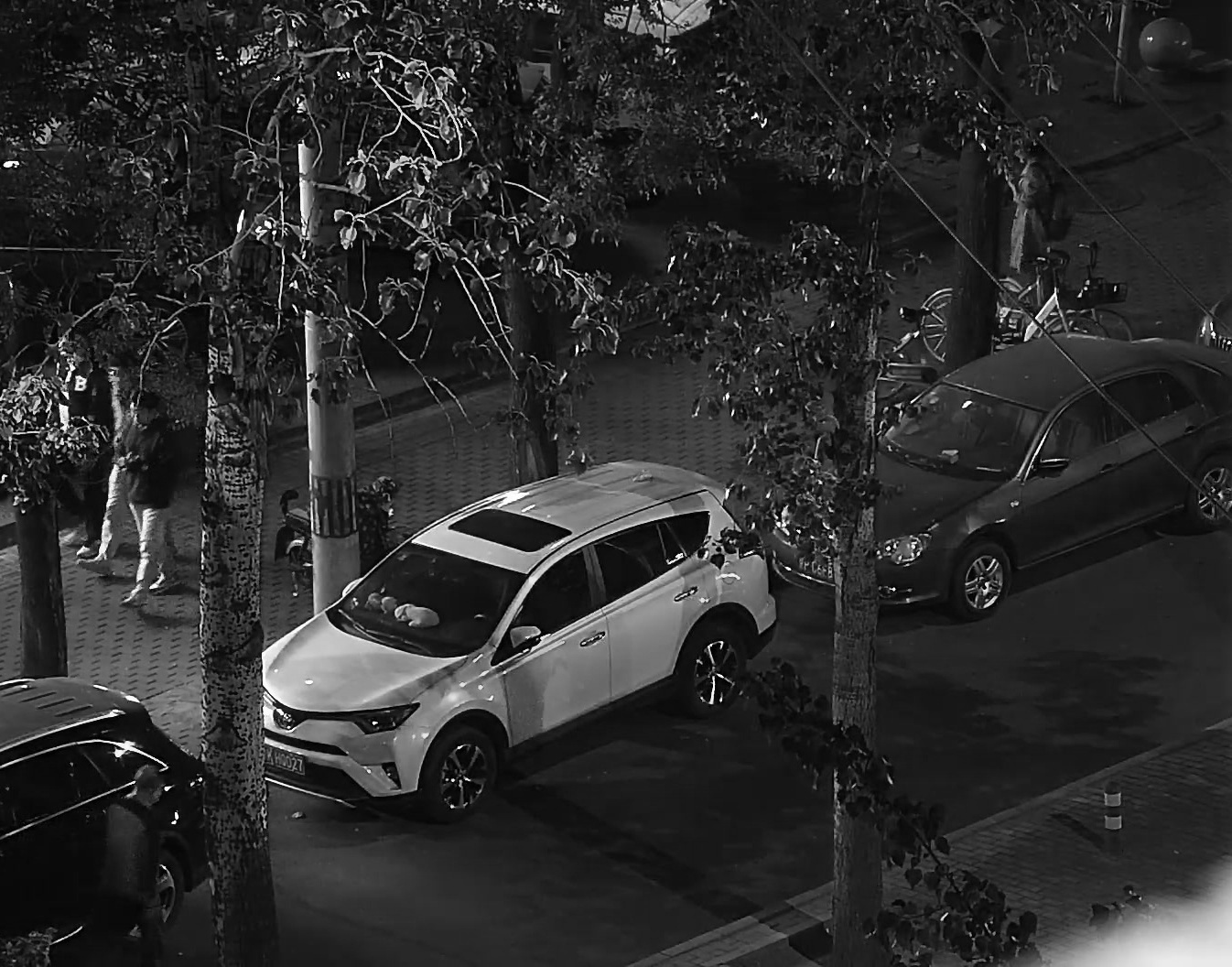}\vspace{2pt}
      \includegraphics[width=\linewidth]{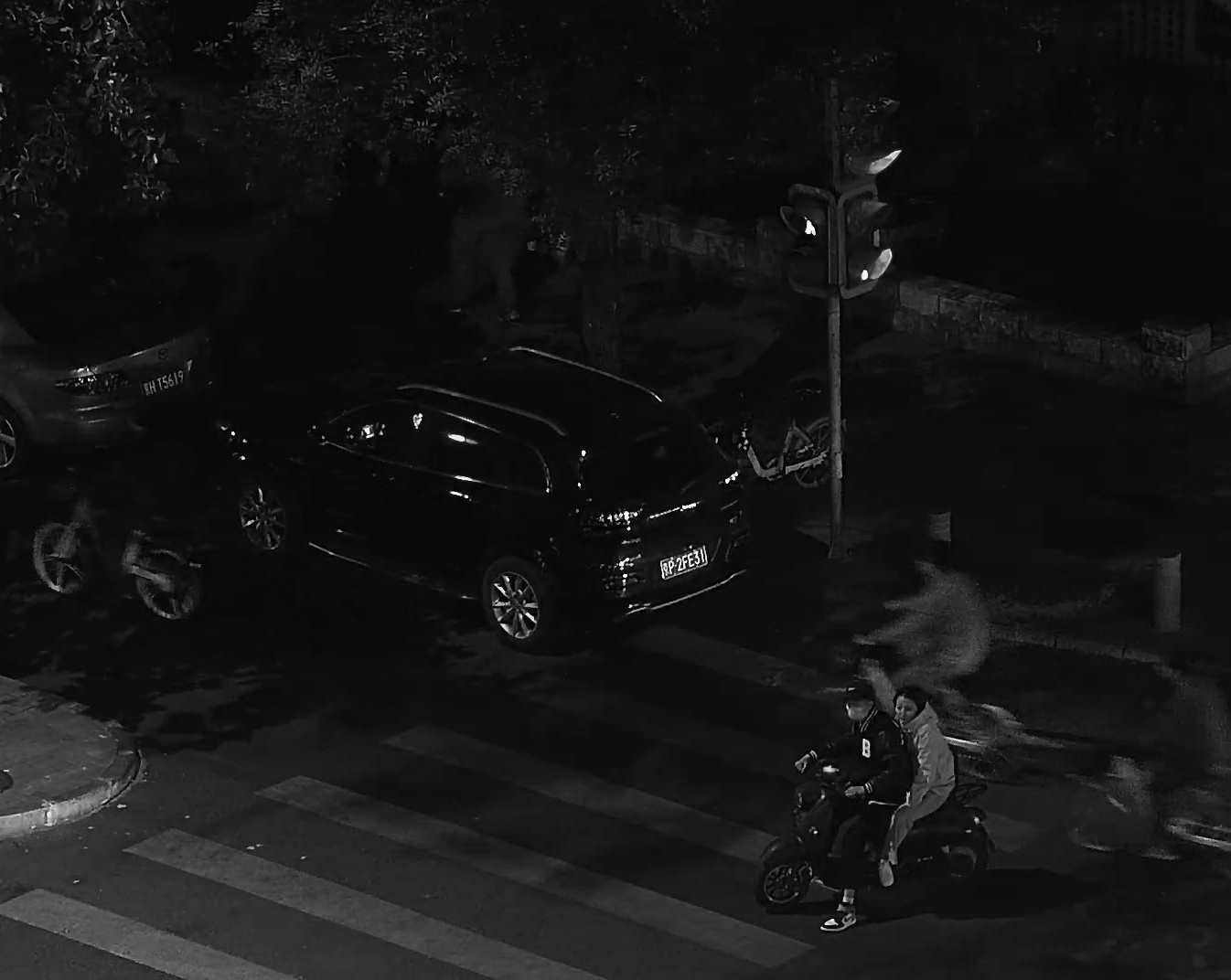}\vspace{2pt}
      \includegraphics[width=\linewidth]{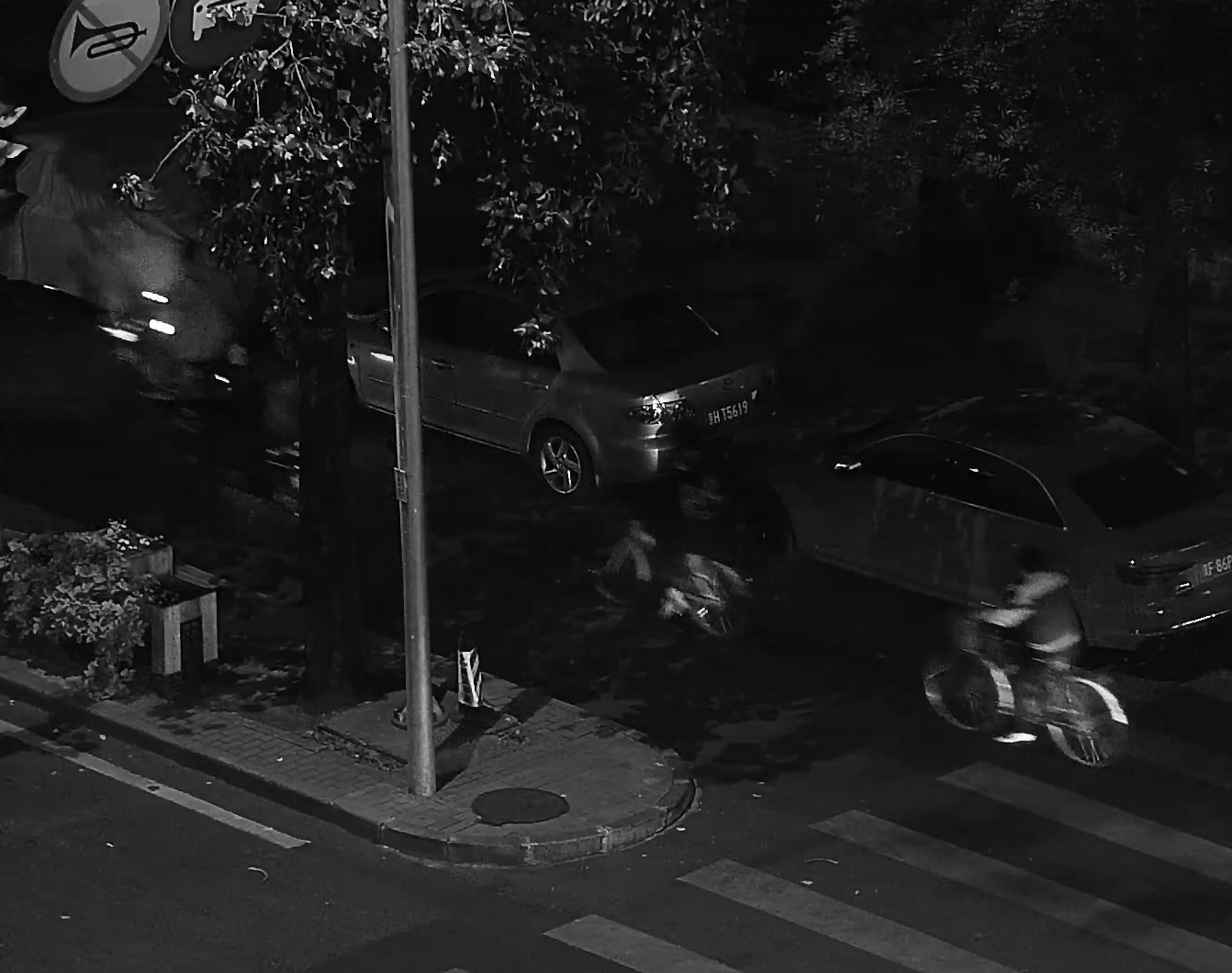}\vspace{2pt}
    \end{minipage}
  }
  \subfigure[infrared]{
    \begin{minipage}[b]{0.127\linewidth}  
      \centering
      \includegraphics[width=\linewidth]{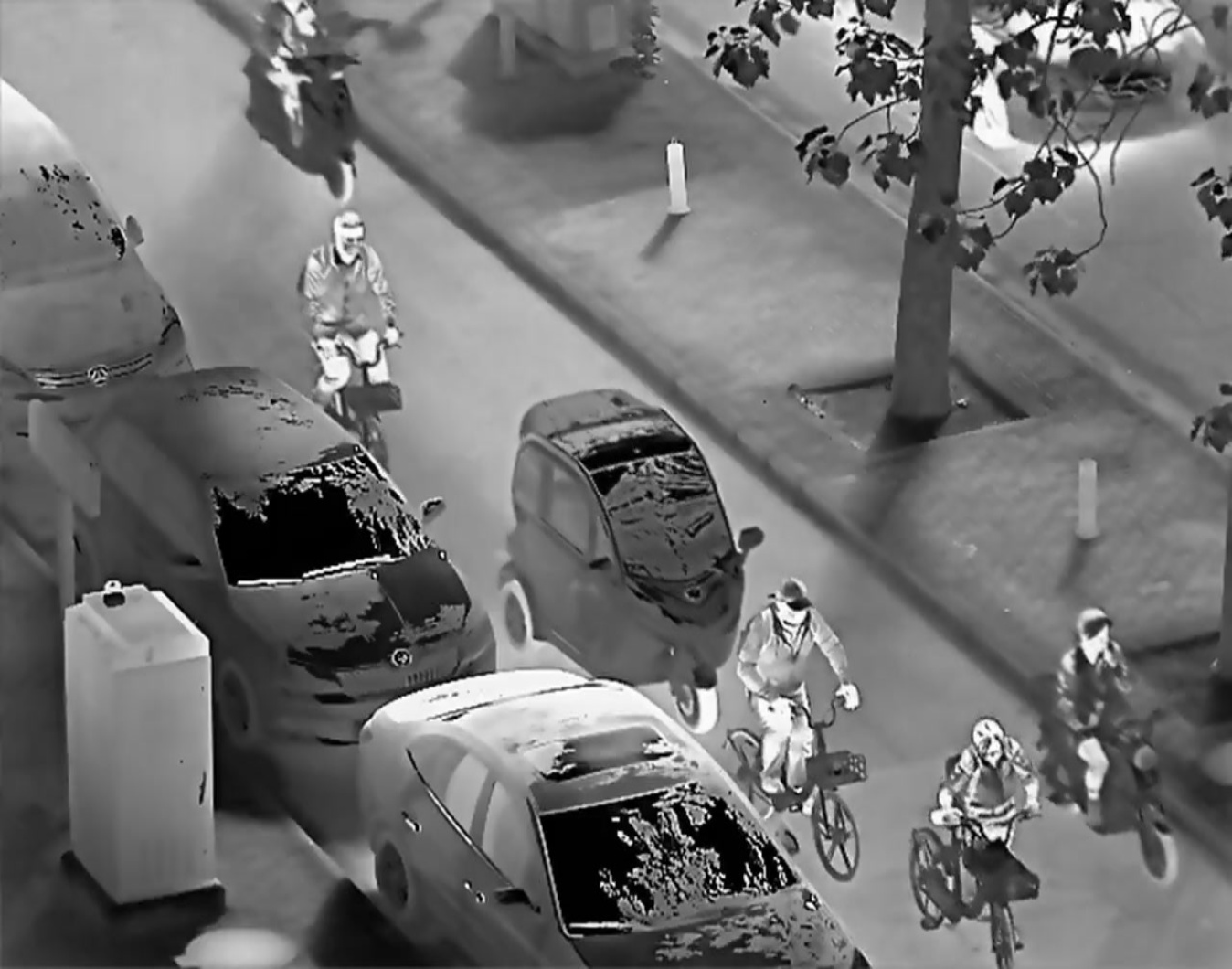}\vspace{2pt}
      \includegraphics[width=\linewidth]{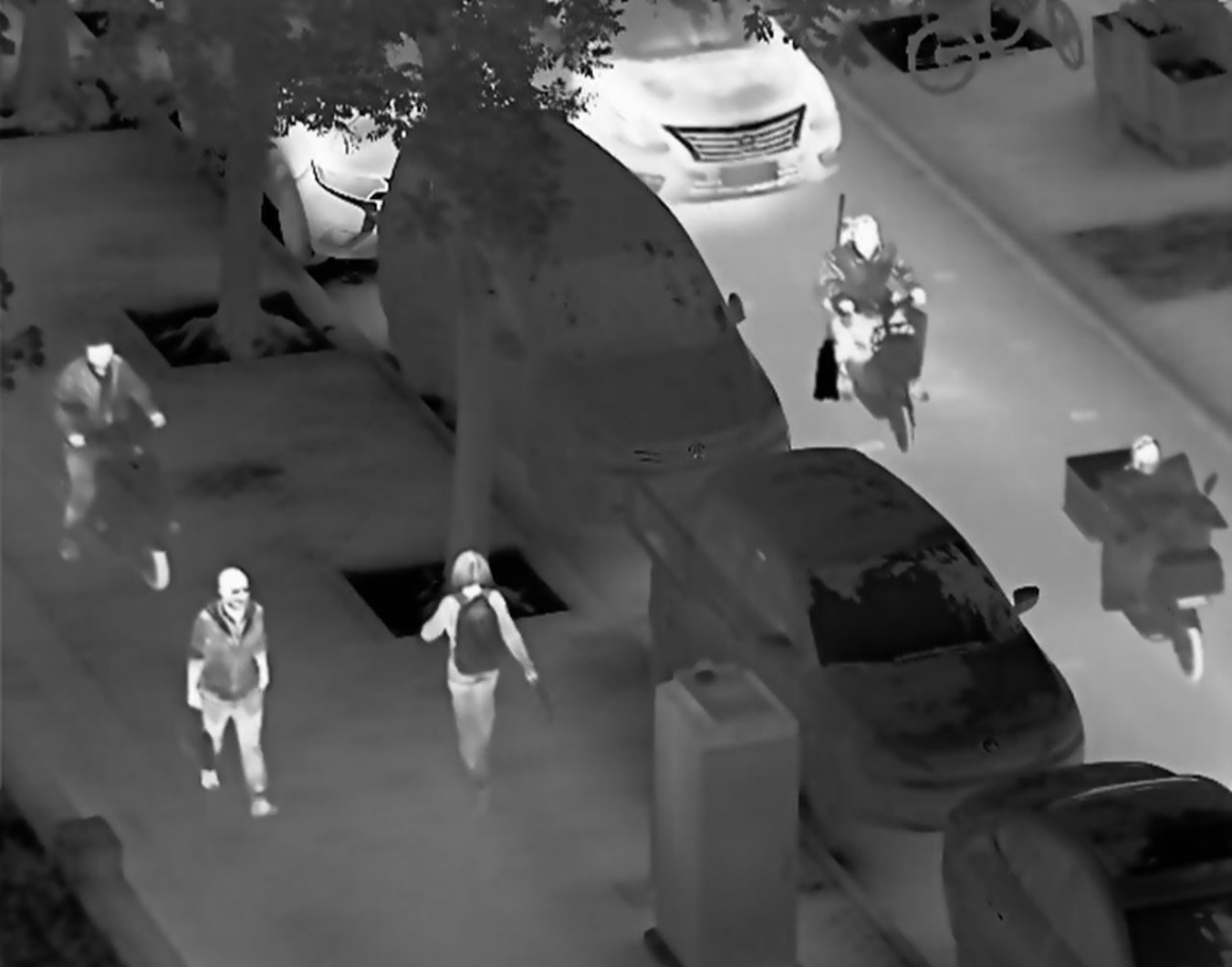}\vspace{2pt}
      \includegraphics[width=\linewidth]{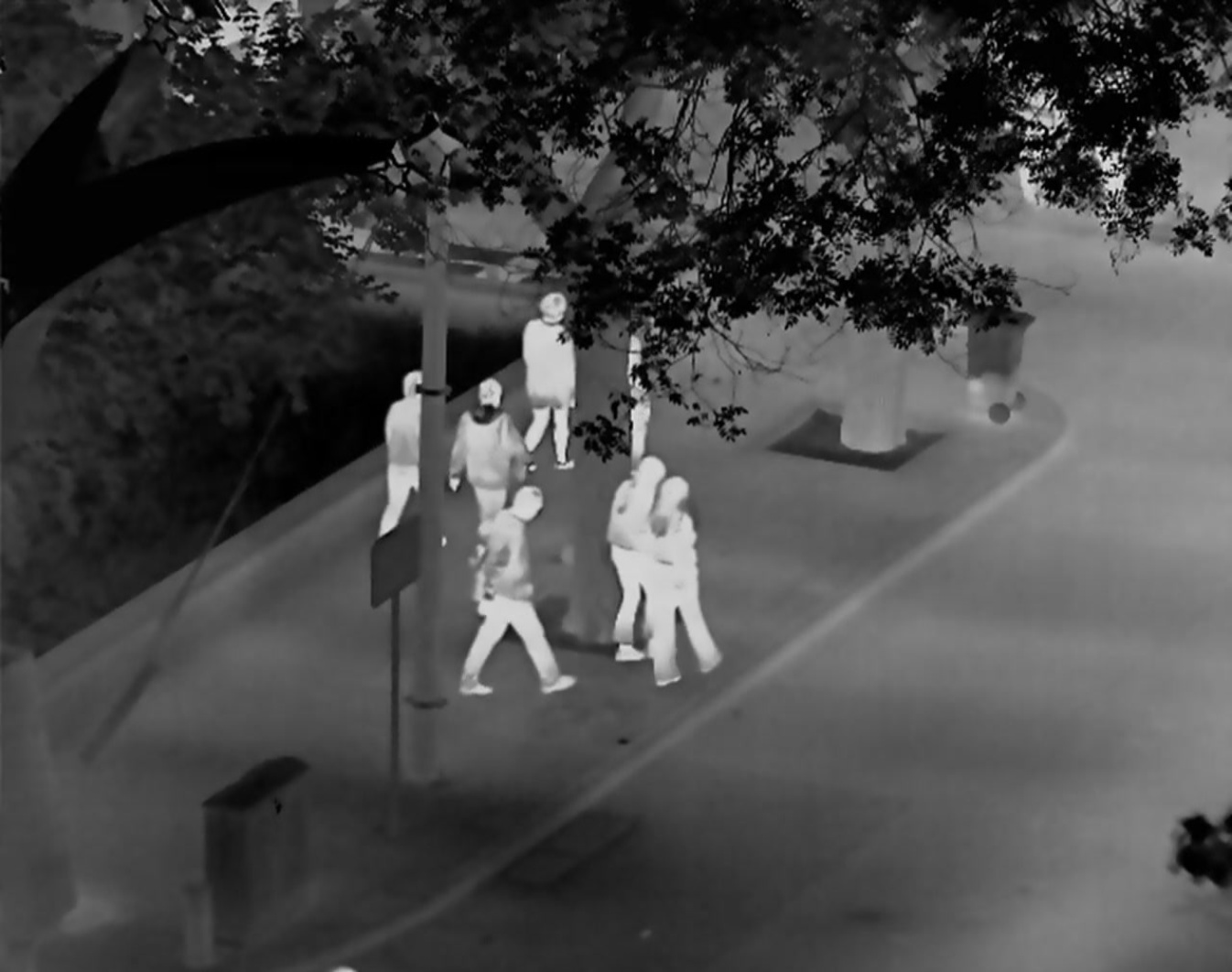}\vspace{2pt}
      \includegraphics[width=\linewidth]{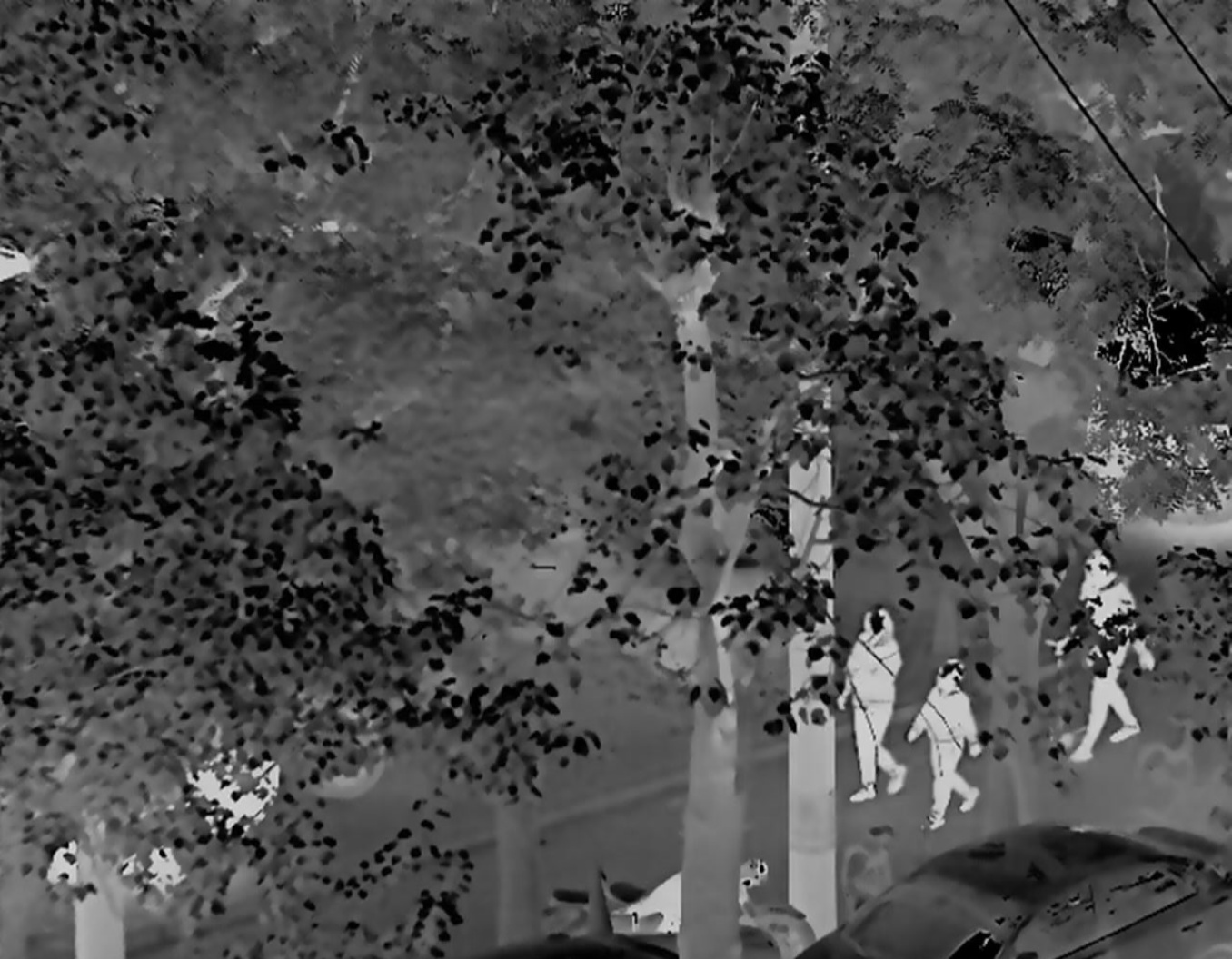}\vspace{2pt}
      \includegraphics[width=\linewidth]{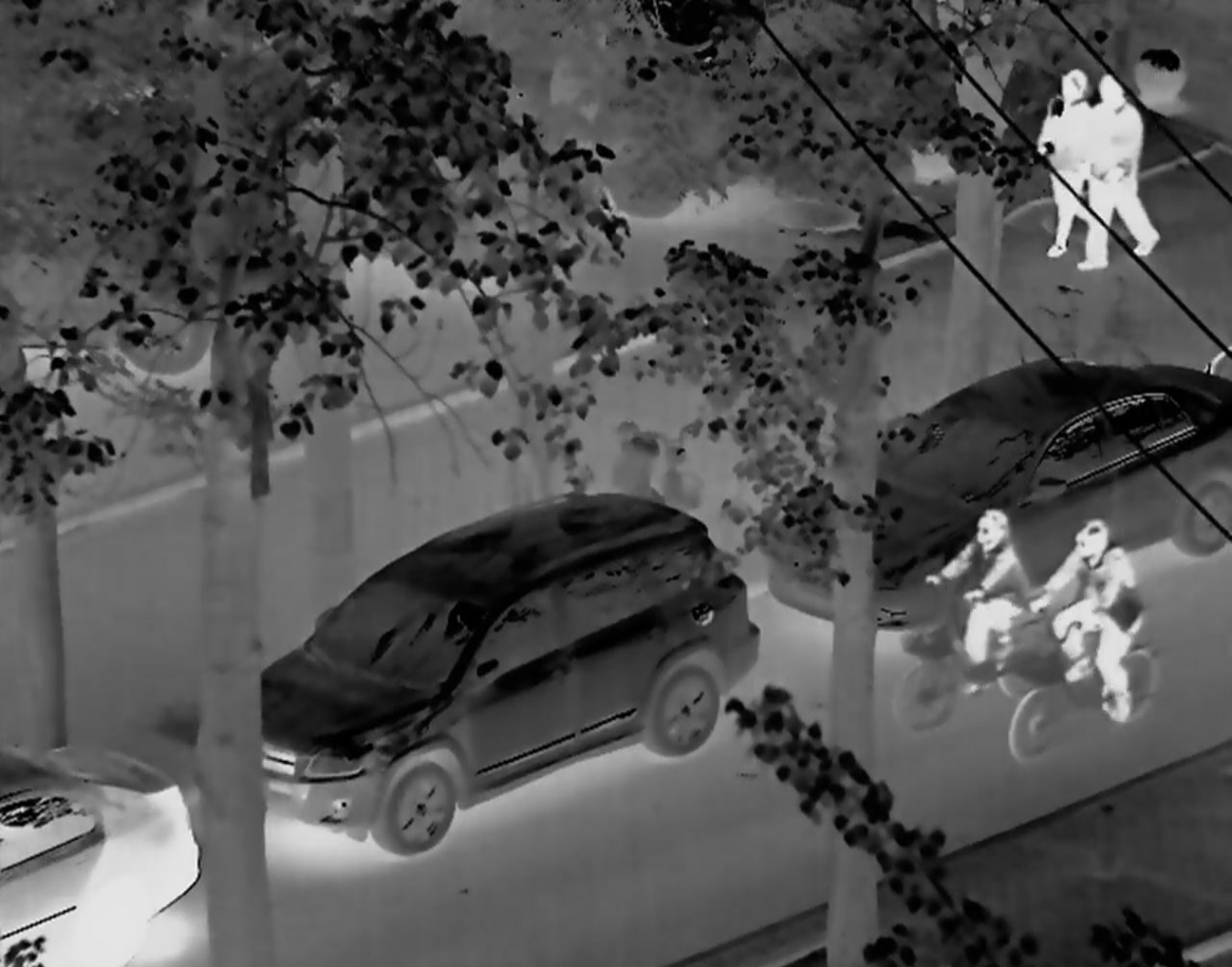}\vspace{2pt}
      \includegraphics[width=\linewidth]{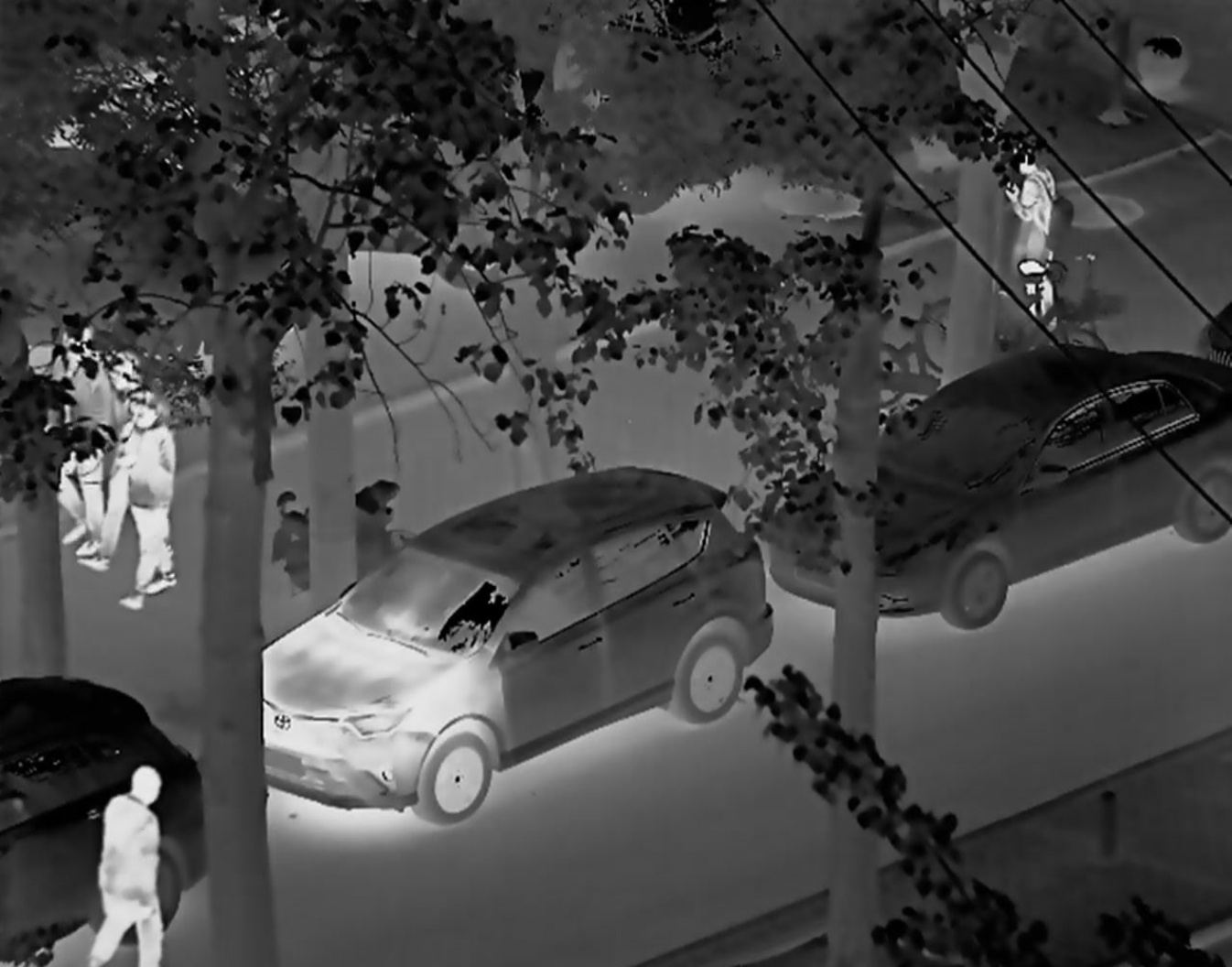}\vspace{2pt}
      \includegraphics[width=\linewidth]{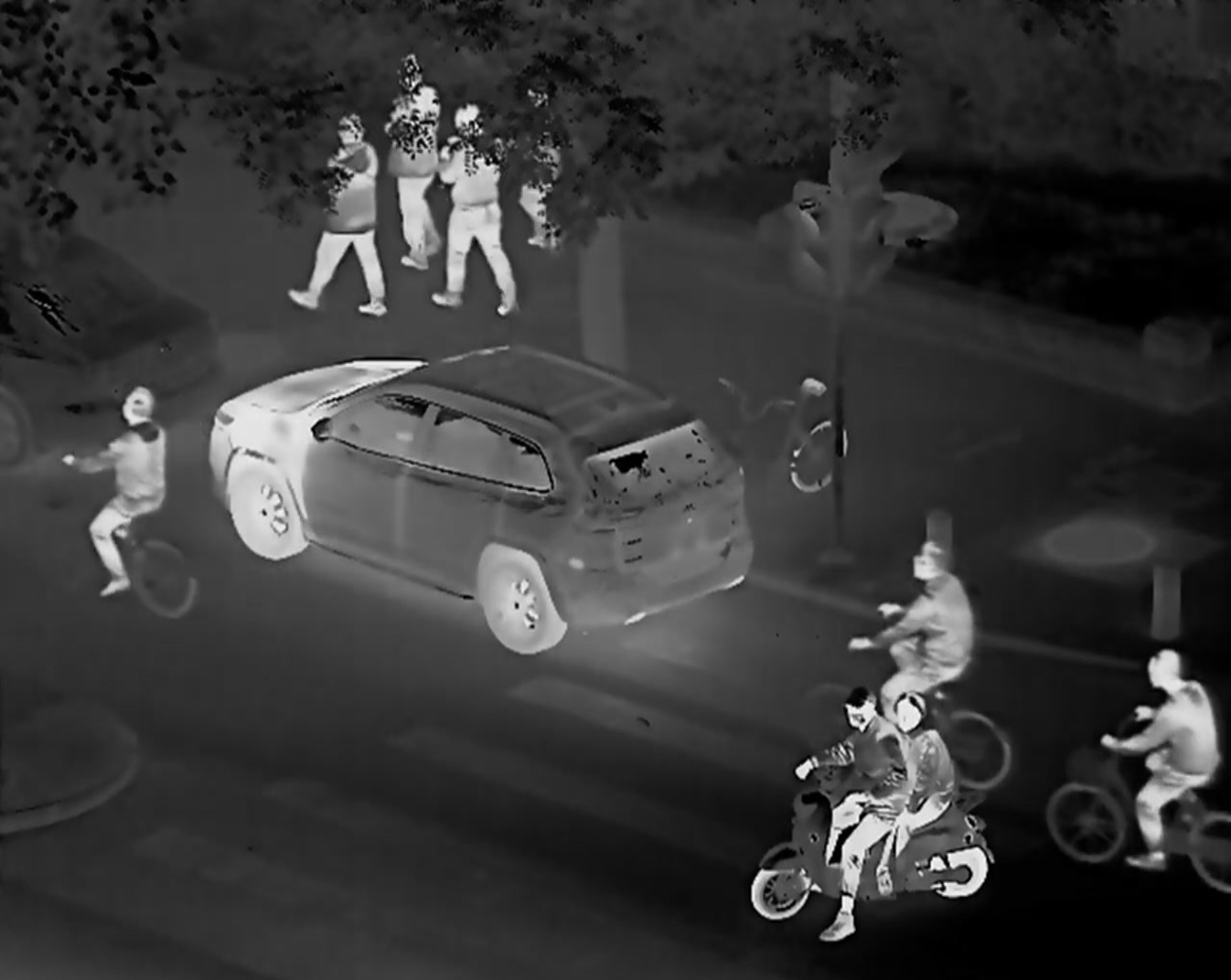}\vspace{2pt}
      \includegraphics[width=\linewidth]{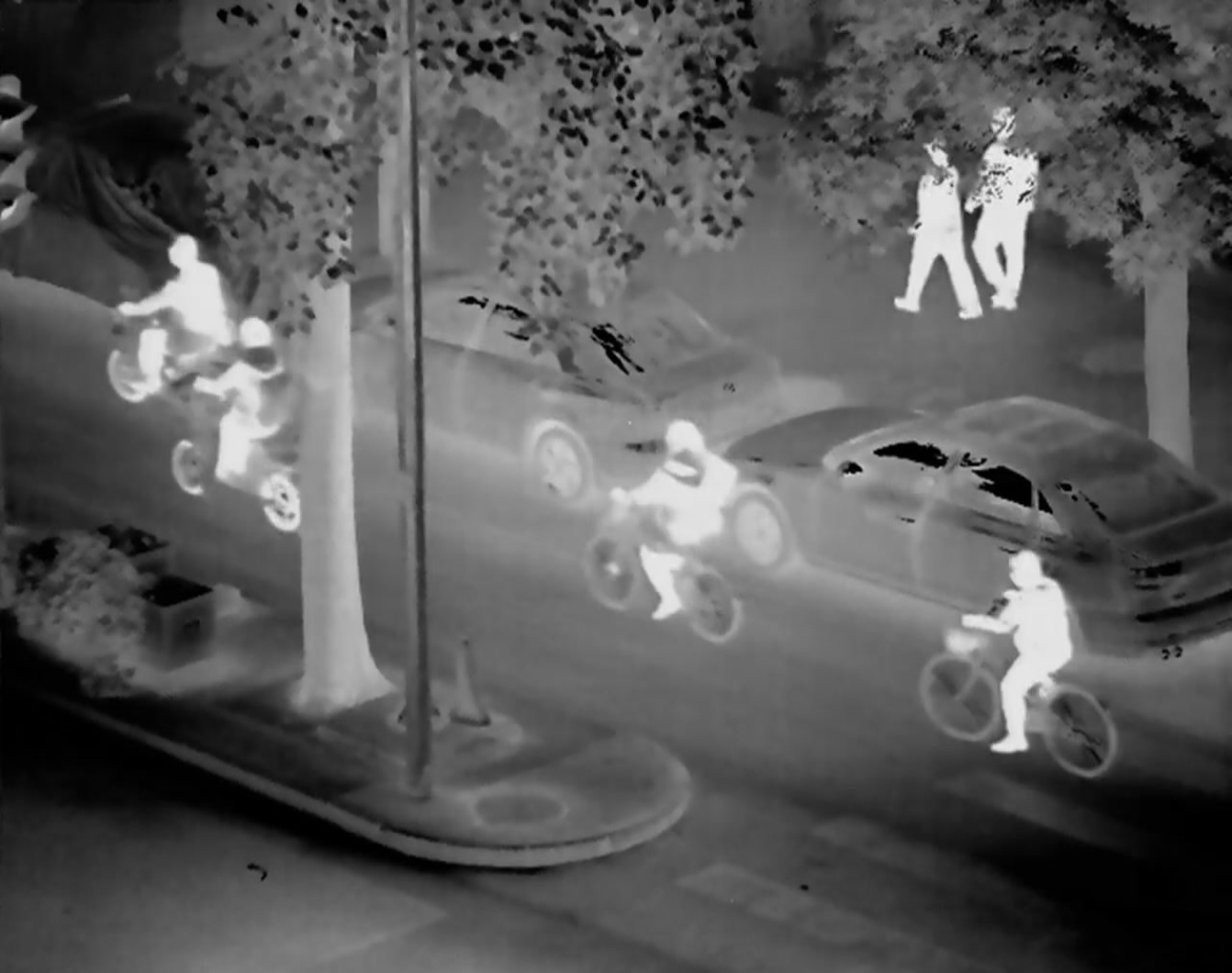}\vspace{2pt}
    \end{minipage}
  }
  \subfigure[GTF]{
    \begin{minipage}[b]{0.127\linewidth}  
      \centering
      \includegraphics[width=\linewidth]{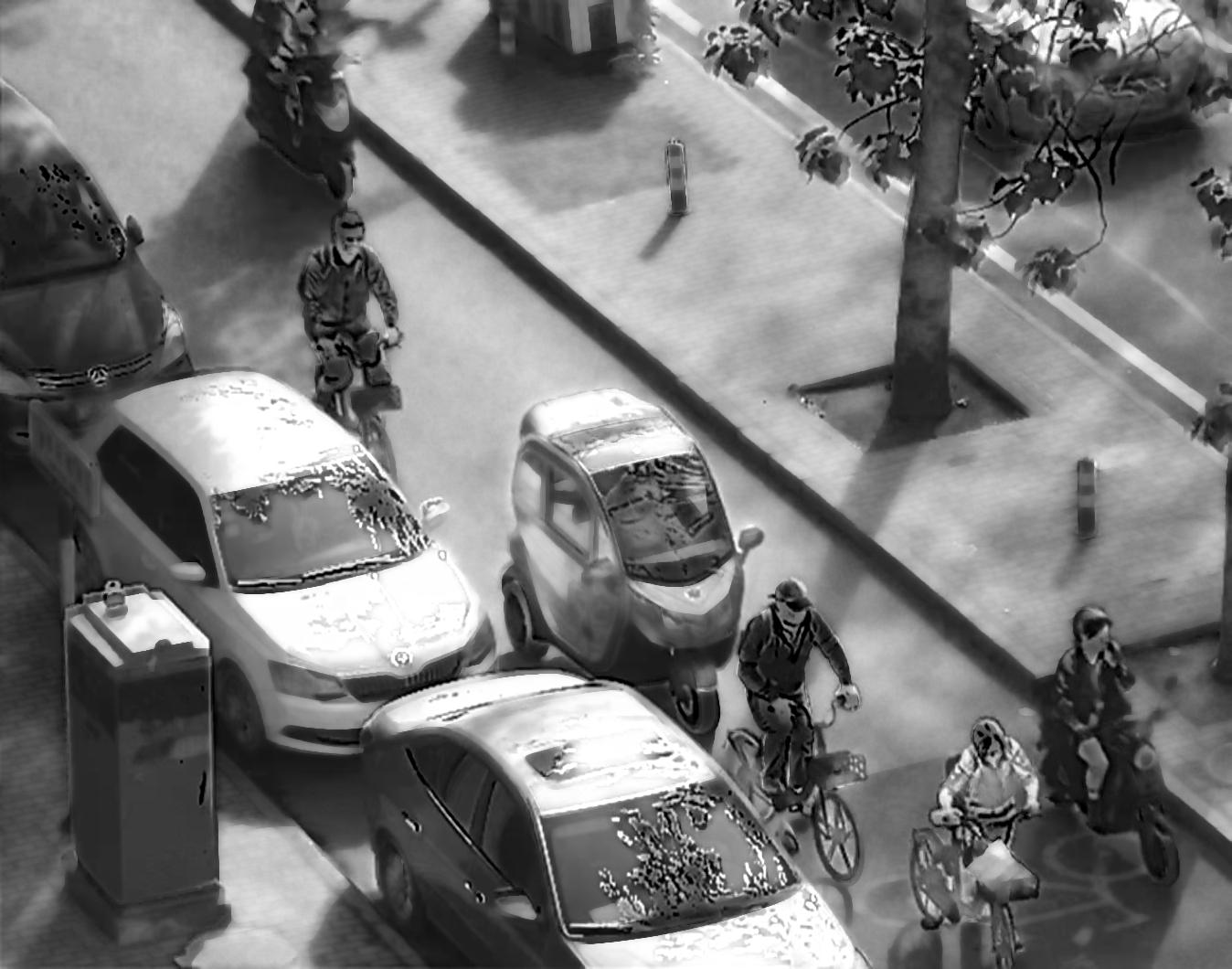}\vspace{2pt}
      \includegraphics[width=\linewidth]{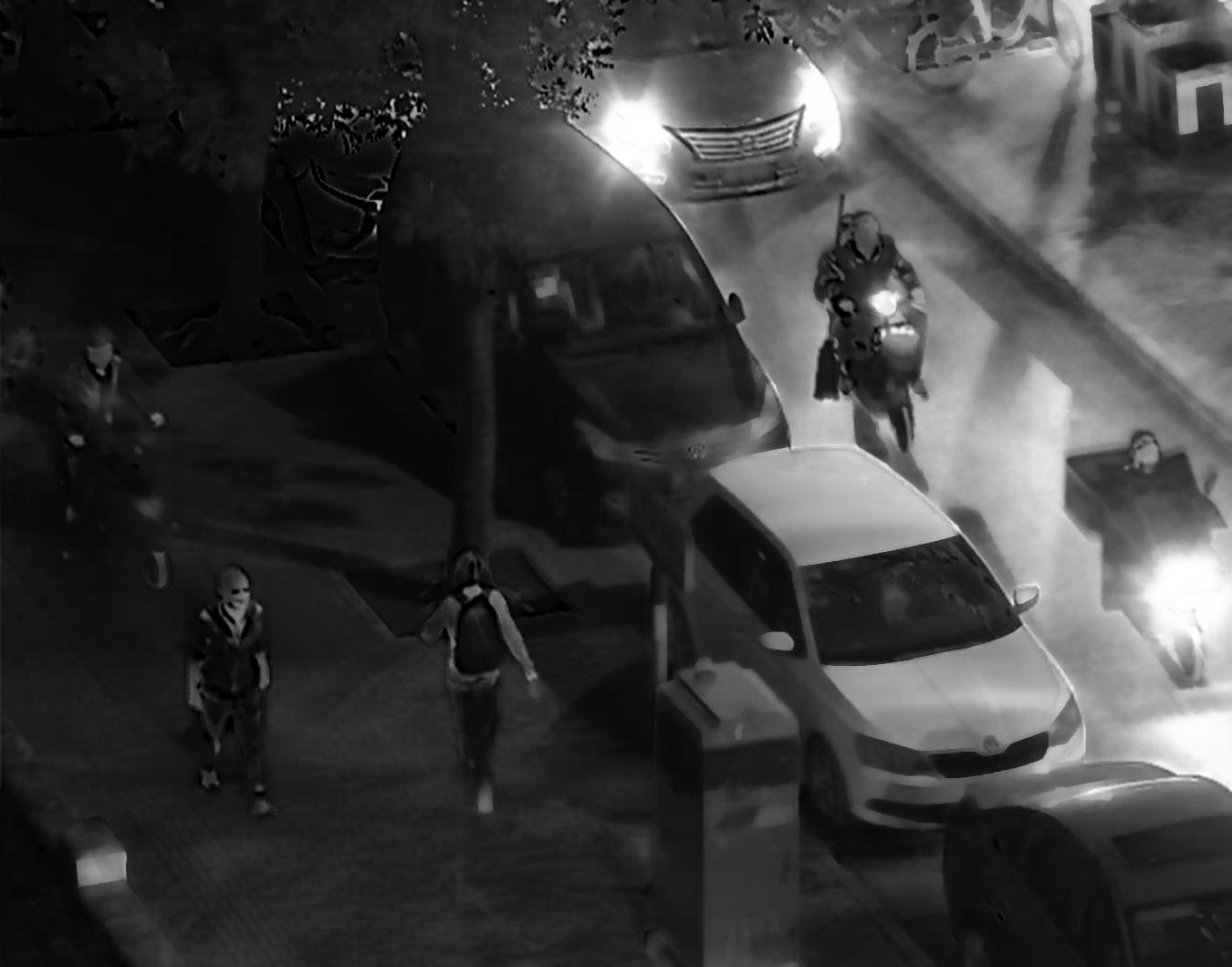}\vspace{2pt}
      \includegraphics[width=\linewidth]{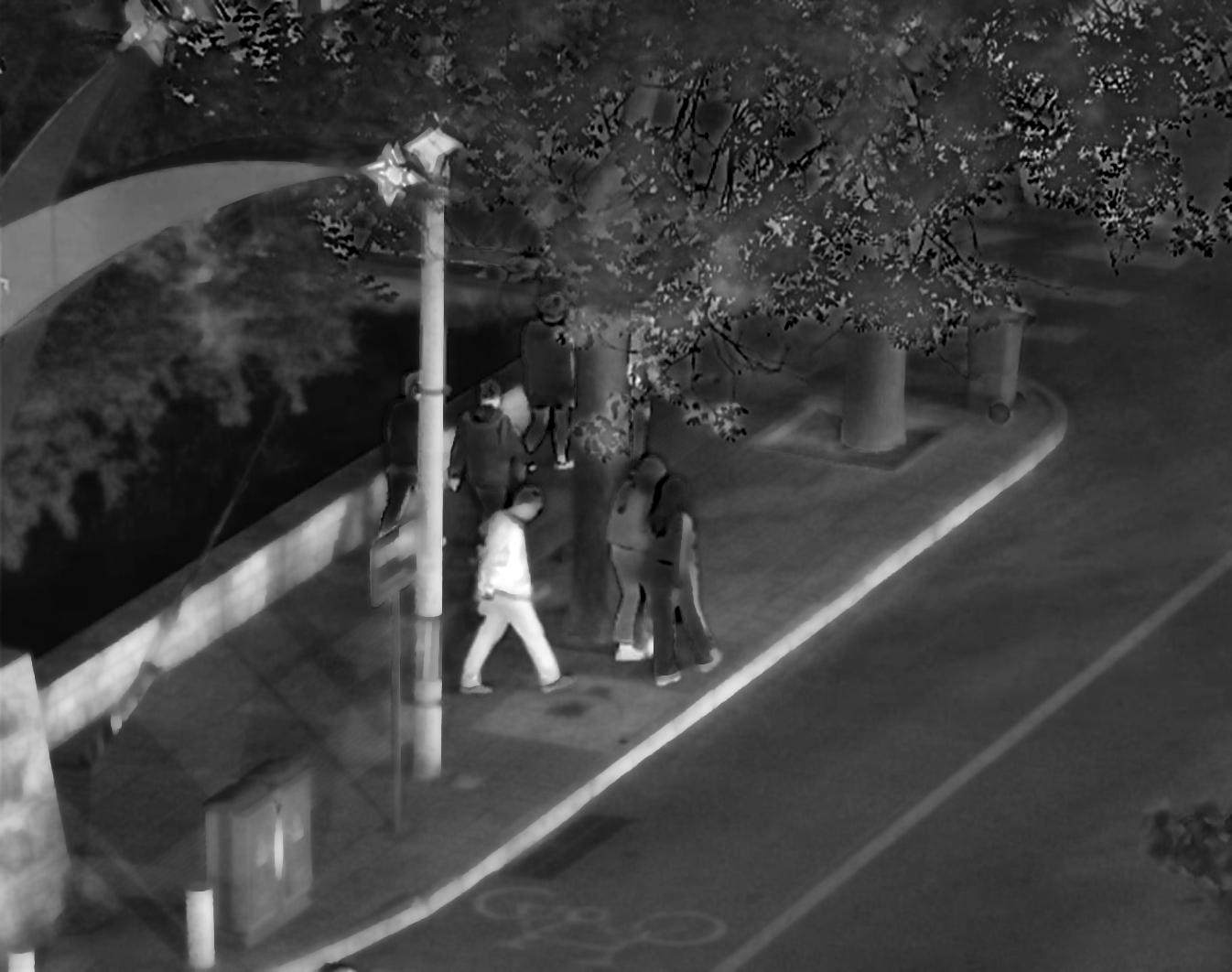}\vspace{2pt}
      \includegraphics[width=\linewidth]{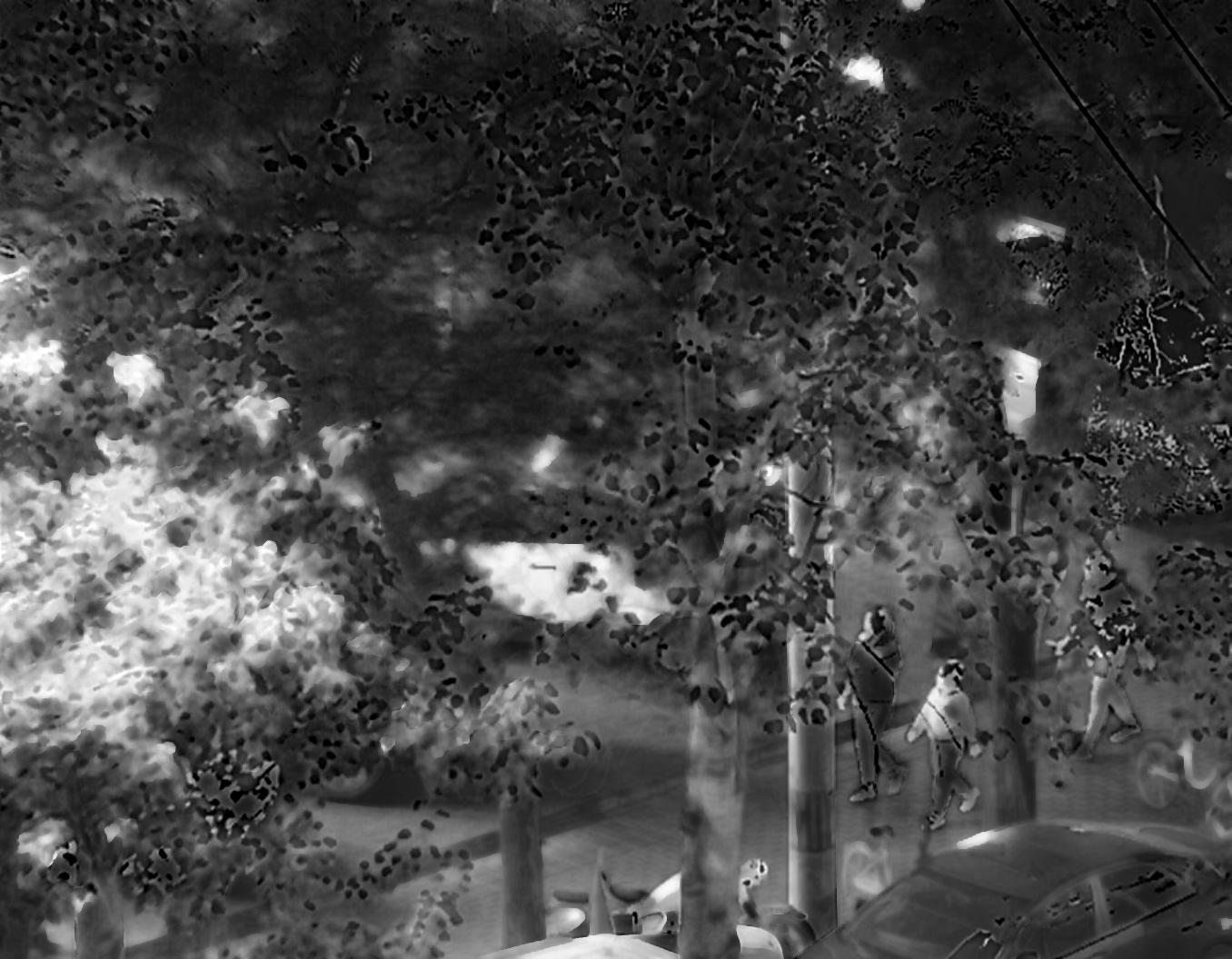}\vspace{2pt}
      \includegraphics[width=\linewidth]{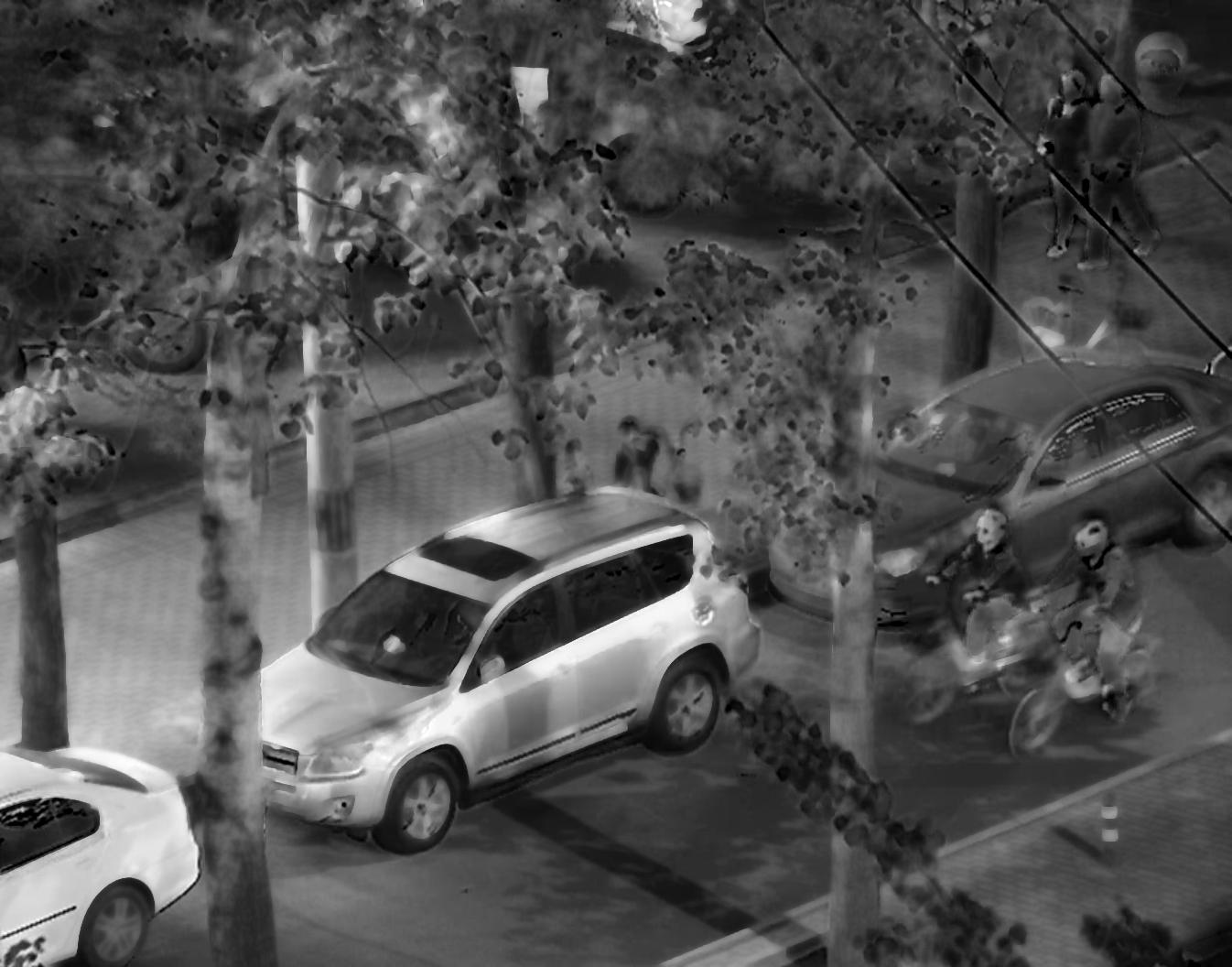}\vspace{2pt}
      \includegraphics[width=\linewidth]{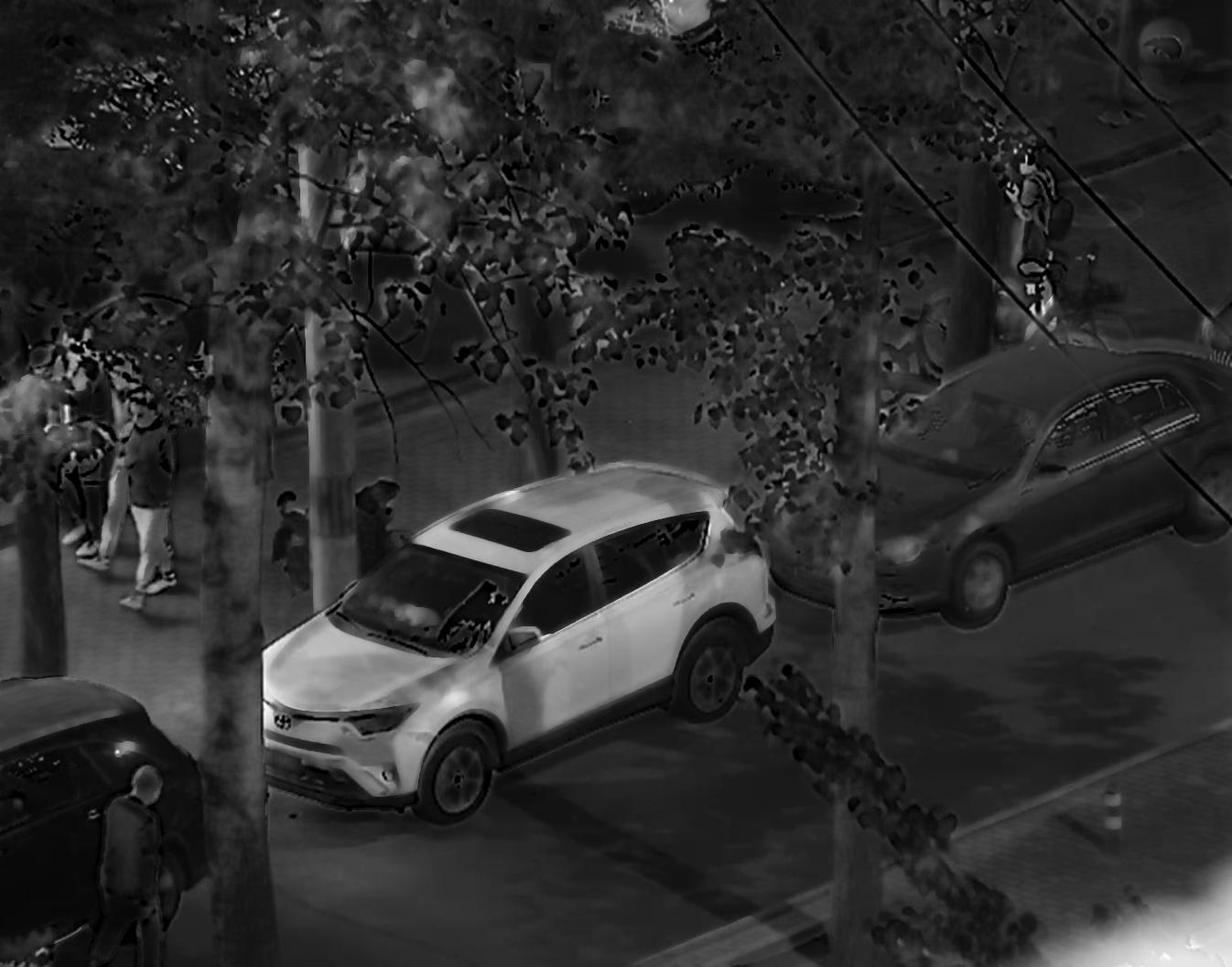}\vspace{2pt}
      \includegraphics[width=\linewidth]{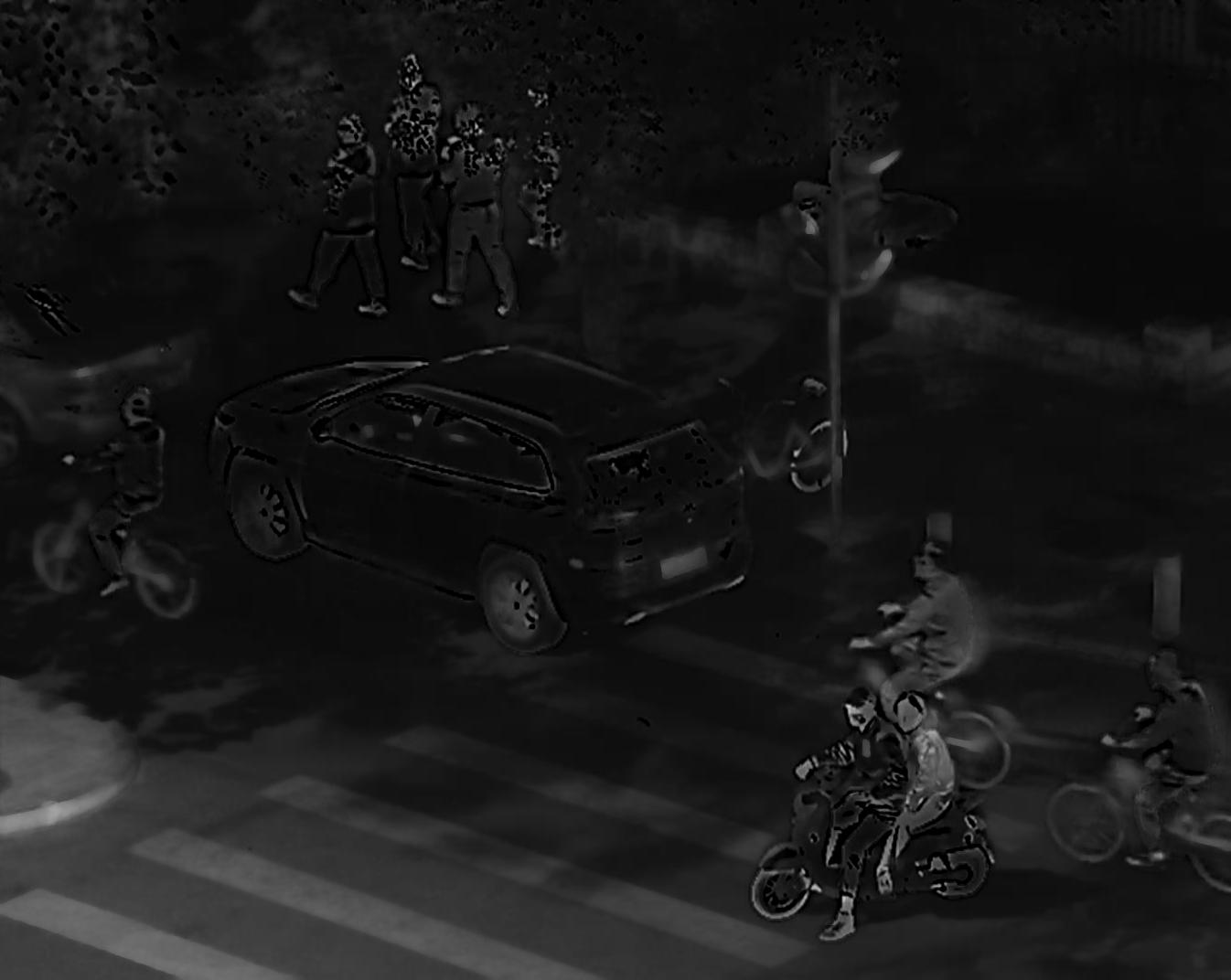}\vspace{2pt}
      \includegraphics[width=\linewidth]{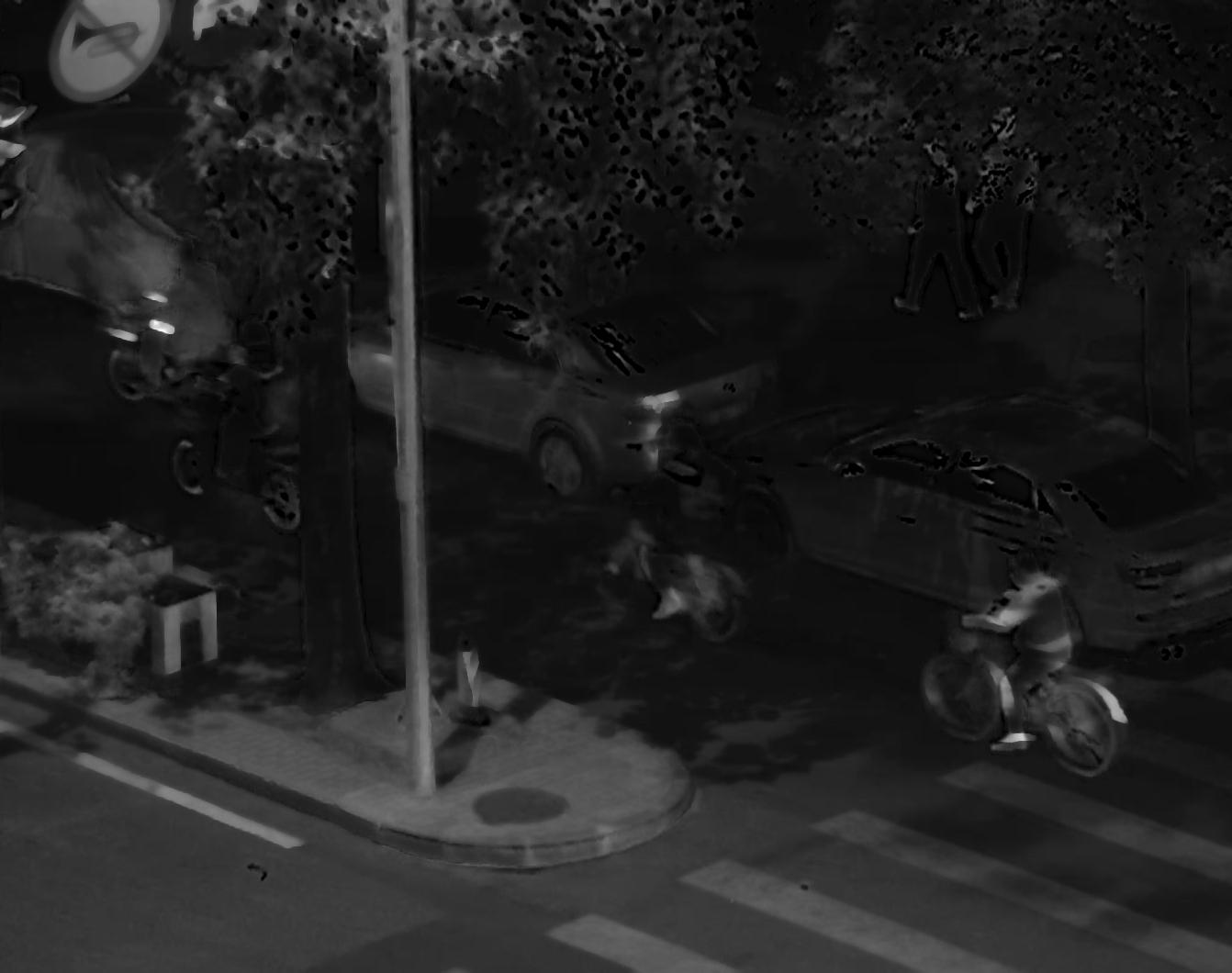}\vspace{2pt}
    \end{minipage}
  }
  \subfigure[densefuse\_add]{
    \begin{minipage}[b]{0.127\linewidth}  
      \centering
      \includegraphics[width=\linewidth]{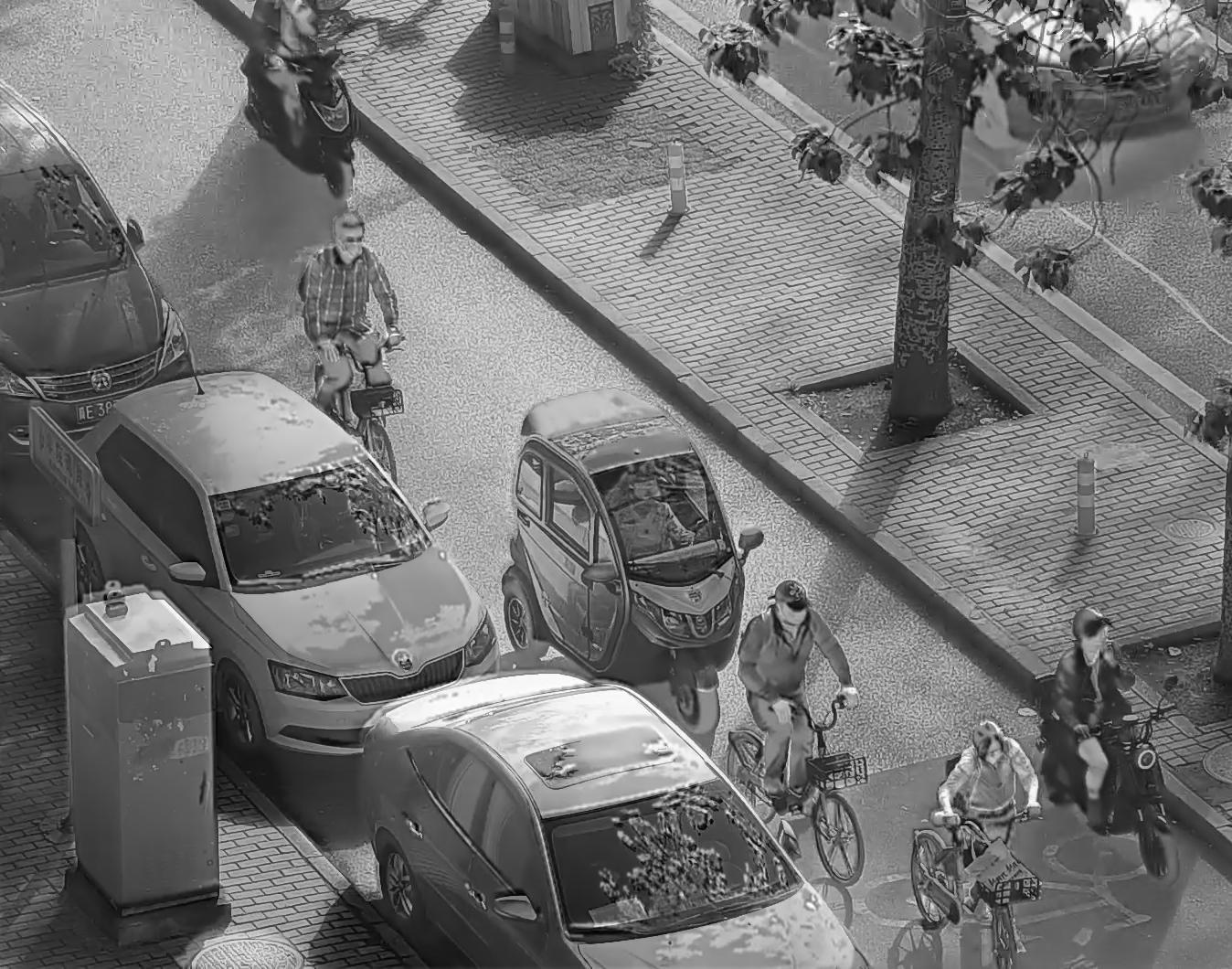}\vspace{2pt}
      \includegraphics[width=\linewidth]{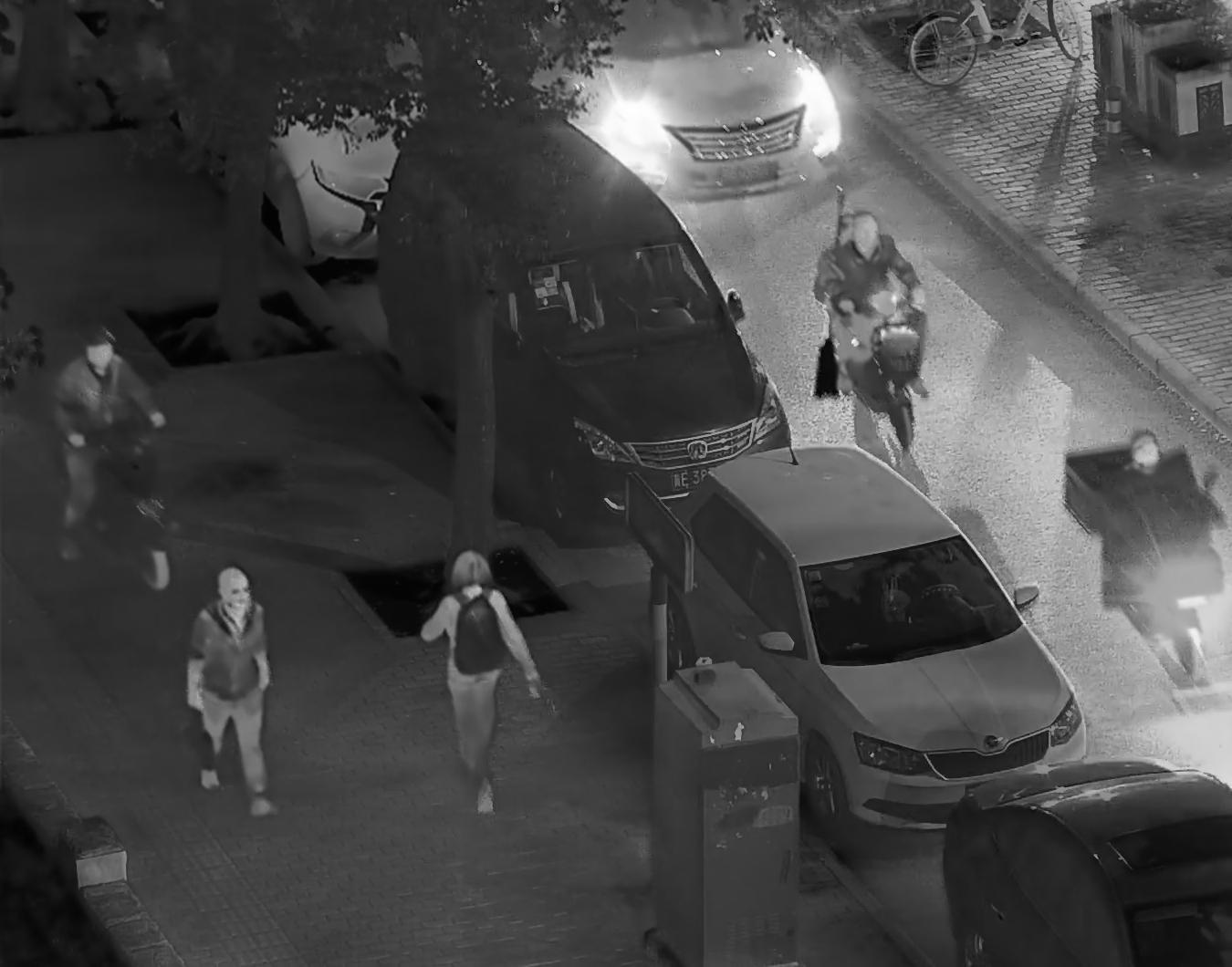}\vspace{2pt}
      \includegraphics[width=\linewidth]{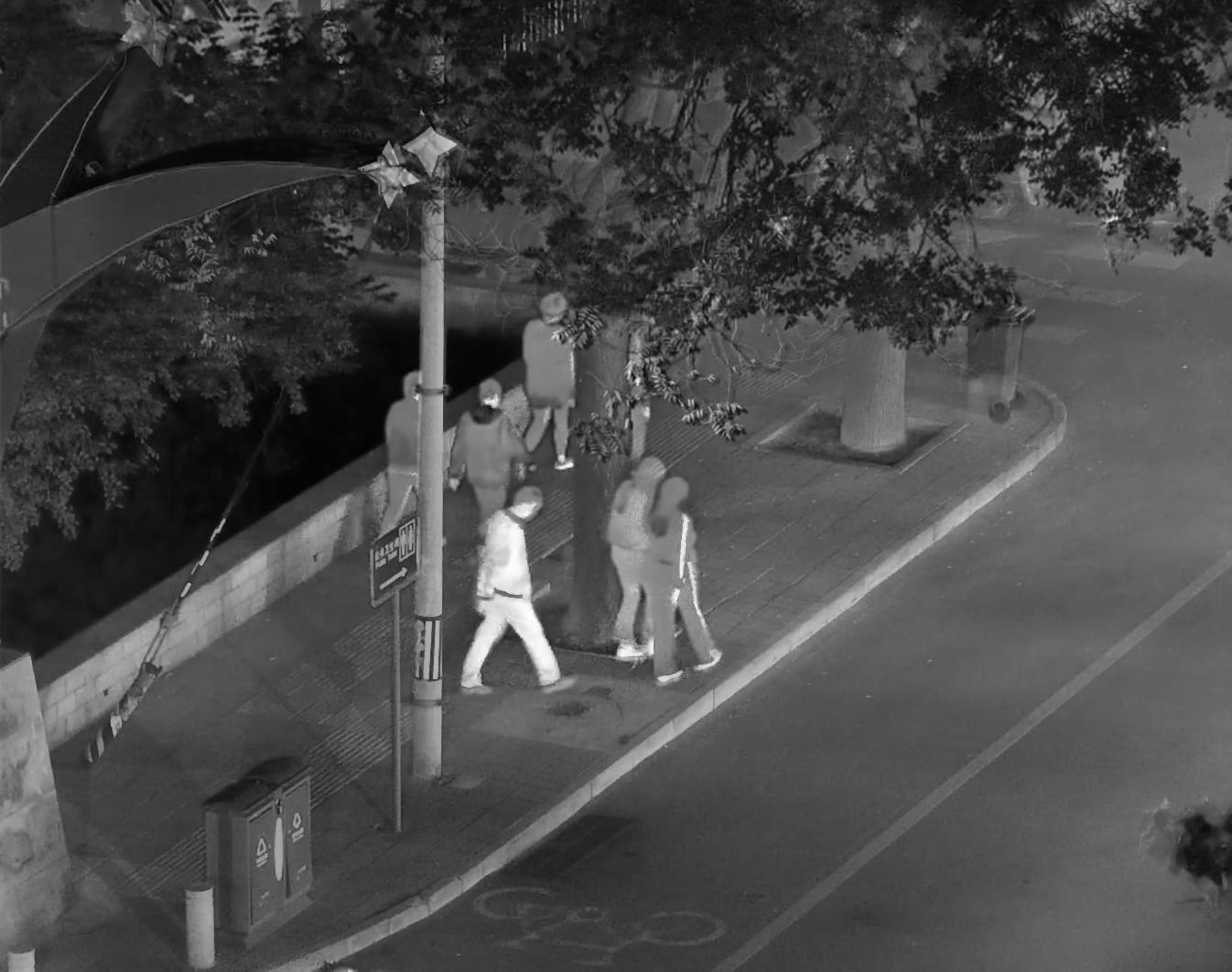}\vspace{2pt}
      \includegraphics[width=\linewidth]{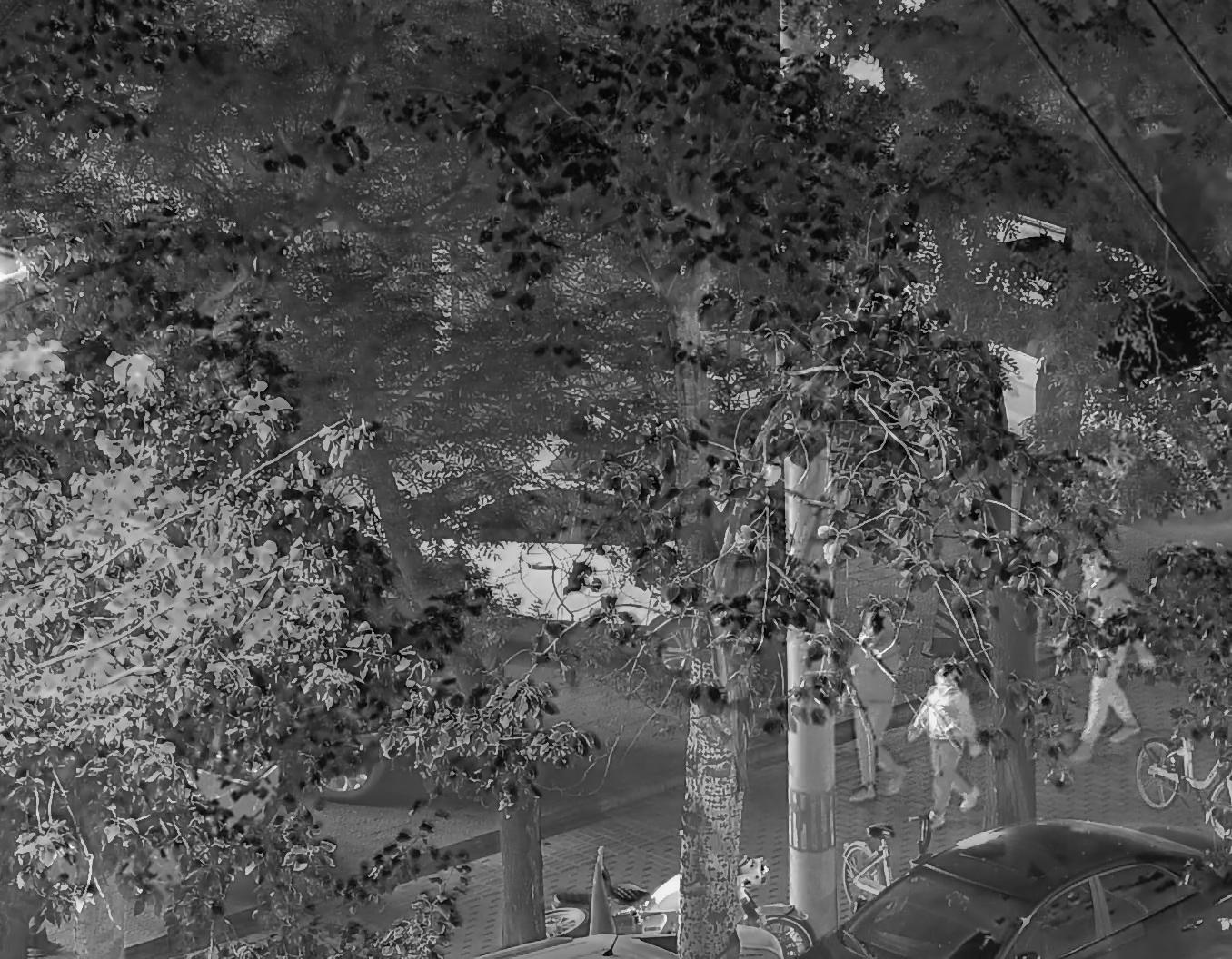}\vspace{2pt}
      \includegraphics[width=\linewidth]{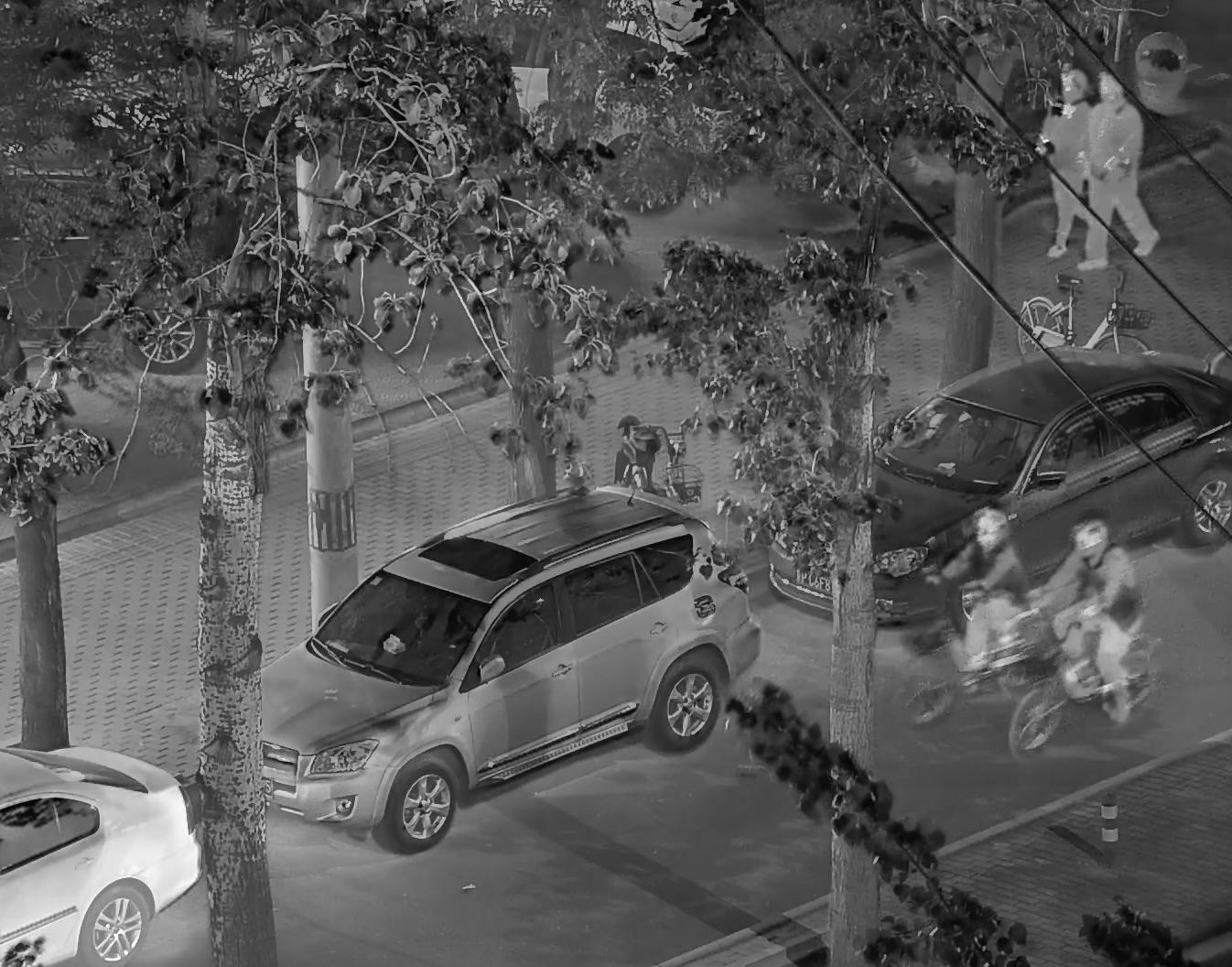}\vspace{2pt}
      \includegraphics[width=\linewidth]{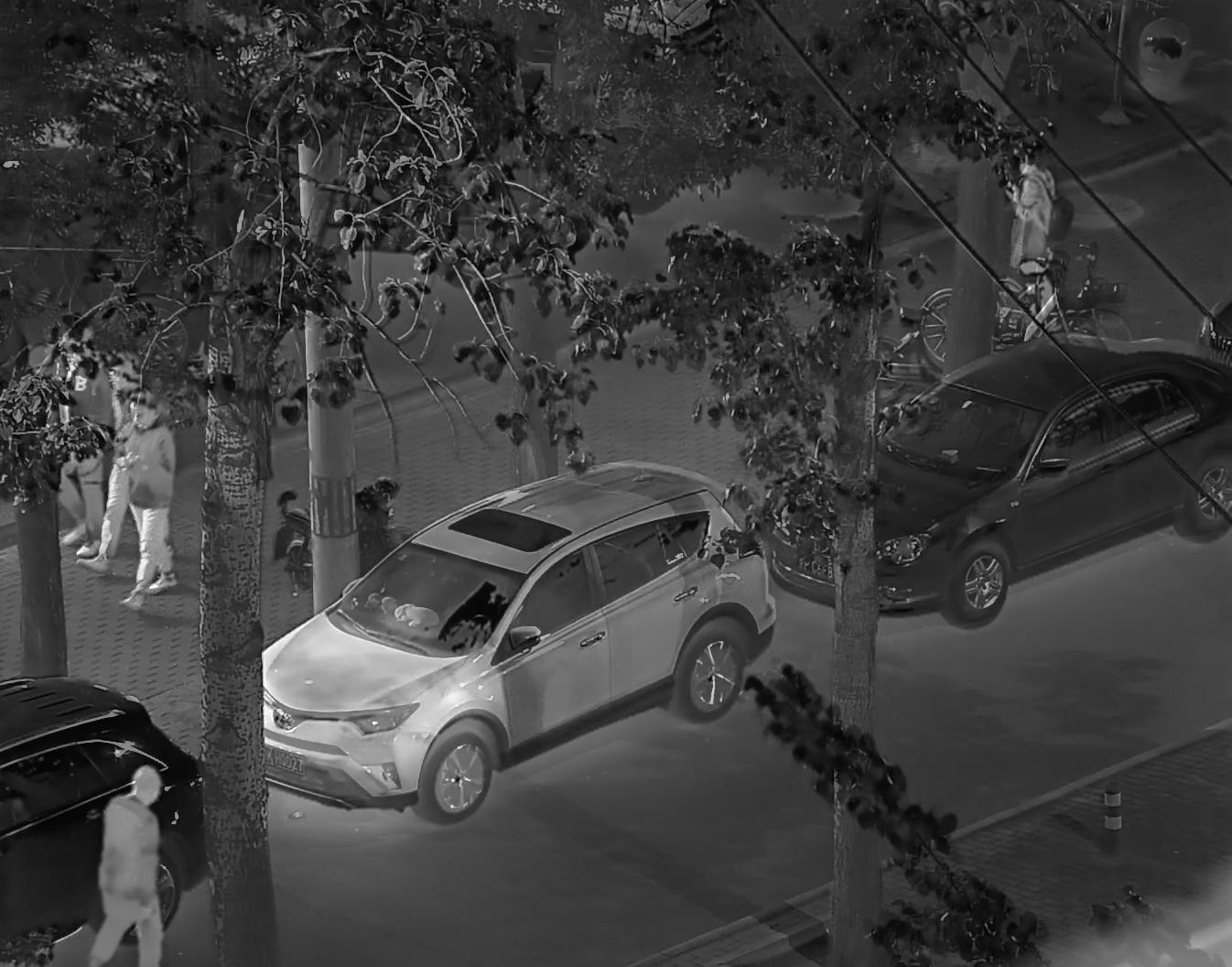}\vspace{2pt}
      \includegraphics[width=\linewidth]{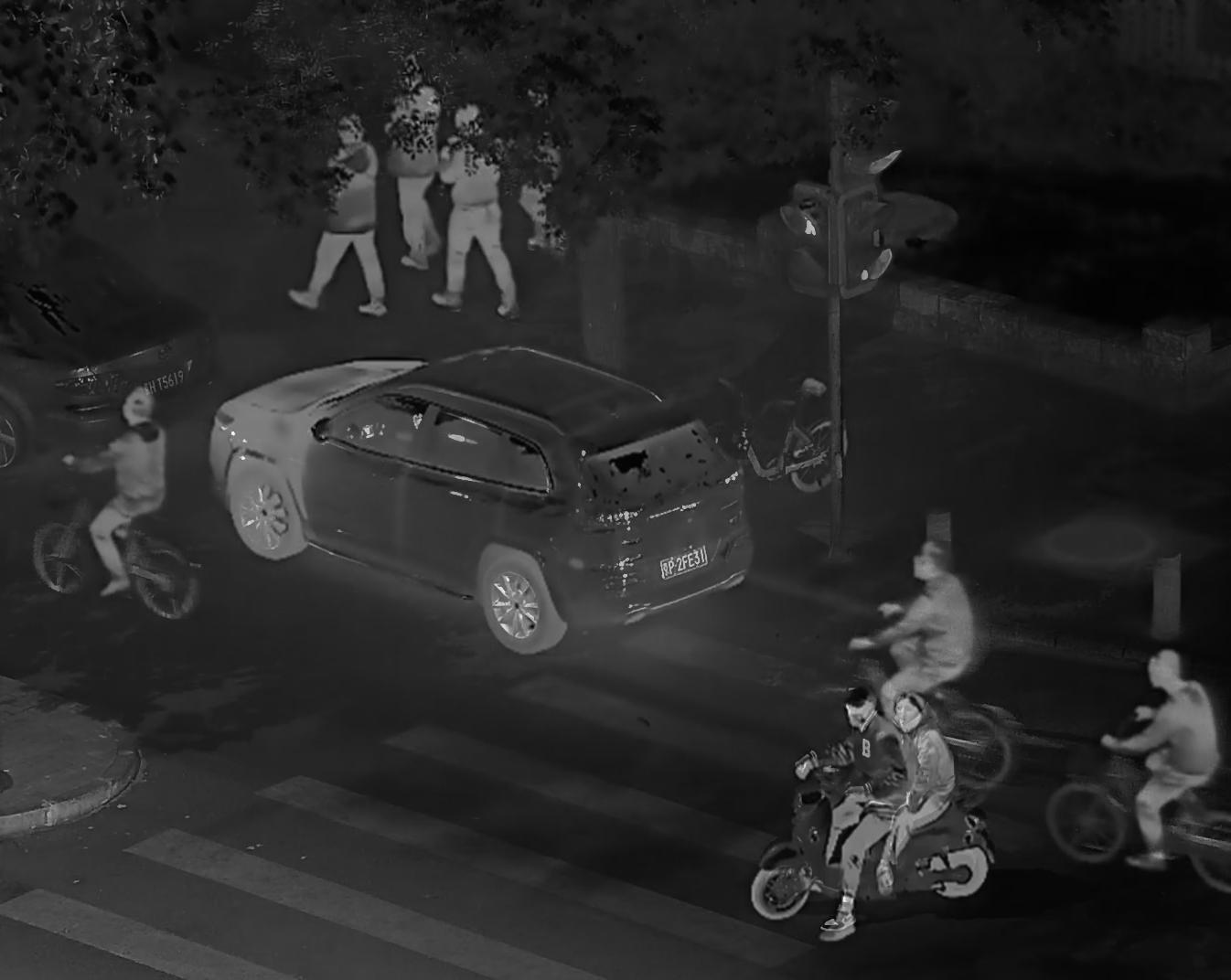}\vspace{2pt}
      \includegraphics[width=\linewidth]{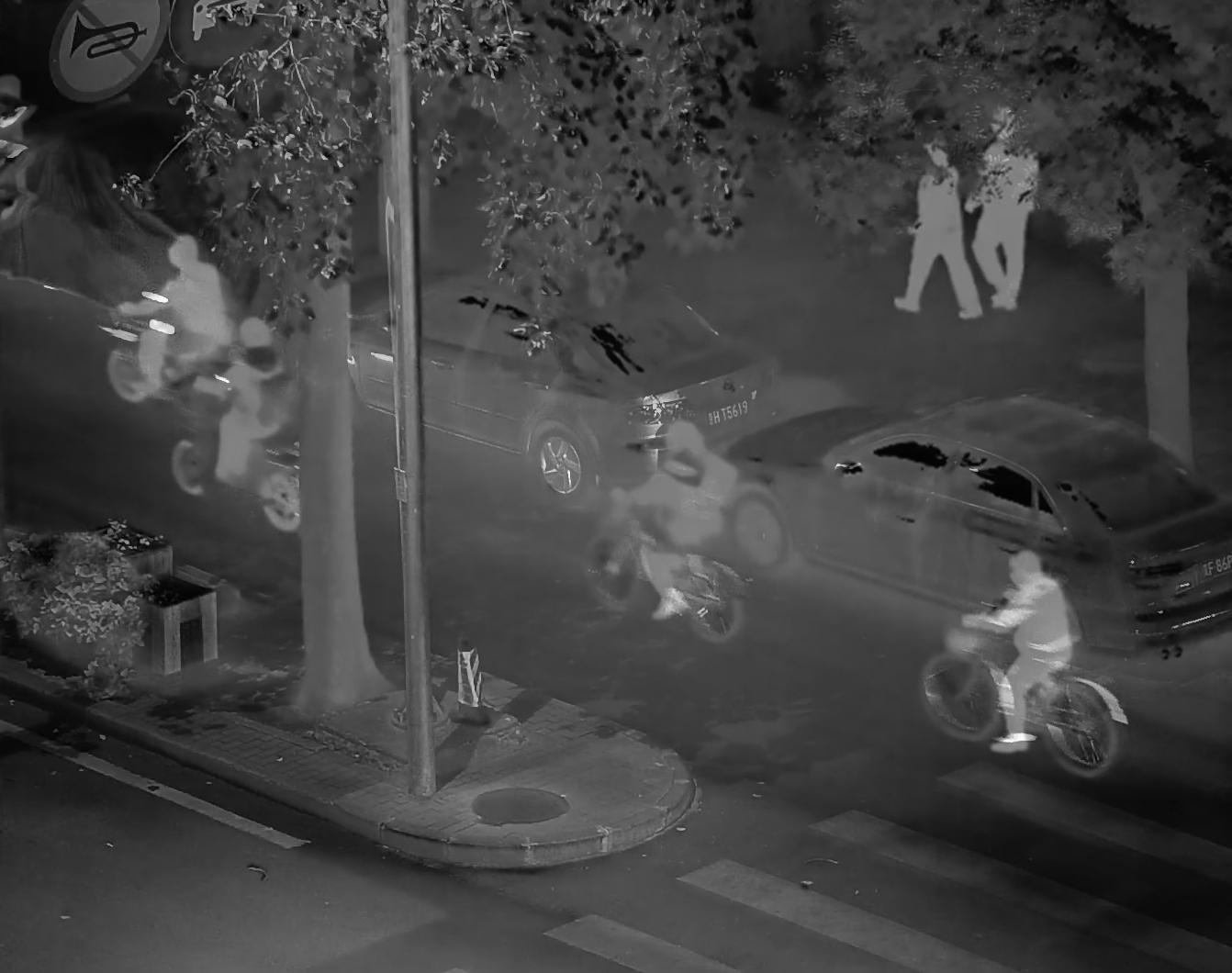}\vspace{2pt}
    \end{minipage}
  }
  \subfigure[densefuse\_$l_1$]{
    \begin{minipage}[b]{0.127\linewidth}  
      \centering
      \includegraphics[width=\linewidth]{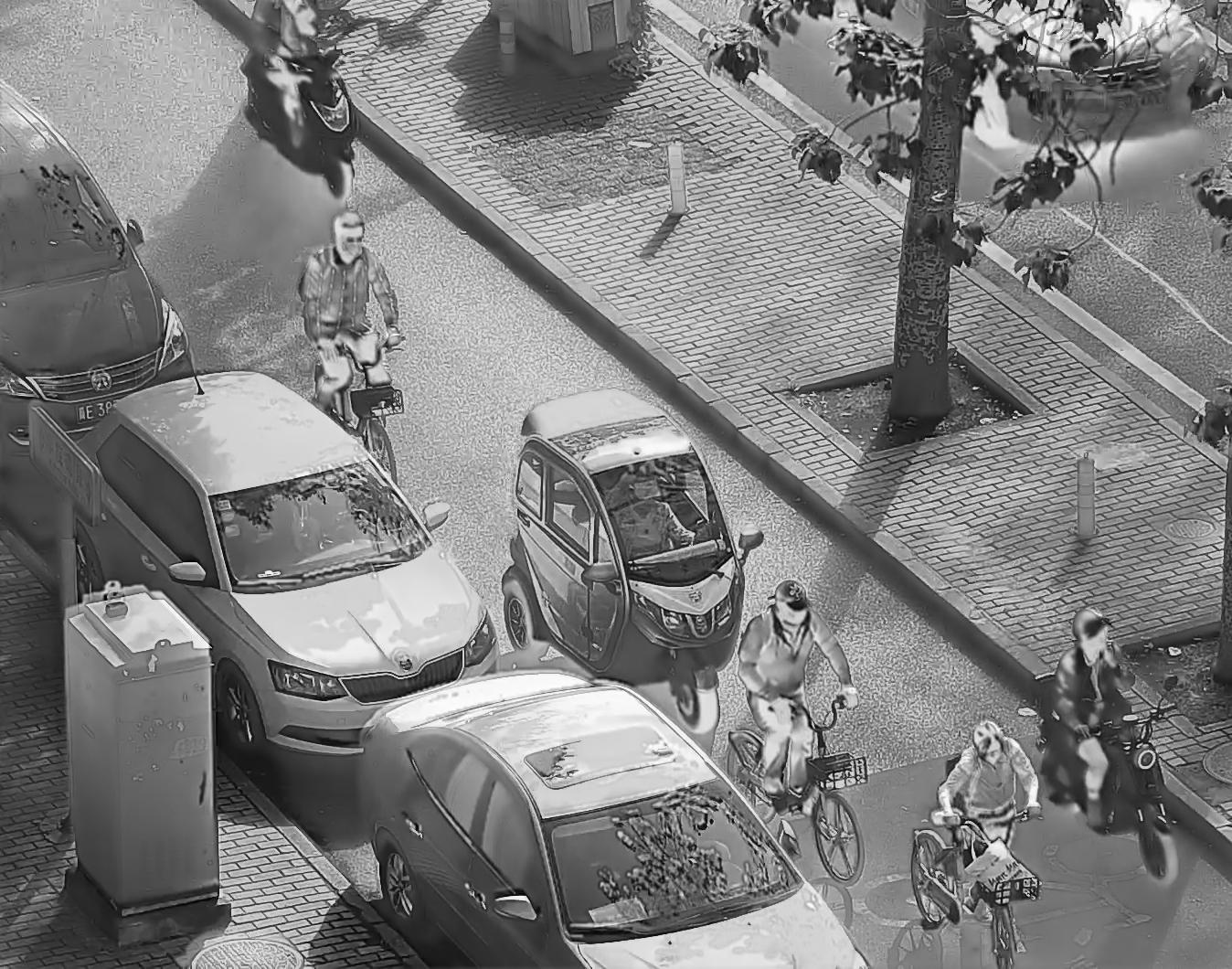}\vspace{2pt}
      \includegraphics[width=\linewidth]{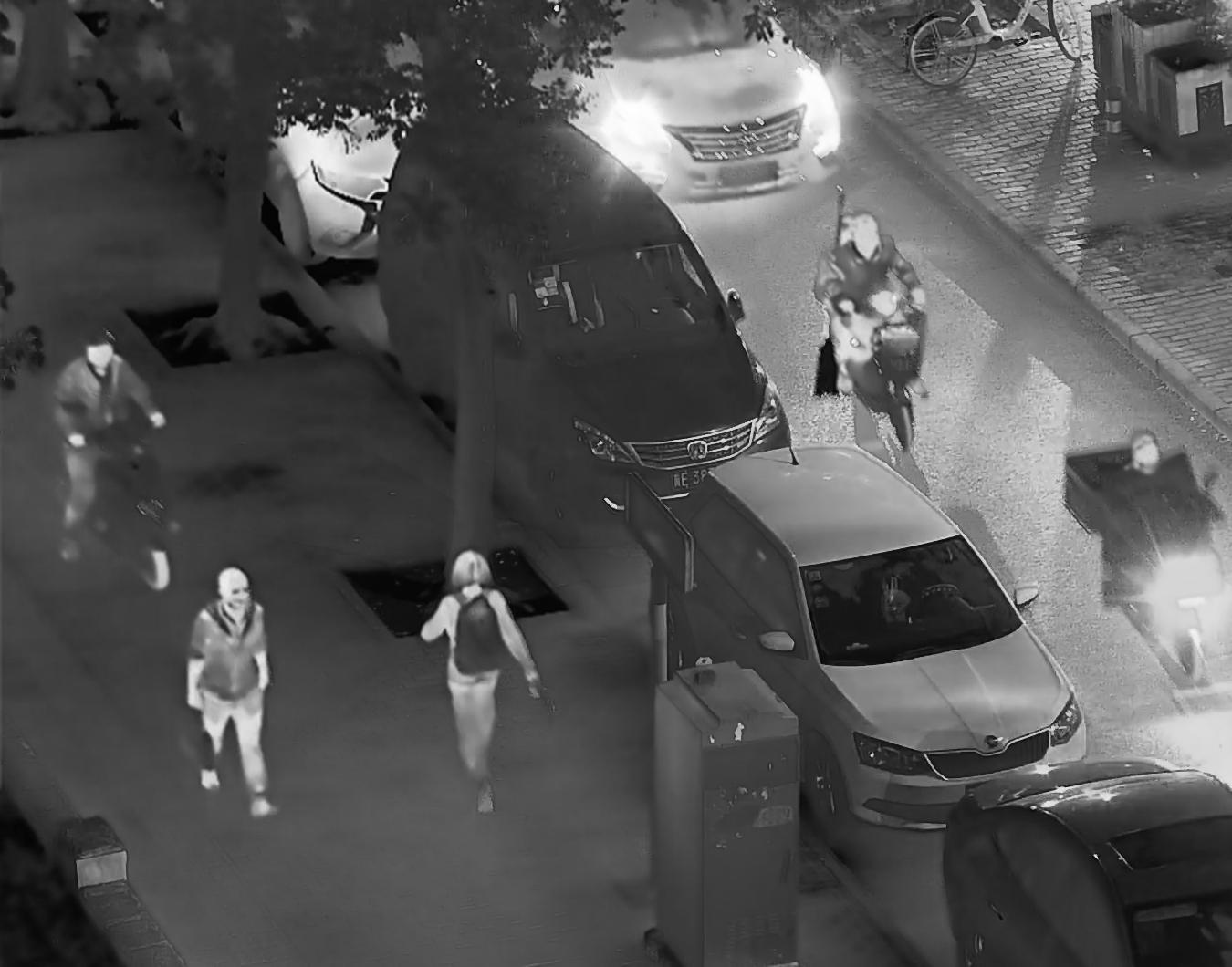}\vspace{2pt}
      \includegraphics[width=\linewidth]{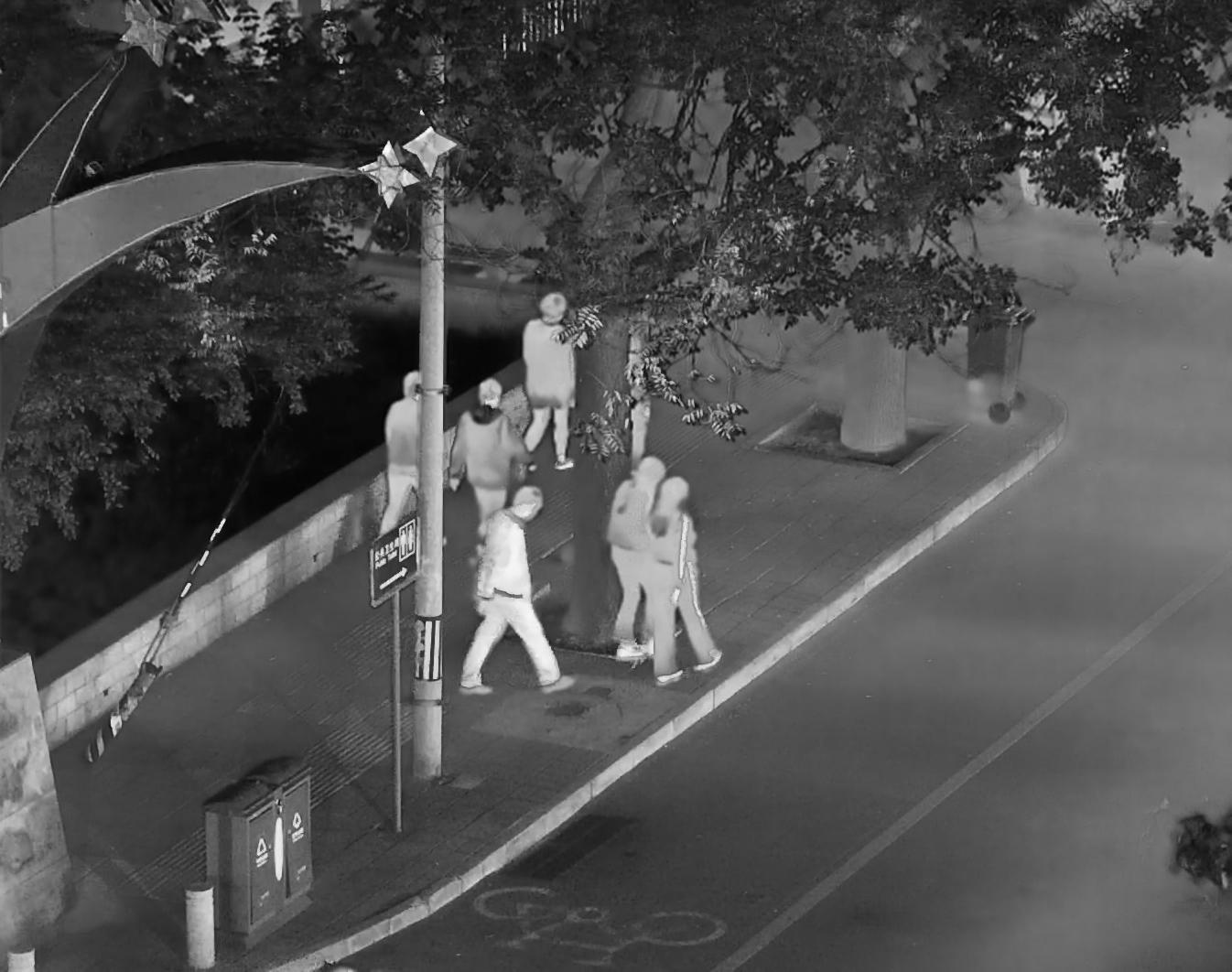}\vspace{2pt}
      \includegraphics[width=\linewidth]{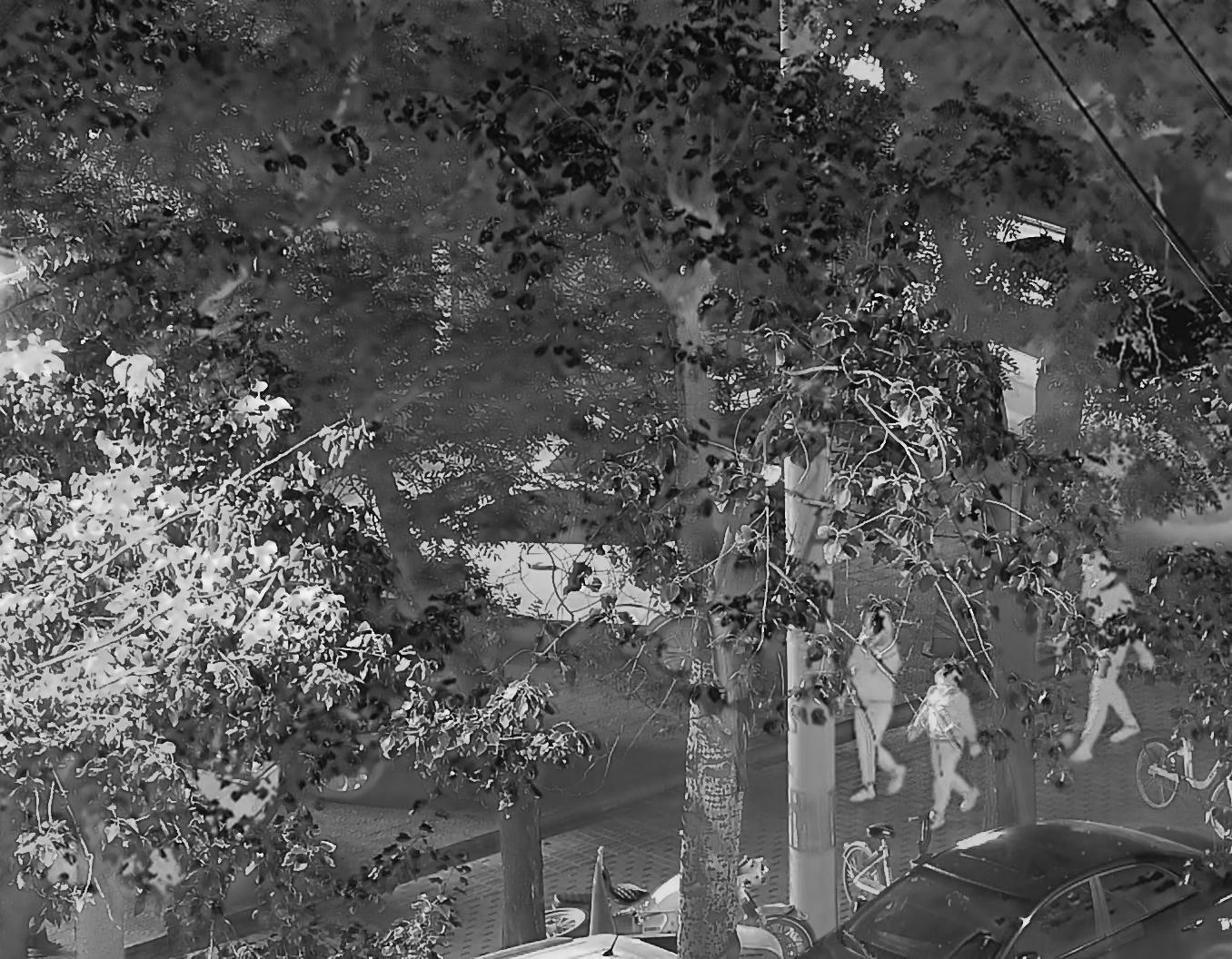}\vspace{2pt}
      \includegraphics[width=\linewidth]{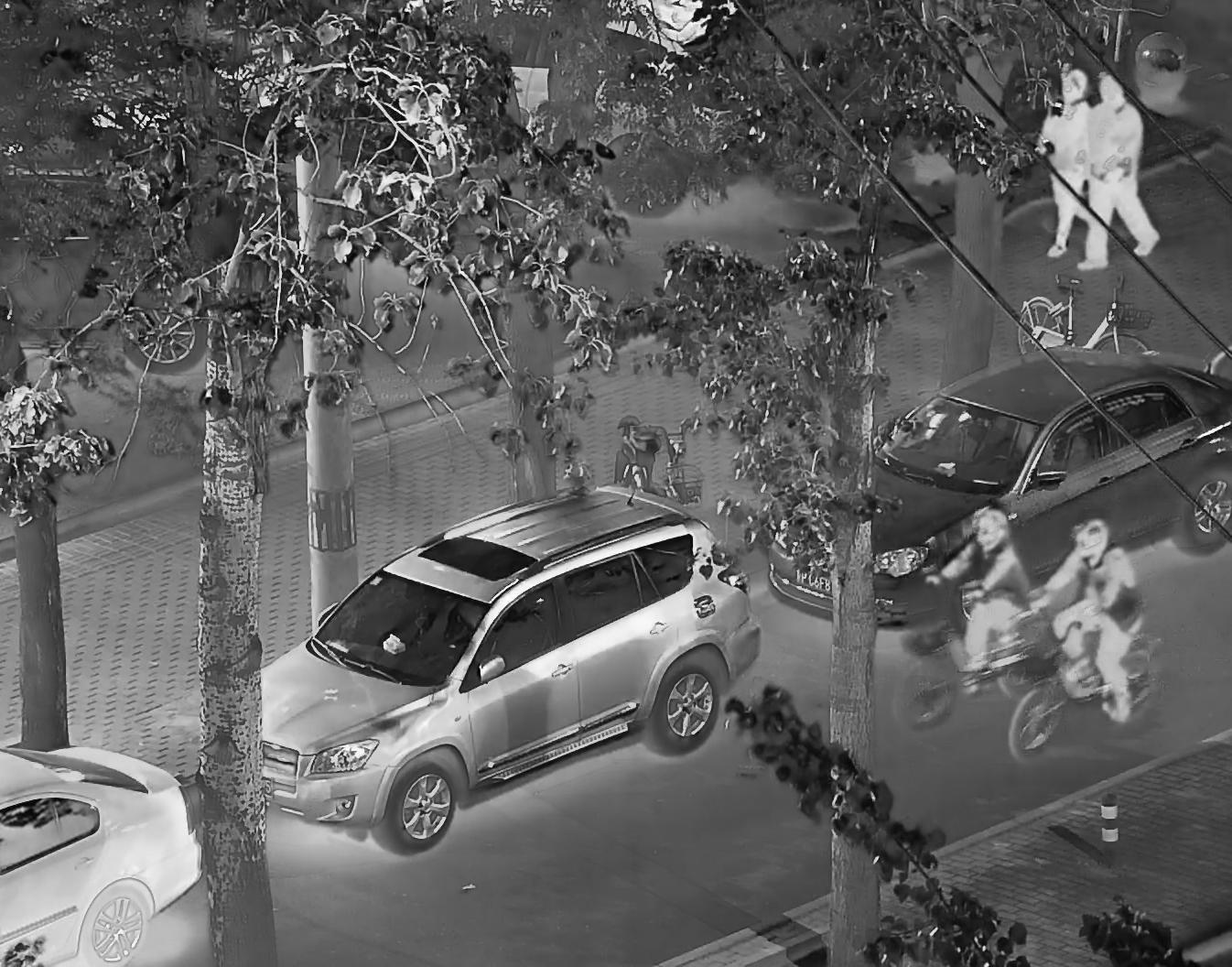}\vspace{2pt}
      \includegraphics[width=\linewidth]{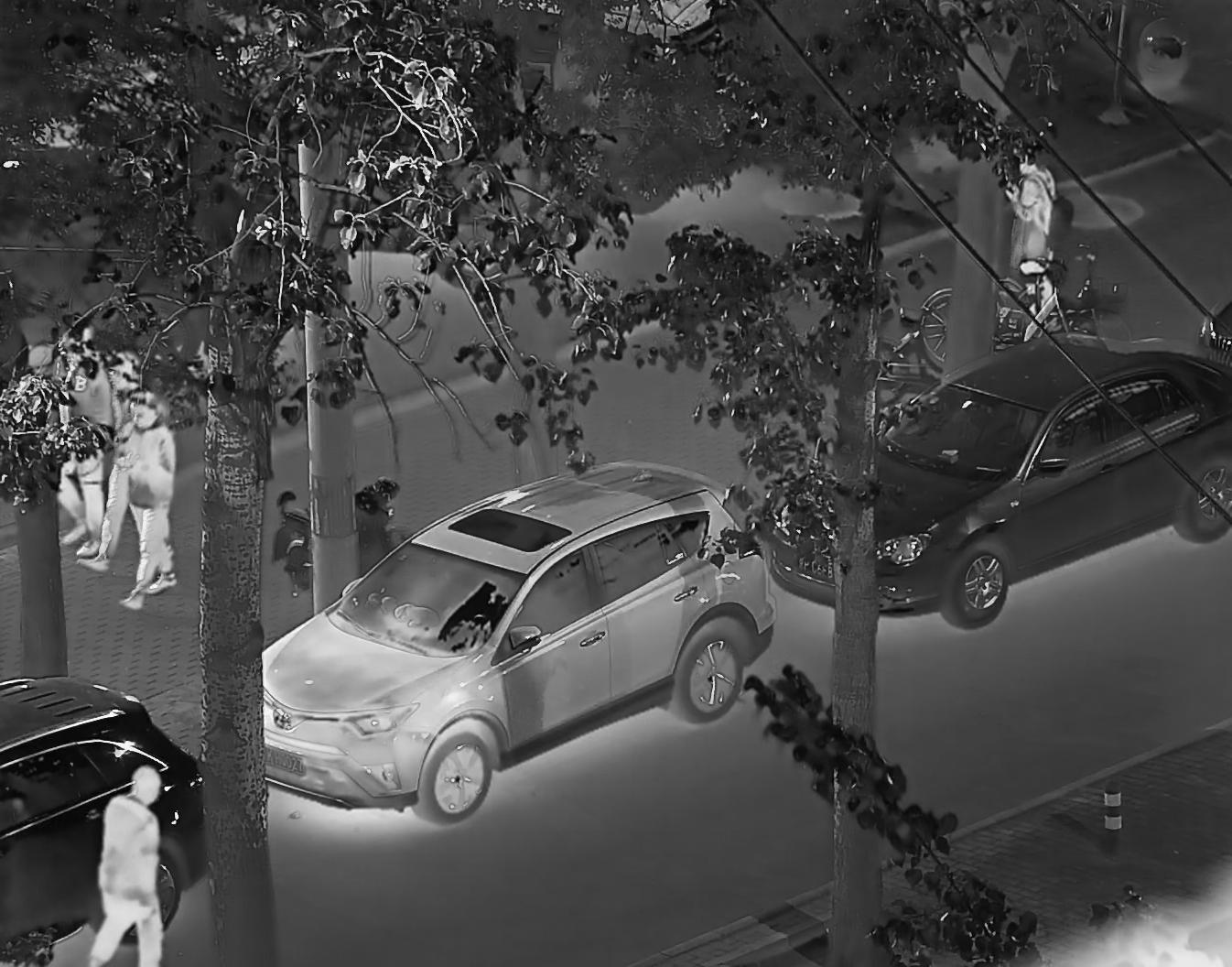}\vspace{2pt}
      \includegraphics[width=\linewidth]{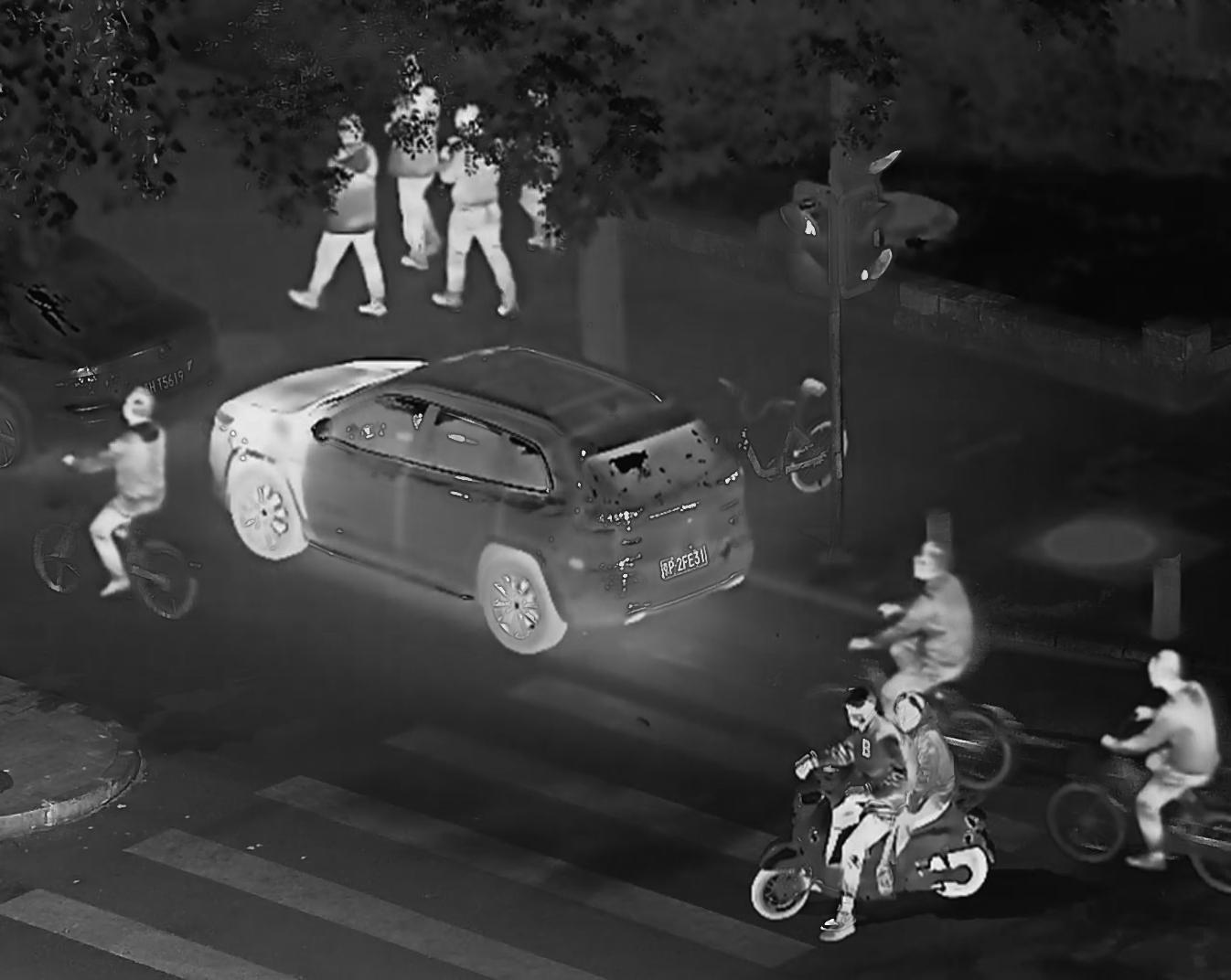}\vspace{2pt}
      \includegraphics[width=\linewidth]{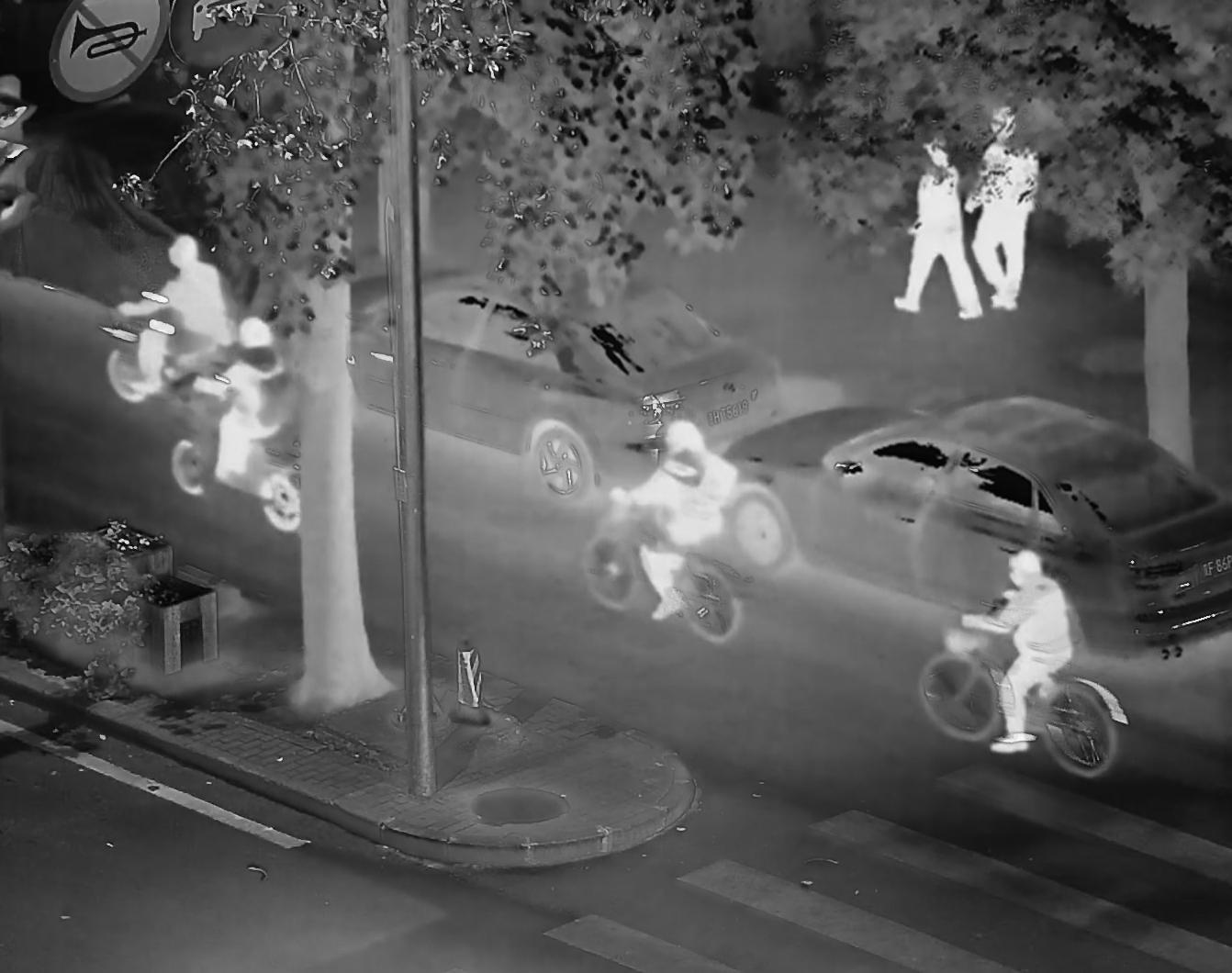}\vspace{2pt}
    \end{minipage}
  }
  \subfigure[FusionGAN]{
    \begin{minipage}[b]{0.127\linewidth}  
      \centering
      \includegraphics[width=\linewidth]{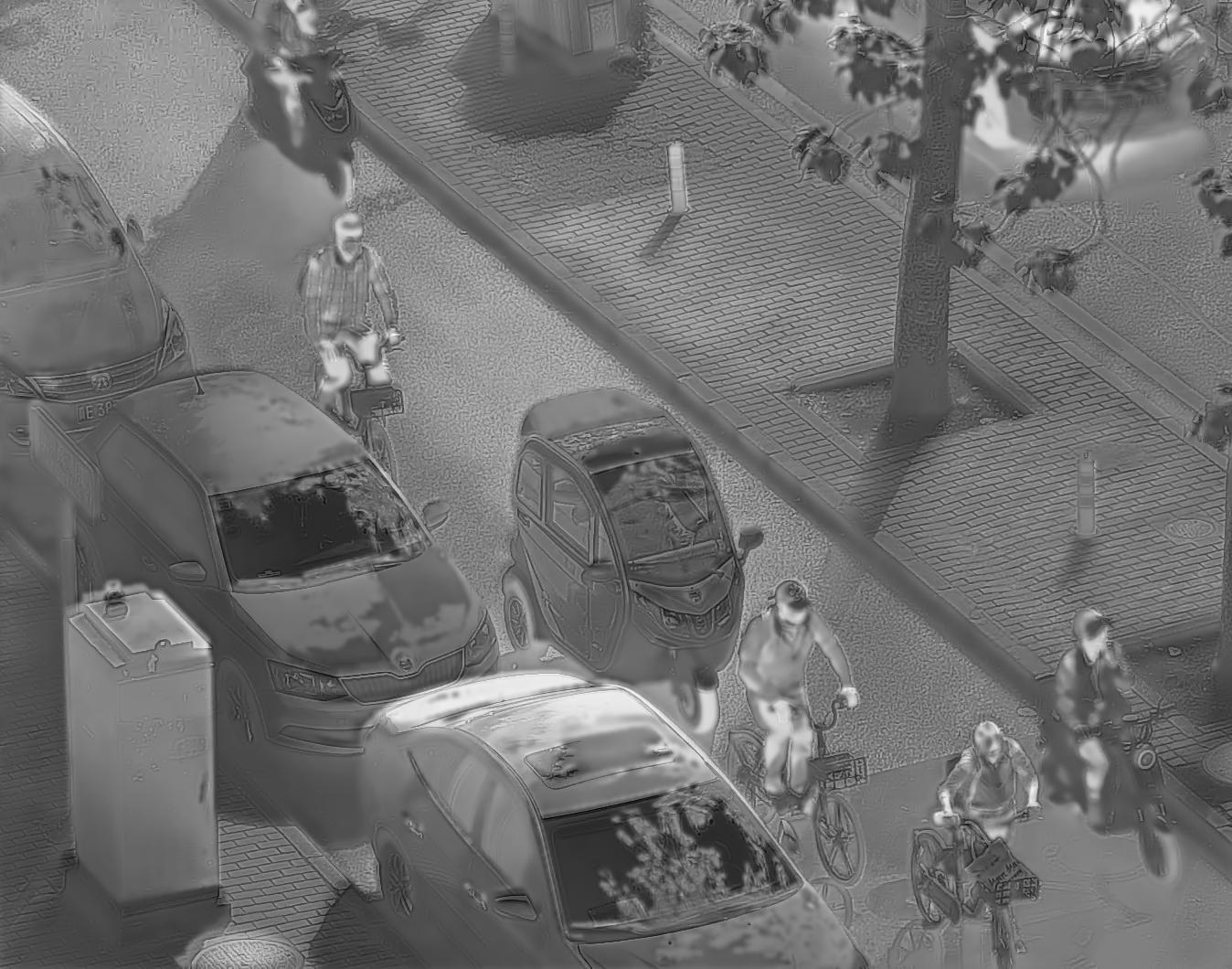}\vspace{2pt}
      \includegraphics[width=\linewidth]{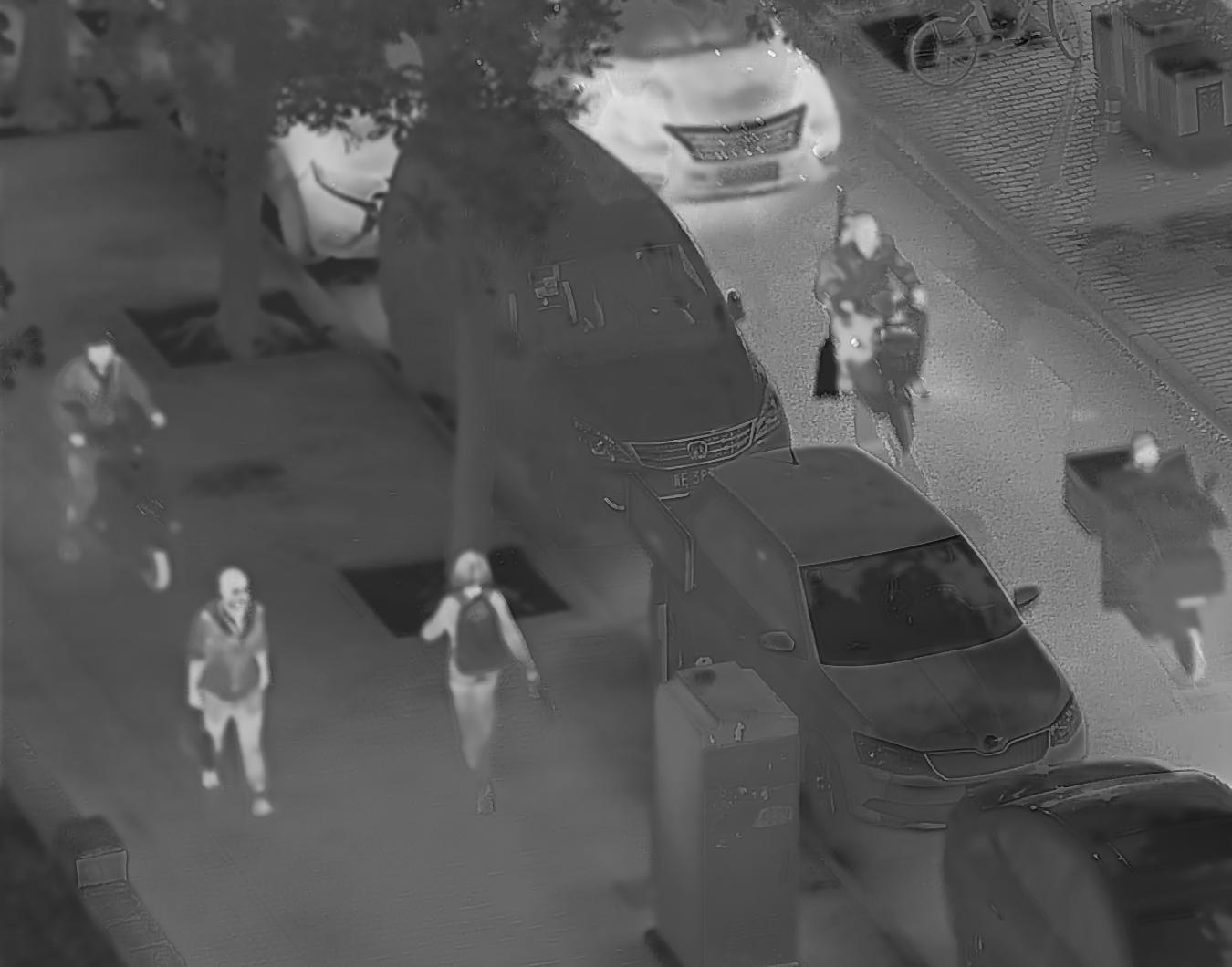}\vspace{2pt}
      \includegraphics[width=\linewidth]{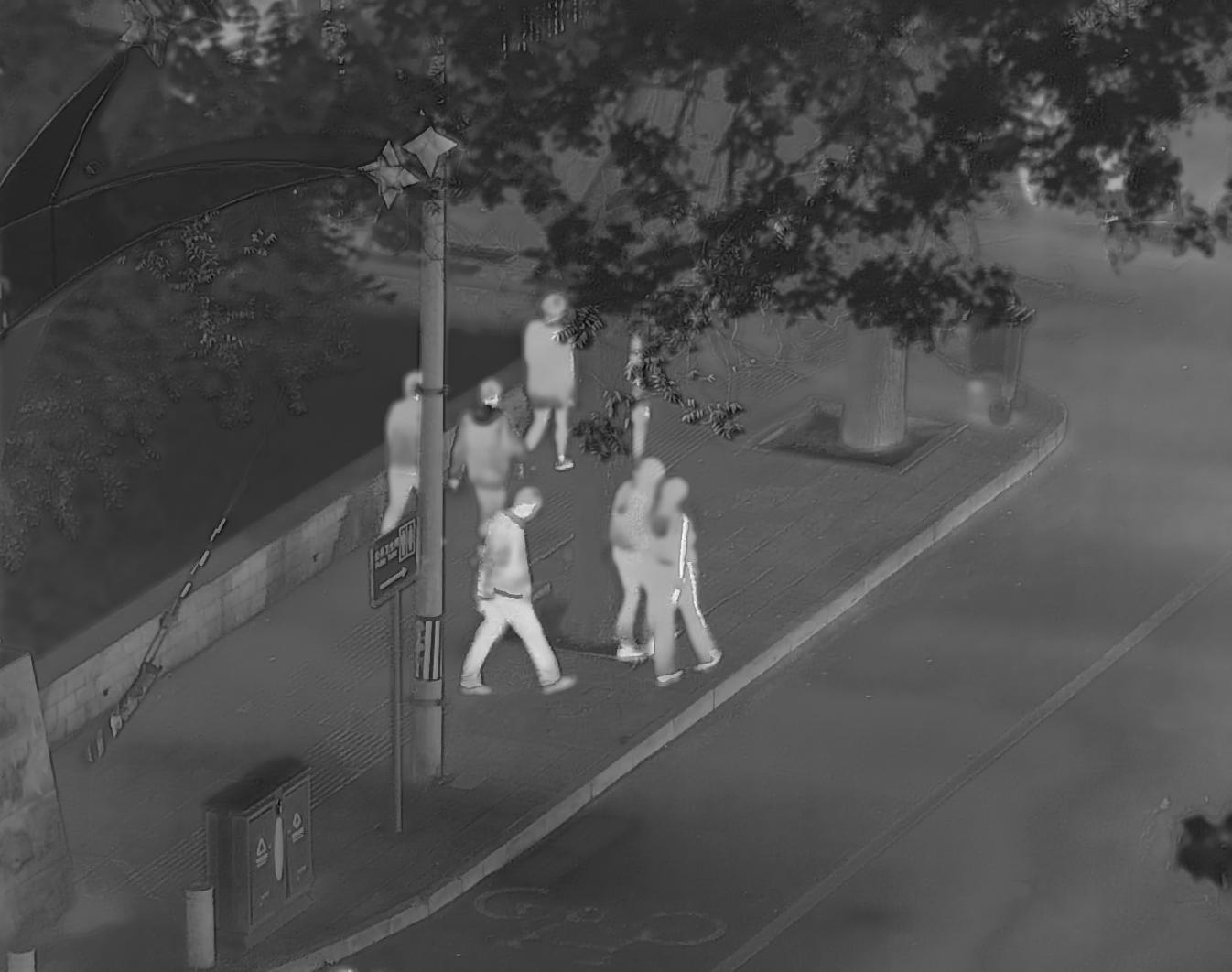}\vspace{2pt}
      \includegraphics[width=\linewidth]{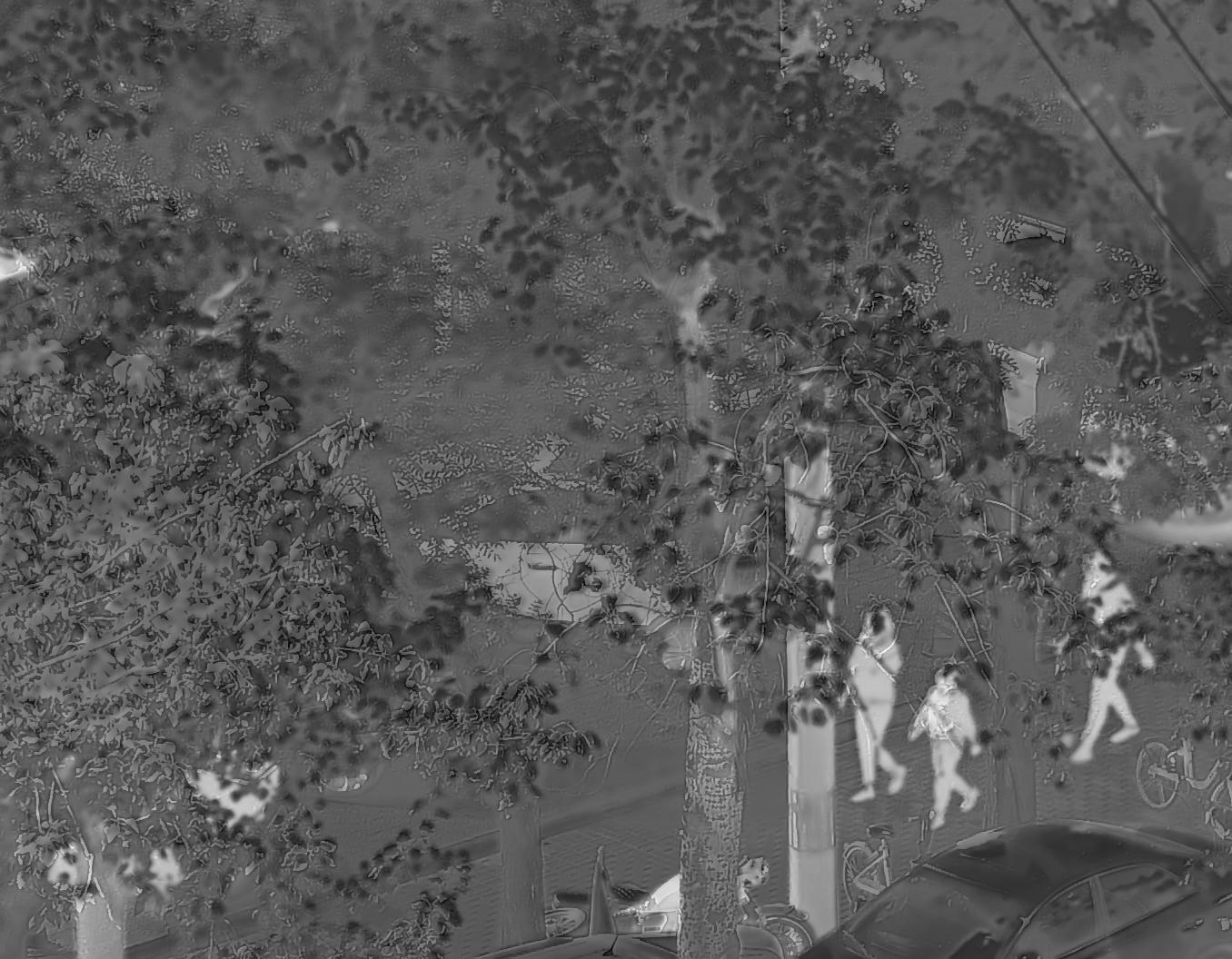}\vspace{2pt}
      \includegraphics[width=\linewidth]{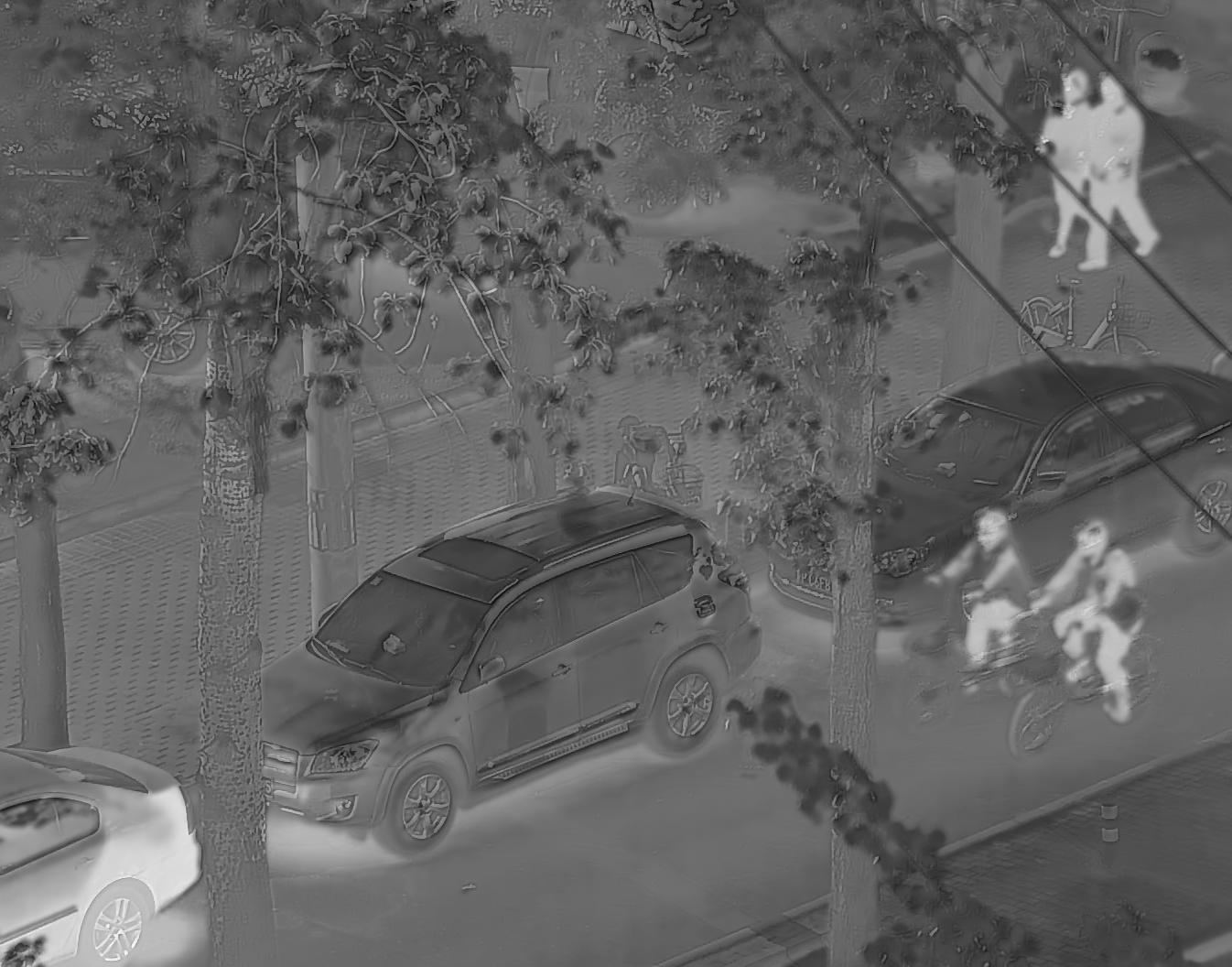}\vspace{2pt}
      \includegraphics[width=\linewidth]{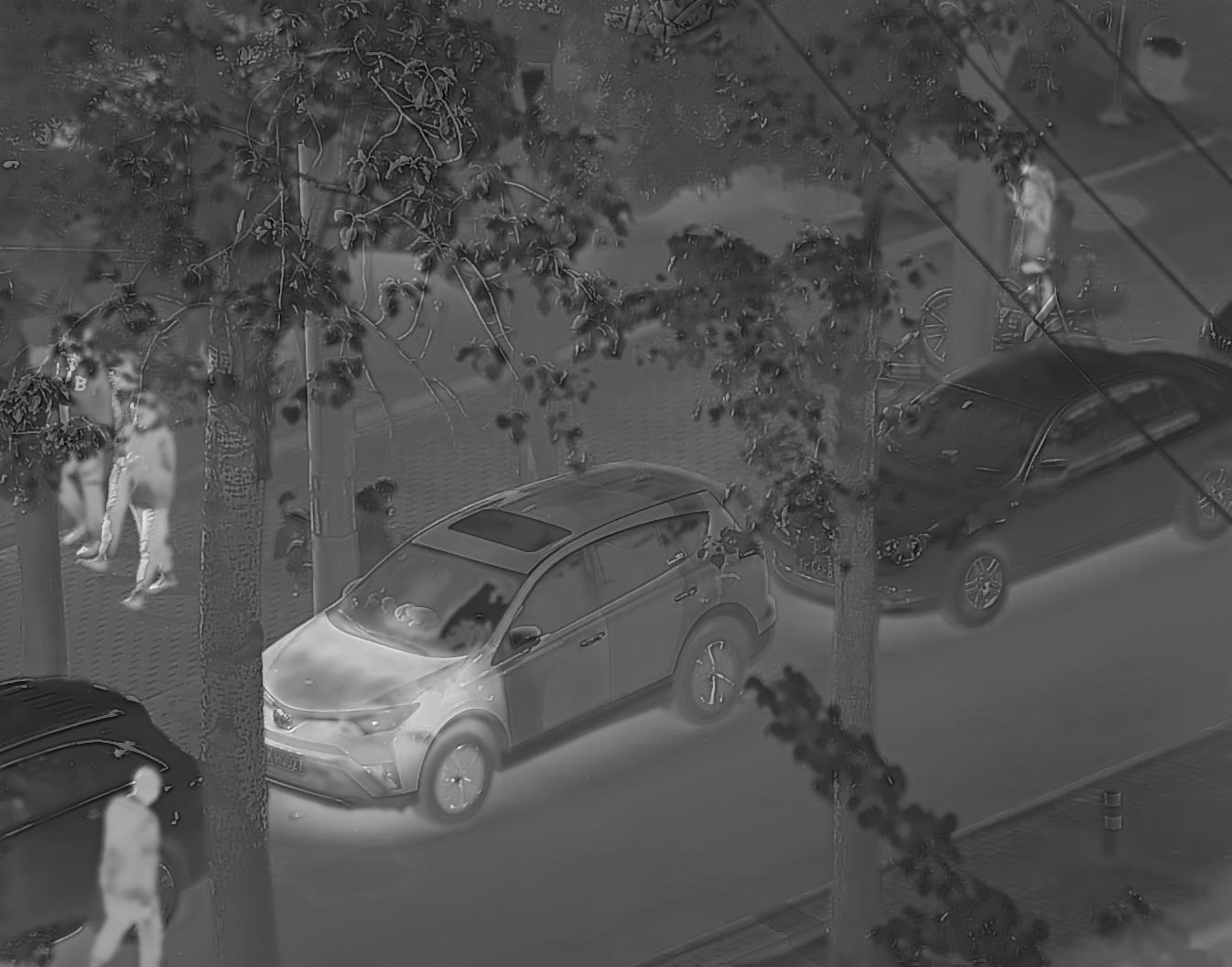}\vspace{2pt}
      \includegraphics[width=\linewidth]{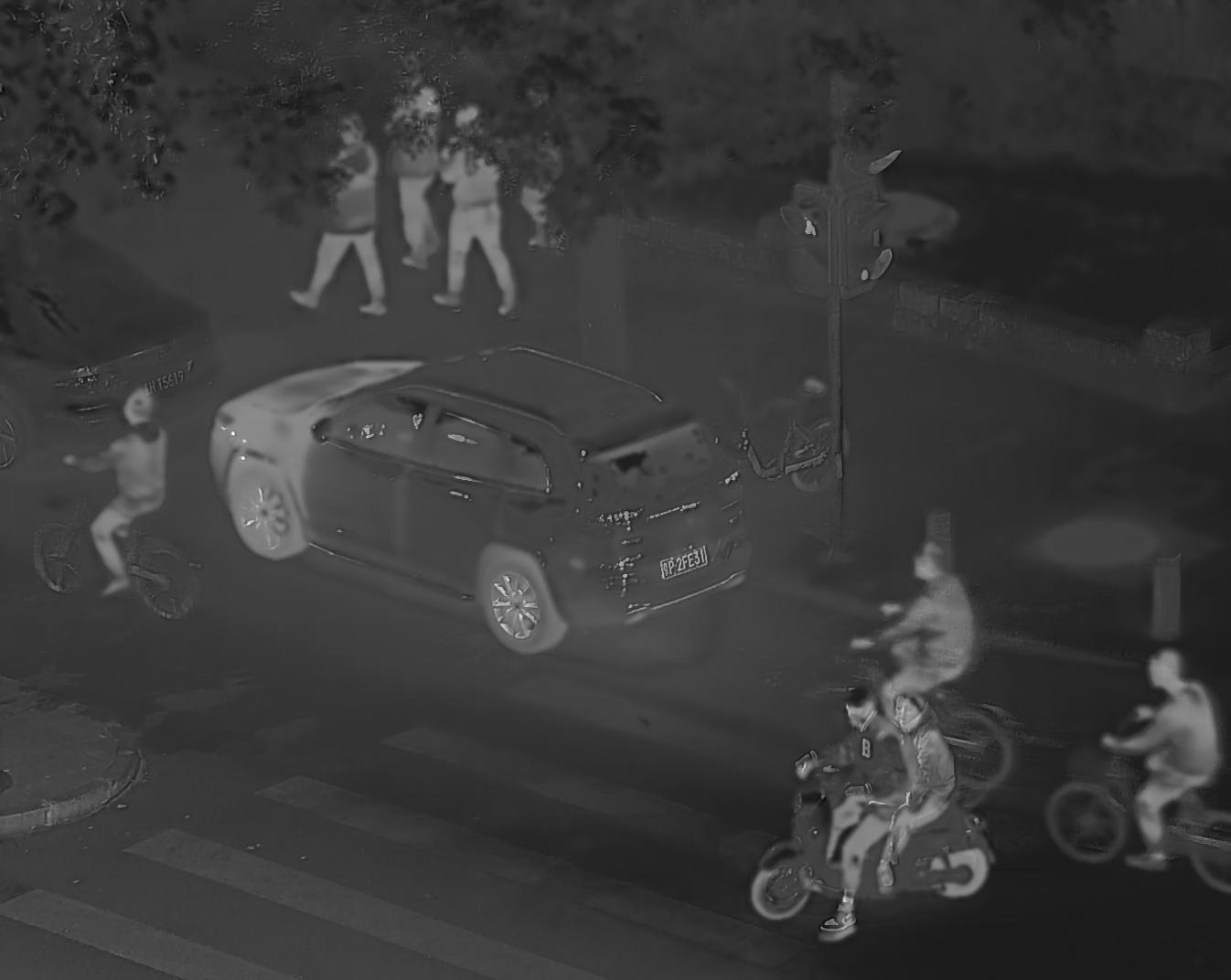}\vspace{2pt}
      \includegraphics[width=\linewidth]{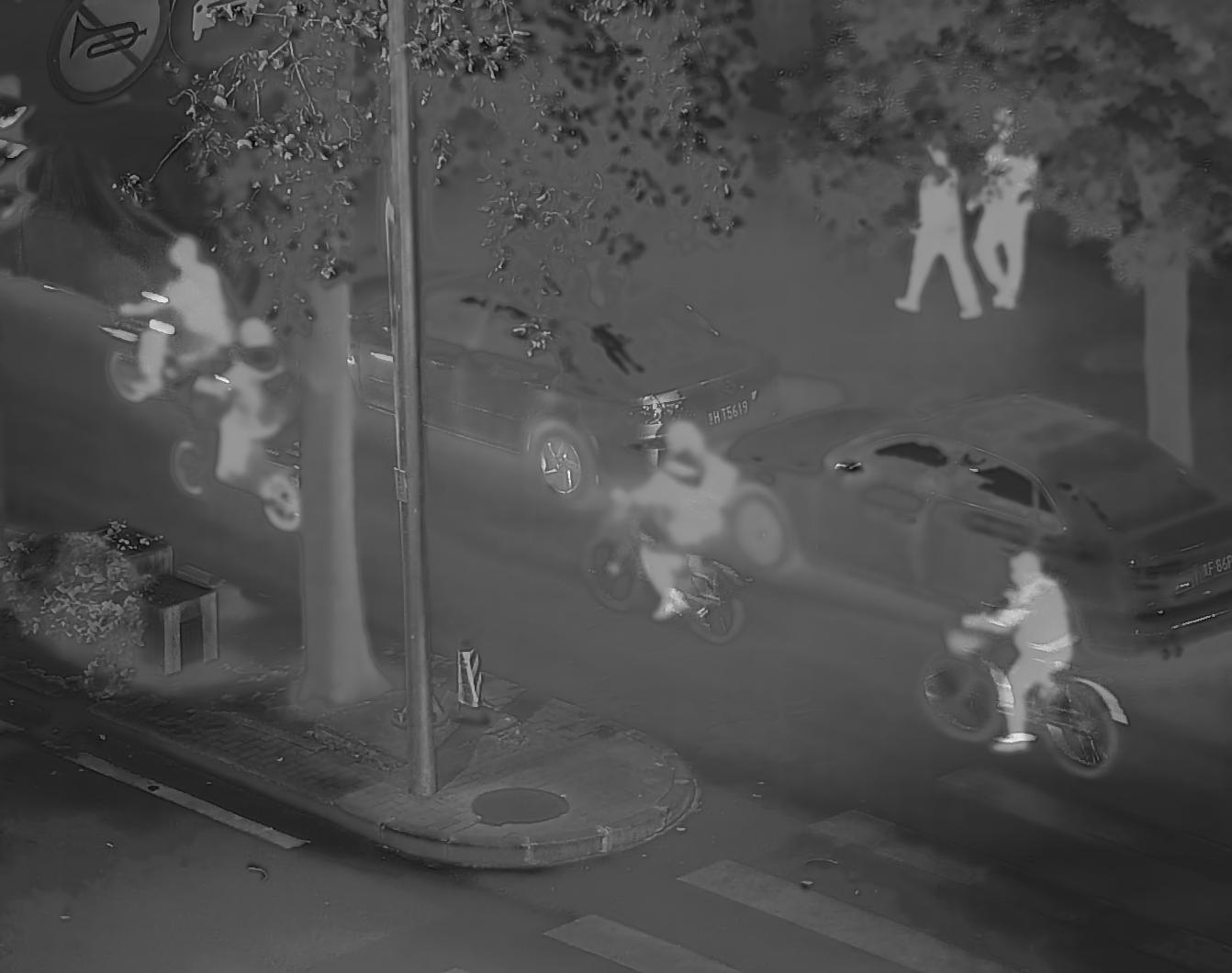}\vspace{2pt}
    \end{minipage}
  }
  \subfigure[IFCNN]{
    \begin{minipage}[b]{0.127\linewidth}  
      \centering
      \includegraphics[width=\linewidth]{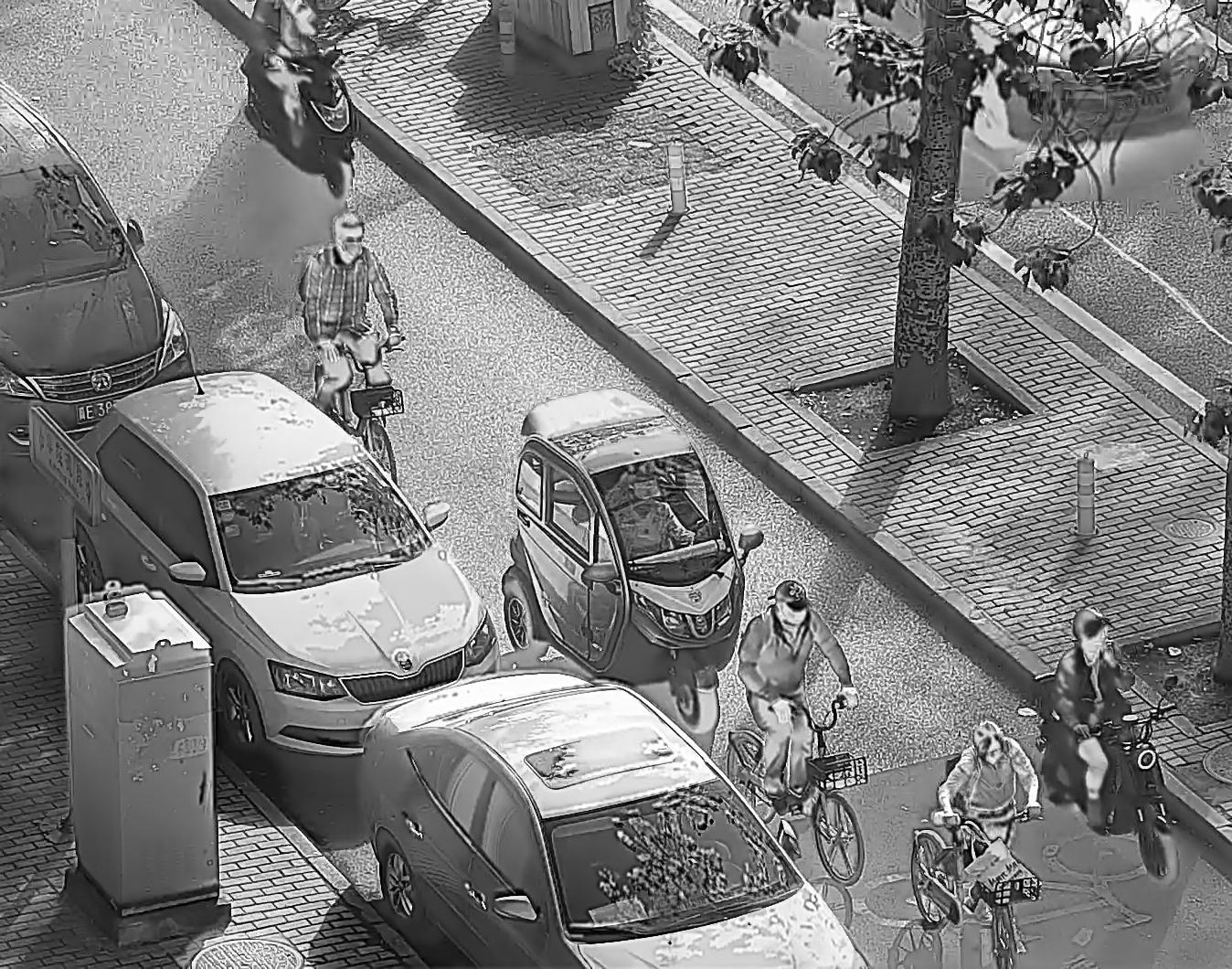}\vspace{2pt}
      \includegraphics[width=\linewidth]{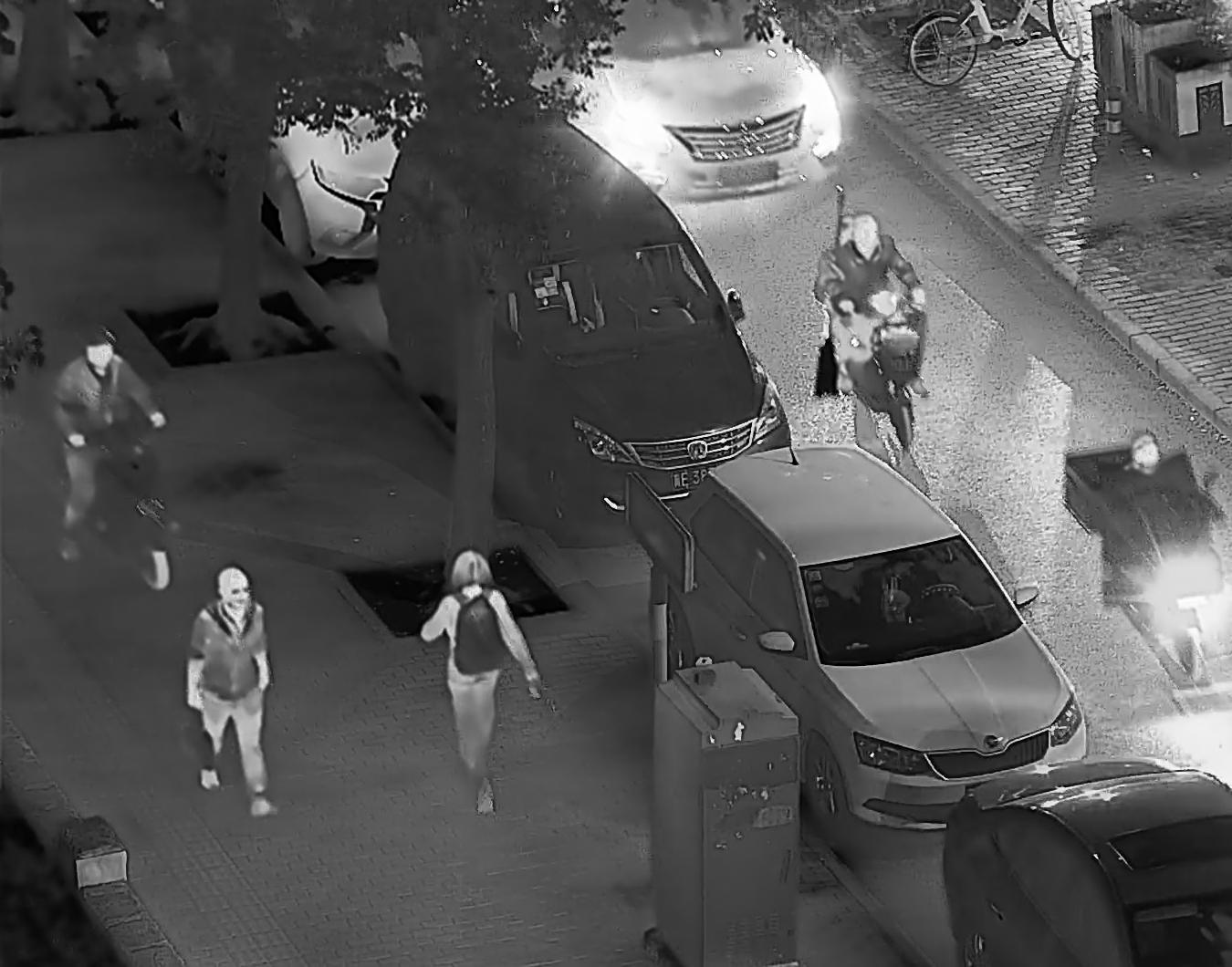}\vspace{2pt}
      \includegraphics[width=\linewidth]{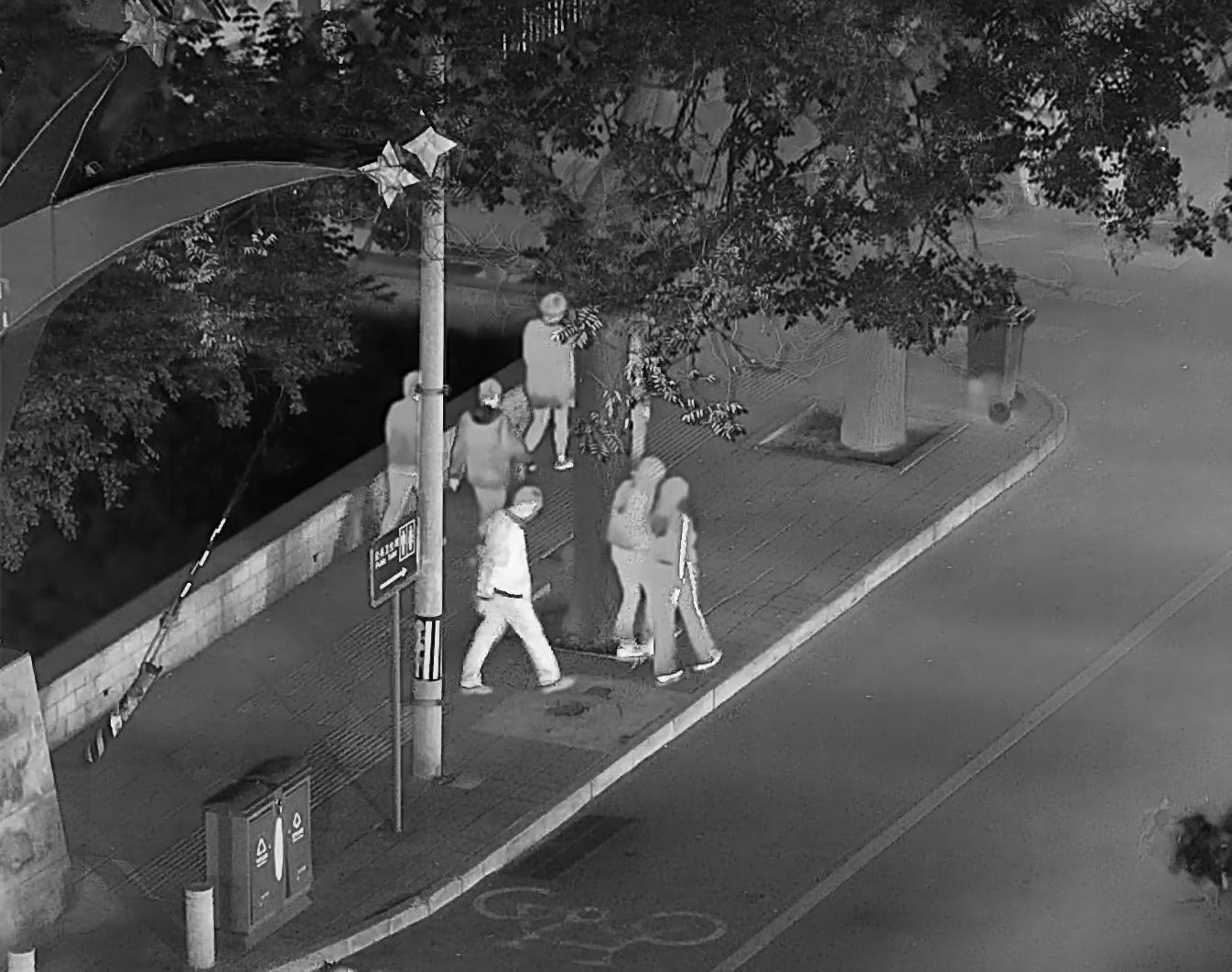}\vspace{2pt}
      \includegraphics[width=\linewidth]{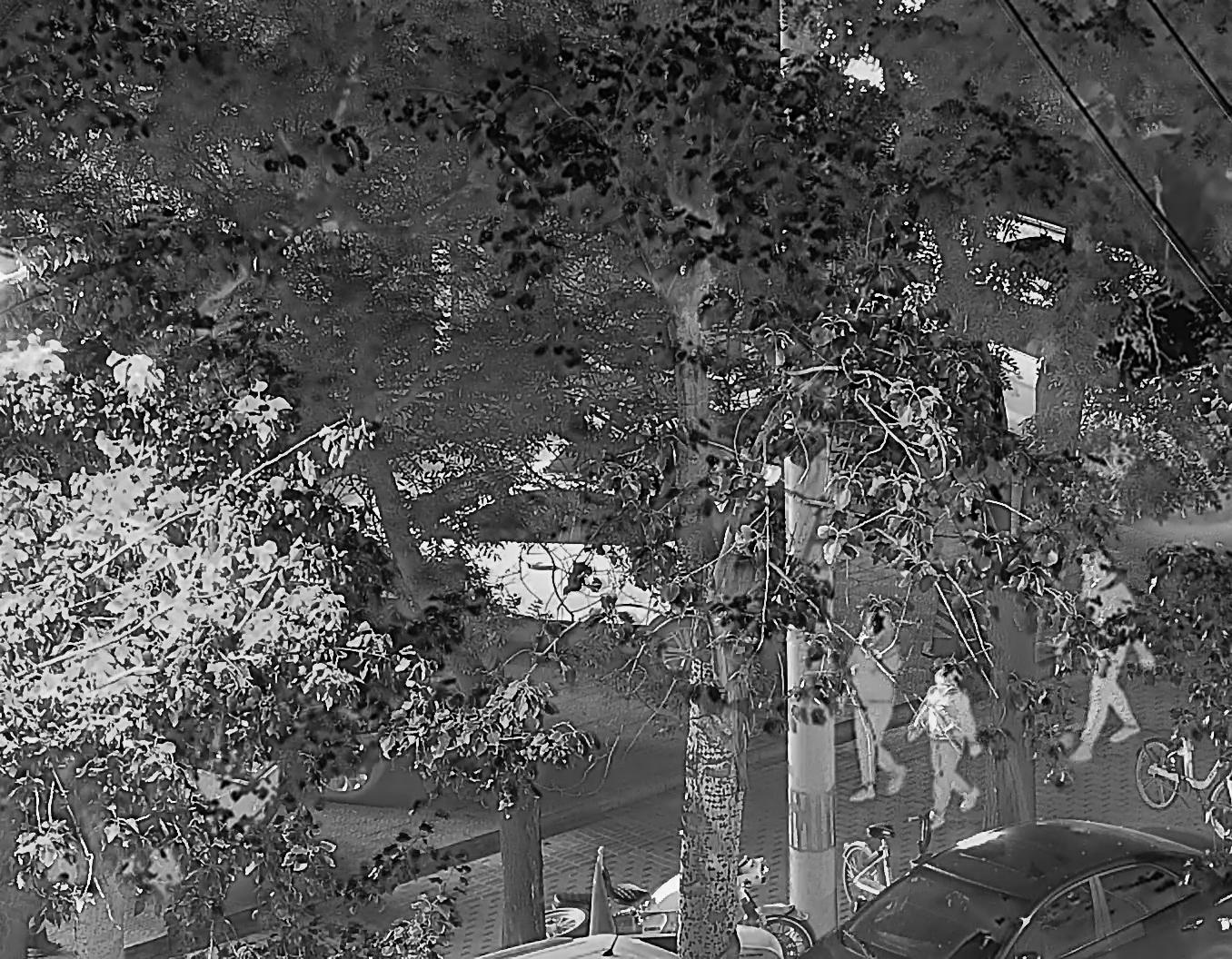}\vspace{2pt}
      \includegraphics[width=\linewidth]{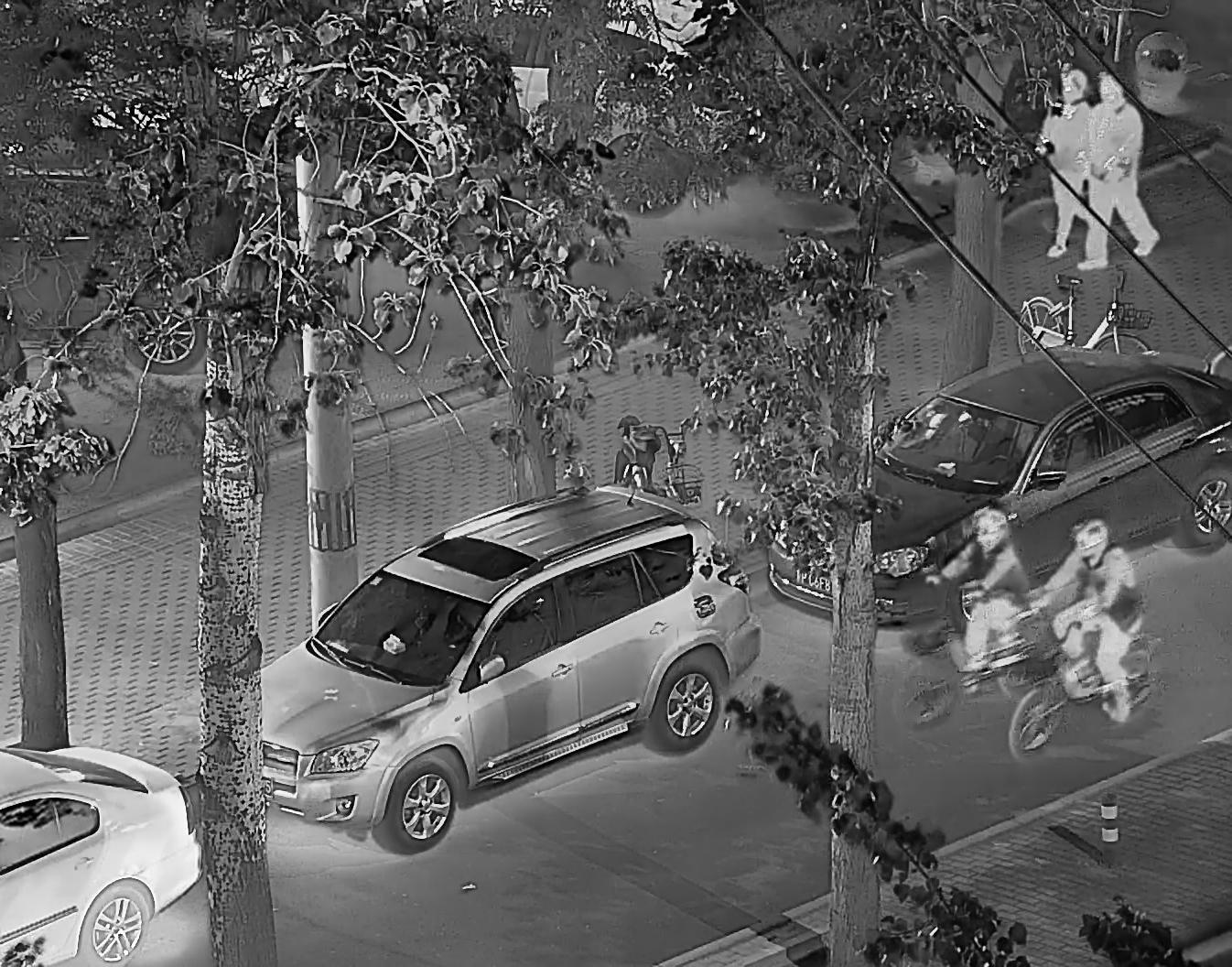}\vspace{2pt}
      \includegraphics[width=\linewidth]{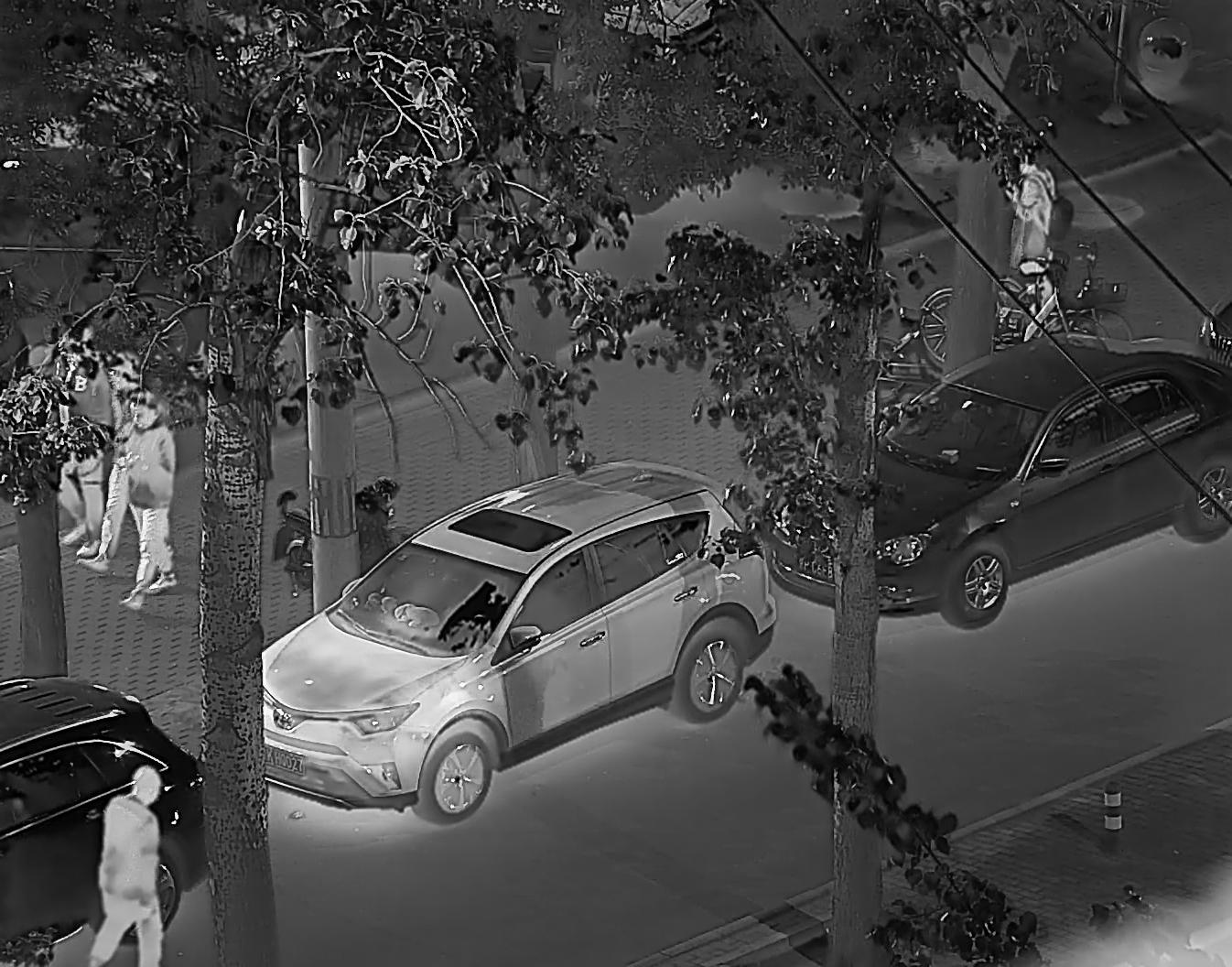}\vspace{2pt}
      \includegraphics[width=\linewidth]{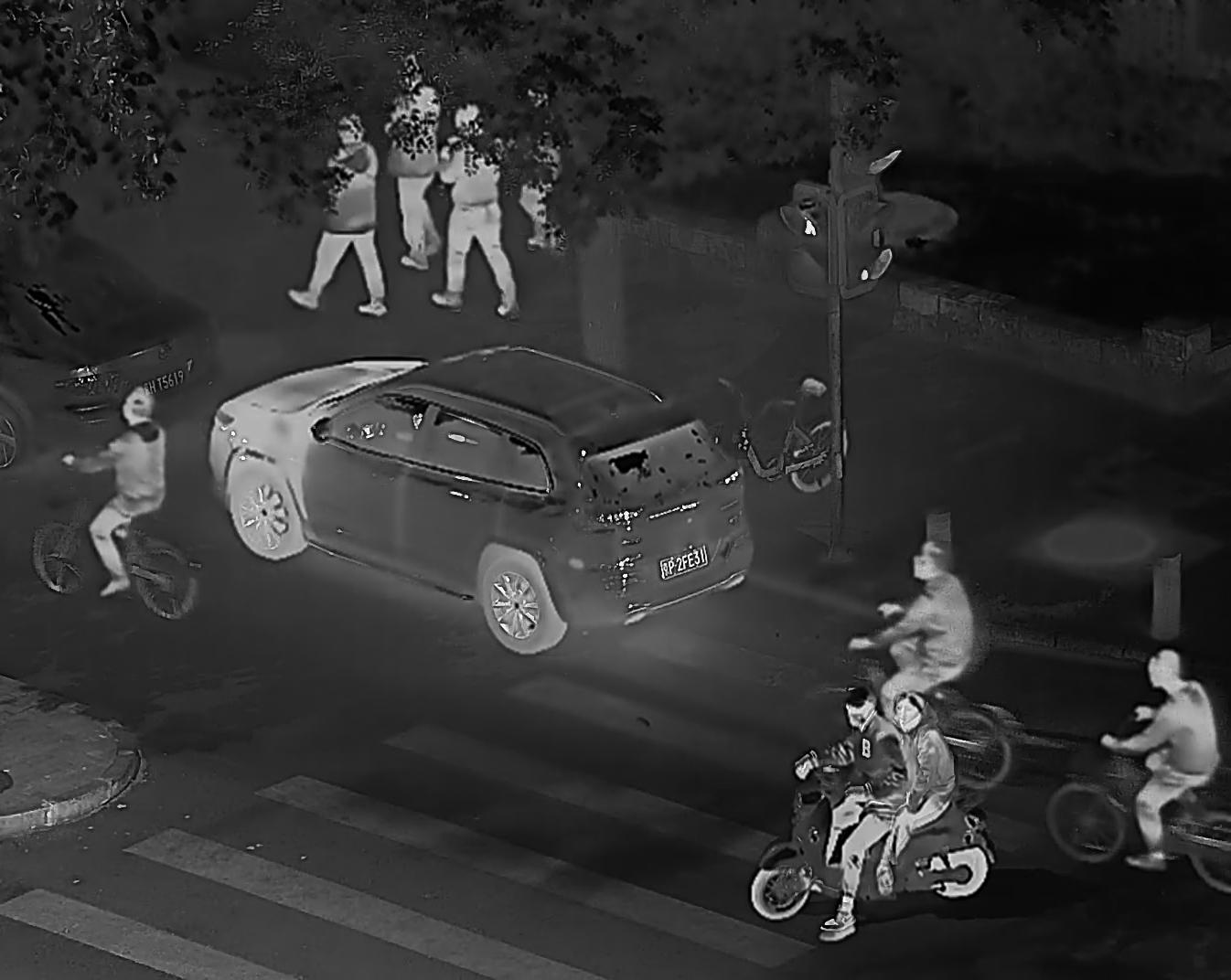}\vspace{2pt}
      \includegraphics[width=\linewidth]{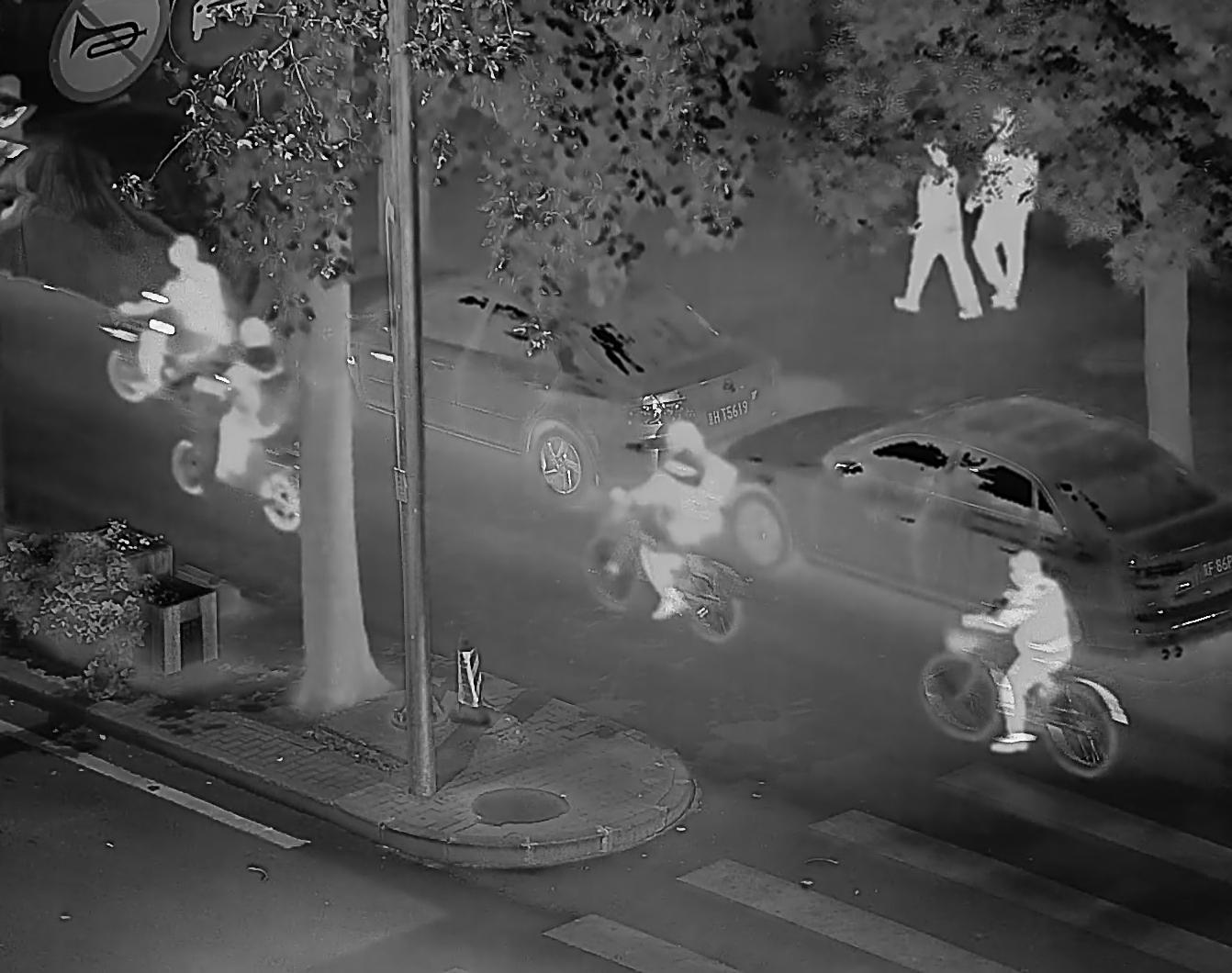}\vspace{2pt}
    \end{minipage}
  }
  \end{minipage}
  \vfill
  \end{center}
  \caption{Examples of fusion results of several fusion algorithms on the LLVIP dataset. From left to right: (a) visible images, (b) infrared images, (c) GTF results, (d) densefuse\_add results, (e) densefuse\_$l_1$ results, (f) FusionGAN results, (g) IFCNN results.}
\label{fig-fused} 
\end{figure*}

\vspace{-1.2em}
\paragraph{Subjective evaluation.} Fig.~\ref{fig-fused} shows some examples of fused images. From the first column on the left, we can clearly see that when the light condition is poor, visible images can hardly distinguish human body and background. In infrared images, objects such as human body can be easily distinguished with clear outline, but there is no internal texture information.Fusion algorithms combine the information of the two kinds of images more or less, so that human bodies are highlighted and the images contain some texture information. 

Judging from the subjective perception of human eyes, we believe that densefuse\_$l_1$ and IFCNN are the most suitable ones for image fusion at night. Because the fused images obtained by these two methods retain more information from visible and infrared images, \emph{i.e.}, they are not only more detailed, but also highlight the human body.

\begin{figure}[htb]
\begin{center}
\includegraphics[width=0.98\linewidth]{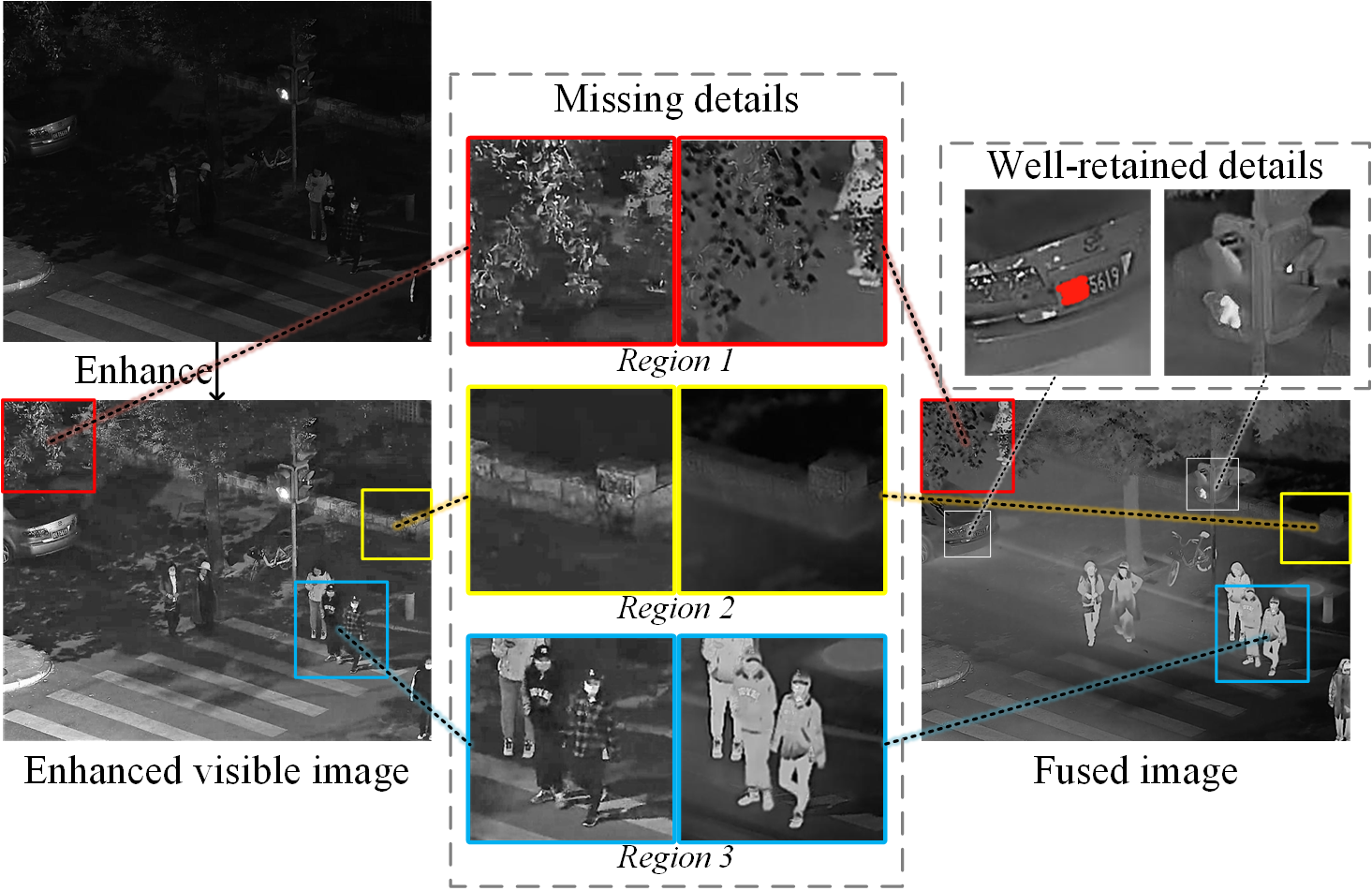}
\end{center}
\caption{Comparison of the enhanced visible image and the fused image in detail. Details that are bright in the original visible image are well preserved in the fused image(we daub the license plate numbers), but many other details are lost.} 
\label{fig-comparision} 
\end{figure}

In order to get a clearer view of the details of the visible image retained in fused image, we enhance the low-light visible image. We compare the details in the fused image and the enhanced visible image in Fig.~\ref{fig-comparision}. Details that are bright in the original visible image are well retained in the fused image, such as the license plate number and the traffic light. However, we notice that there are some missing details in the fused image.

On the one hand, the dark details in the original visible image are badly lost in fused image, \eg, the \textit{region 1} and \textit{region 2} of ``missing details" in Fig.~\ref{fig-comparision}. The enhanced image demonstrates that these low-light areas contain a lot of detail, but they are not contained in the fused image, the textures of the leaves and stones are all lost in the fusion image.

On the other hand, a lot of details in people are lost, \eg, the \textit{region 3} of ``bad details" in Fig.~\ref{fig-comparision}. The texture information of people's clothes is not shown in the fusion image, which is not only because of the poor illumination of the visible image, but also because the infrared image dominates the fusion image due to the high pixel intensity here.

In general, when the pixel intensity of one image in the source images is very low, or the pixel intensity of one image is very high, the fusion effect will be worse. In other words, the ability of fusion algorithm to balance two source images is poor. This demonstrates that the existing fusion algorithms still have great room for improvement.

\paragraph{Objective evaluation.} We also provide the average value of six metrics of different fusion algorithms on our LLVIP dataset in Table~\ref{table-average}. In general, densefuse\_$l_1$ and IFCNN perform best on the dataset, but they still have a lot of room for improvement.

\begin{table}[htb]
\begin{center}
\resizebox{0.95\linewidth}{!}{
\begin{tabular}{ccccccc}
\hline
               & EN   & FF    & $Q_{MI}$  & $Q_{abf}$ & VIFF & SSIM \\ \hline
GTF            & 6.36 & 10.81 & 1.65    & 0.23    & 0.20 & 0.63 \\ 
densefuse\_add & 7.02 & 11.11 & 1.61    & 0.47    & 0.45 & \textbf{0.68} \\ 
densefuse\_$l_1$  & \textbf{7.26} & \textbf{11.98} & \textbf{1.70}    & 0.54    & 0.46 & \textbf{0.68} \\ 
FusionGAN      & 6.44 & 10.82 & 1.63    & 0.23    & 0.20 & 0.60 \\ 
IFCNN          & 7.22 & 11.39  & 1.62    & \textbf{0.65}    & \textbf{0.57} & 0.67 \\ \hline
\end{tabular}}
\end{center}
    \caption{Average value of EN, FF, $Q_{MI}$, $Q_{abf}$, VIFF, SSIM.} 
\label{table-average}
\end{table}

\subsection{Pedestrian Detection}

For contrast, we use visible image and infrared image respectively for pedestrian detection experiments.

Yolov5~\cite{glenn_jocher_2020_3983579} is tested on the dataset. The model was first pre-trained on the COCO dataset, and then fine-tuned on our dataset. Pretrained checkpoint yolov5l is selected. 77.6\% of the dataset for training and 22.4\% for testing. The models are trained with 200 epochs, batch-size 8, during which the learning rate decreased from 0.0032 to 0.000384. We use SGD with a momentum of 0.843 and a weight decay of 0.00036. Yolov3~\cite{redmon2018yolov3} is also tested on the dataset, and the experimental Settings are consistent with the default.

\begin{table}[htb]
\begin{center}
\resizebox{0.98\linewidth}{!}{
\begin{tabular}{ccccccc}
\hline
         & \multicolumn{3}{c}{Yolov5} & \multicolumn{3}{c}{Yolov3} \\ 
         & AP50 & AP75 &  AP & AP50 & AP75 &  AP \\ \hline 
visible  & 0.908 & 0.564 &  0.527  & 0.871 & 0.455 & 0.466     \\ 
infrared & 0.965 & 0.764 &  0.670  & 0.940 & 0.661 & 0.582     \\ \hline
\end{tabular}}
\end{center}
\caption{Experiment results of pedestrian detection. AP50 means the AP at IoU threshold of 0.5, AP75 means the AP at IoU threshold of 0.75, and AP means the average of AP at IoU threshold of 0.5 to 0.95, with an interval of 0.05.} 
\label{table-mAP}
\end{table}

\begin{table}[htb]
\begin{center}
\resizebox{0.5\linewidth}{!}{
\begin{tabular}{ccc}
\hline
           & Yolov5   & Yolov3    \\ \hline 
visible    & 22.59\%   & 37.70\%  \\ 
infrared   & 10.66\%   & 19.73\%  \\ \hline
\end{tabular}}
\end{center}
\caption{Log average miss rate of Yolov5 and Yolov3 on the LLVIP dataset. } 
\label{table-missrate}
\end{table}

\begin{figure}[htb]
\begin{center}
\includegraphics[height=0.8\linewidth]{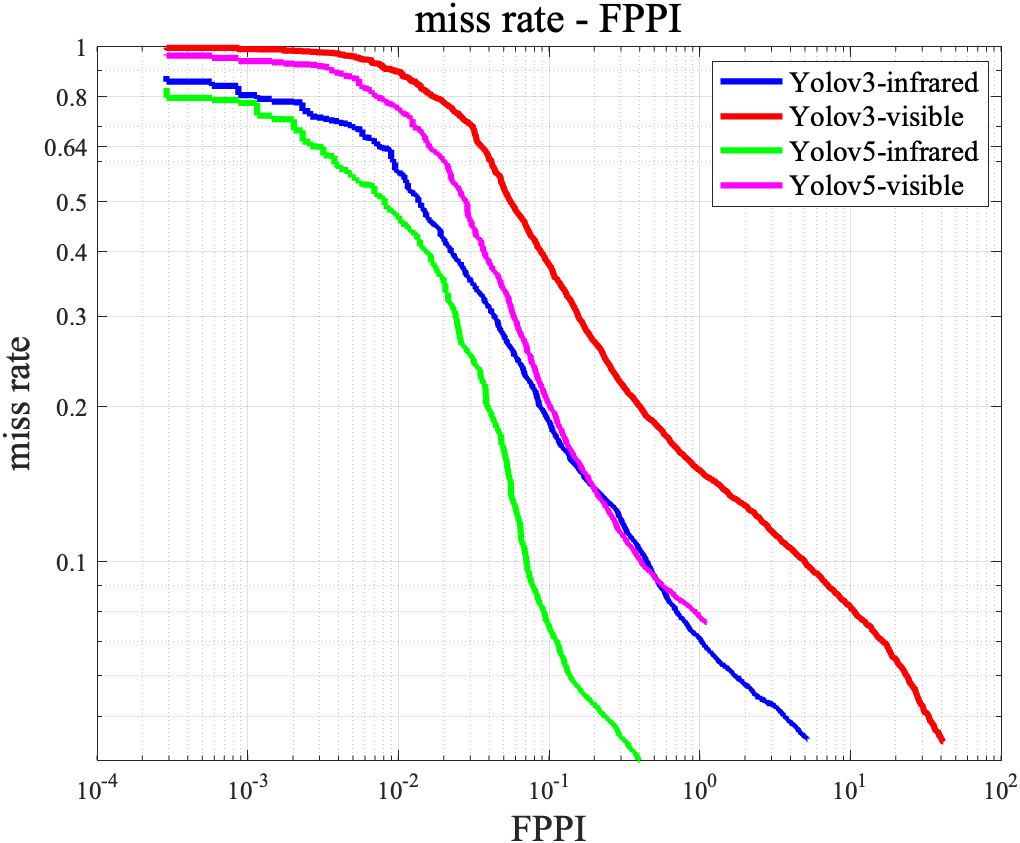}
\end{center}
\caption{Miss rate-FPPI curve on the LLVIP dataset.} 
\label{fig-mr_fppi} 
\end{figure}

\begin{figure}[htb]
\begin{center}
\includegraphics[height=0.18\linewidth]{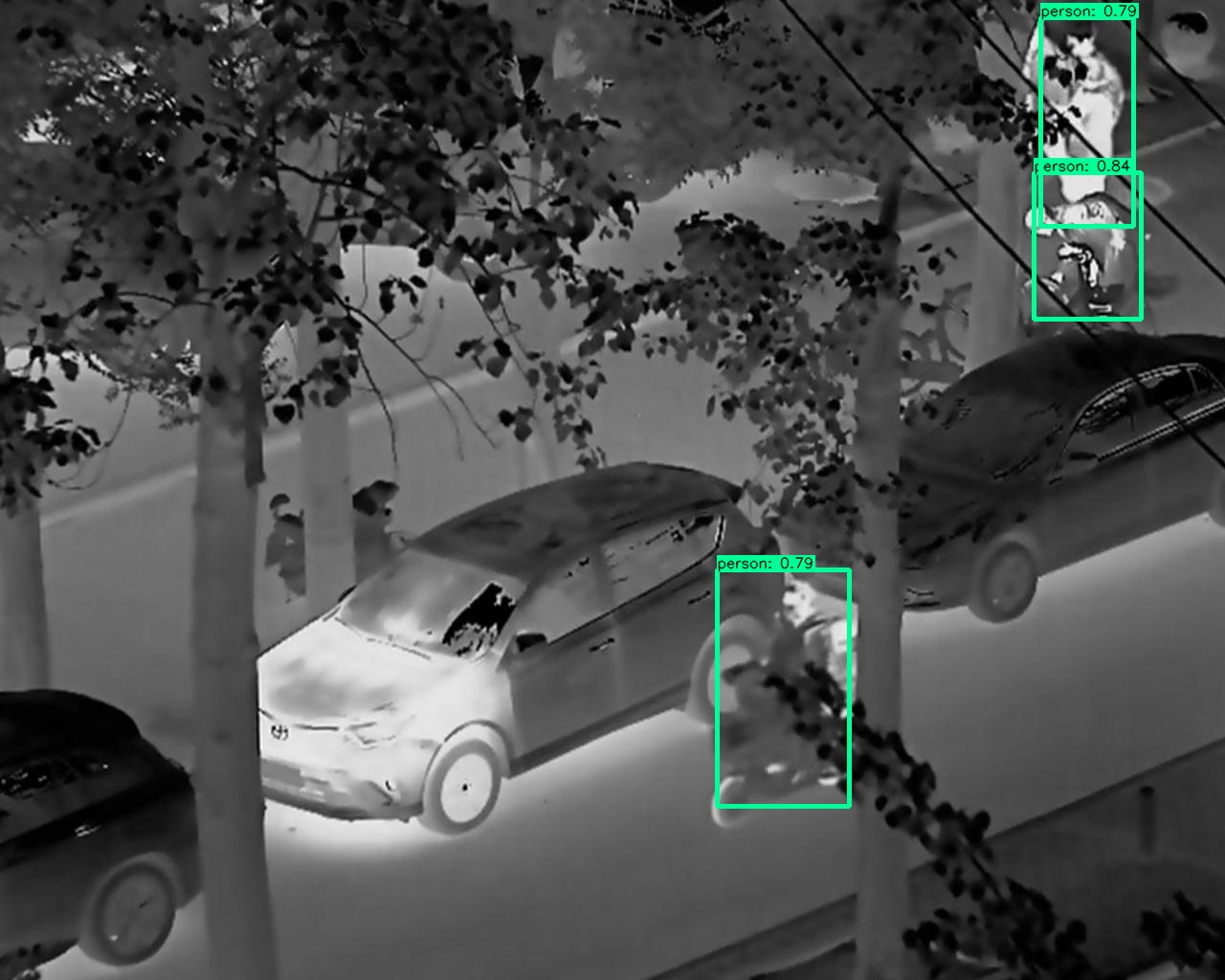}
\includegraphics[height=0.18\linewidth]{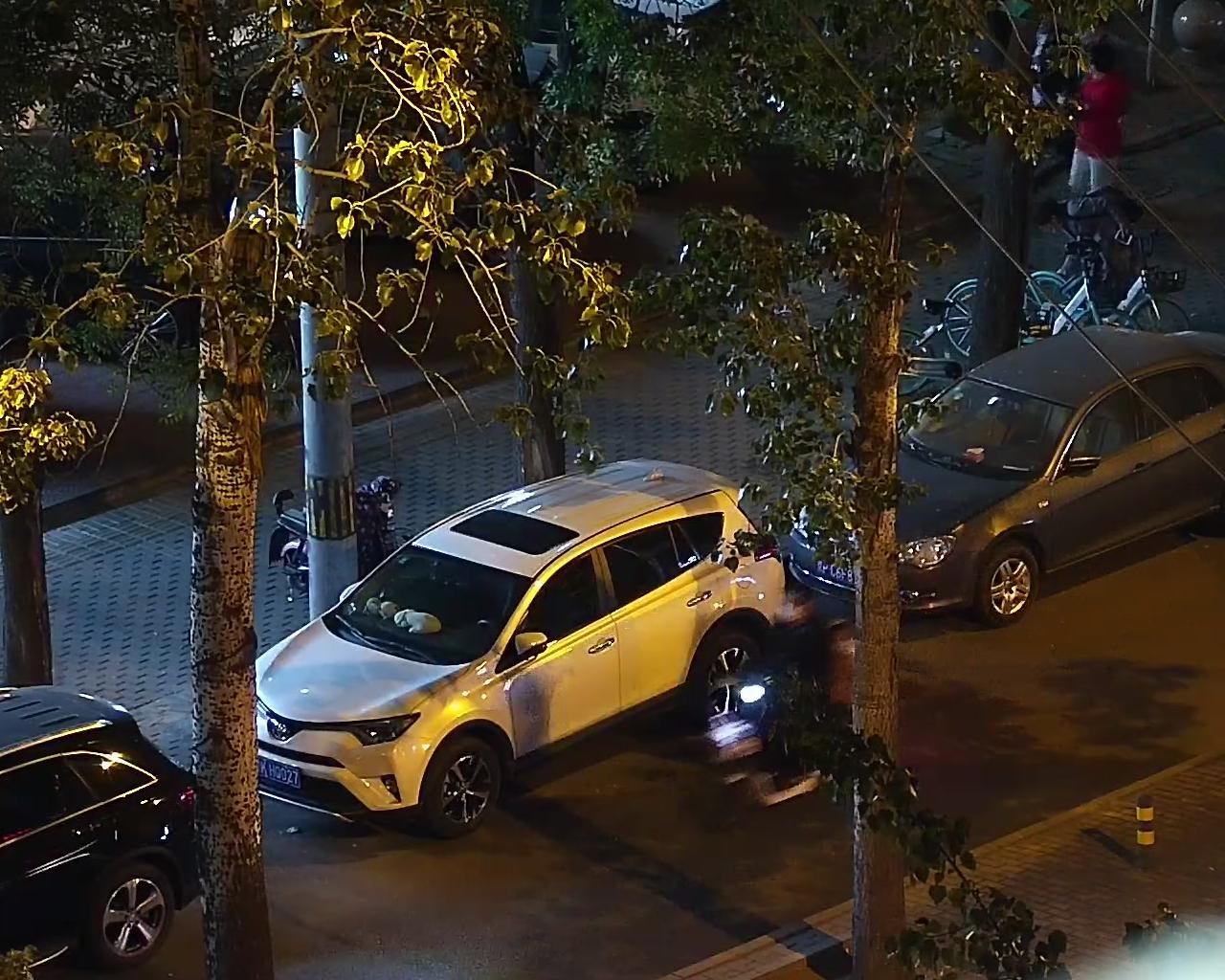}
\hspace{4pt}\vspace{4pt}
\includegraphics[height=0.18\linewidth]{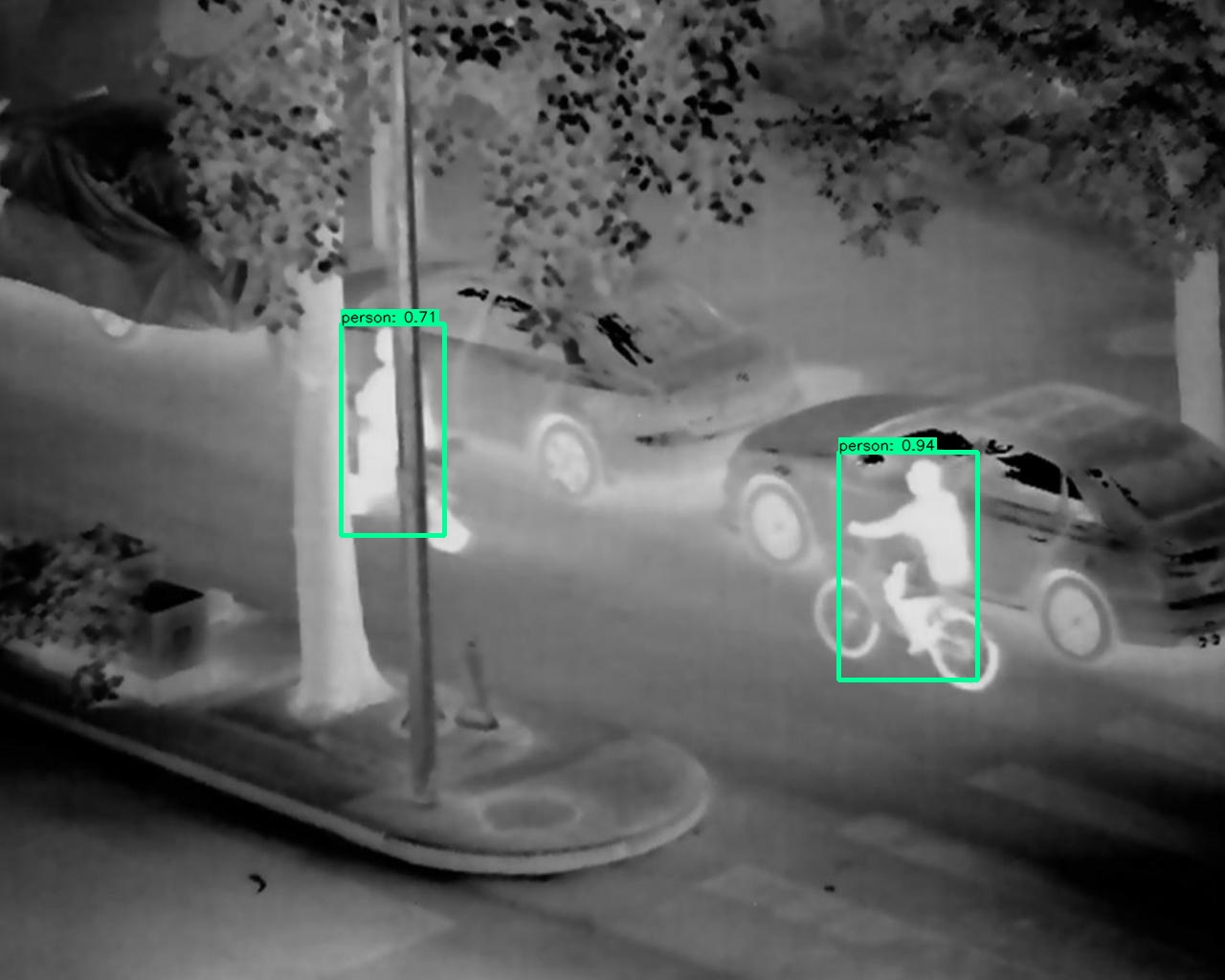}
\includegraphics[height=0.18\linewidth]{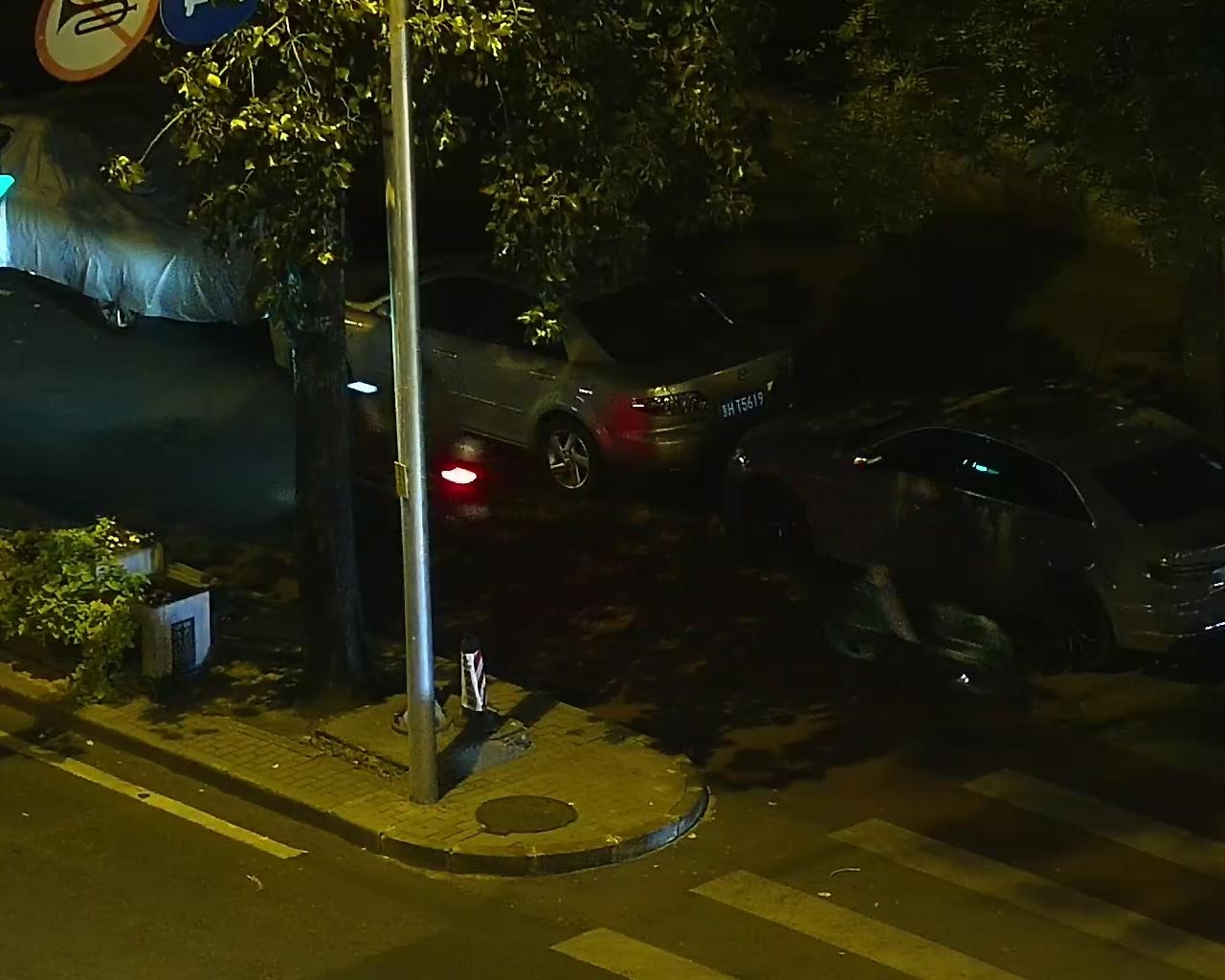}

\includegraphics[height=0.18\linewidth]{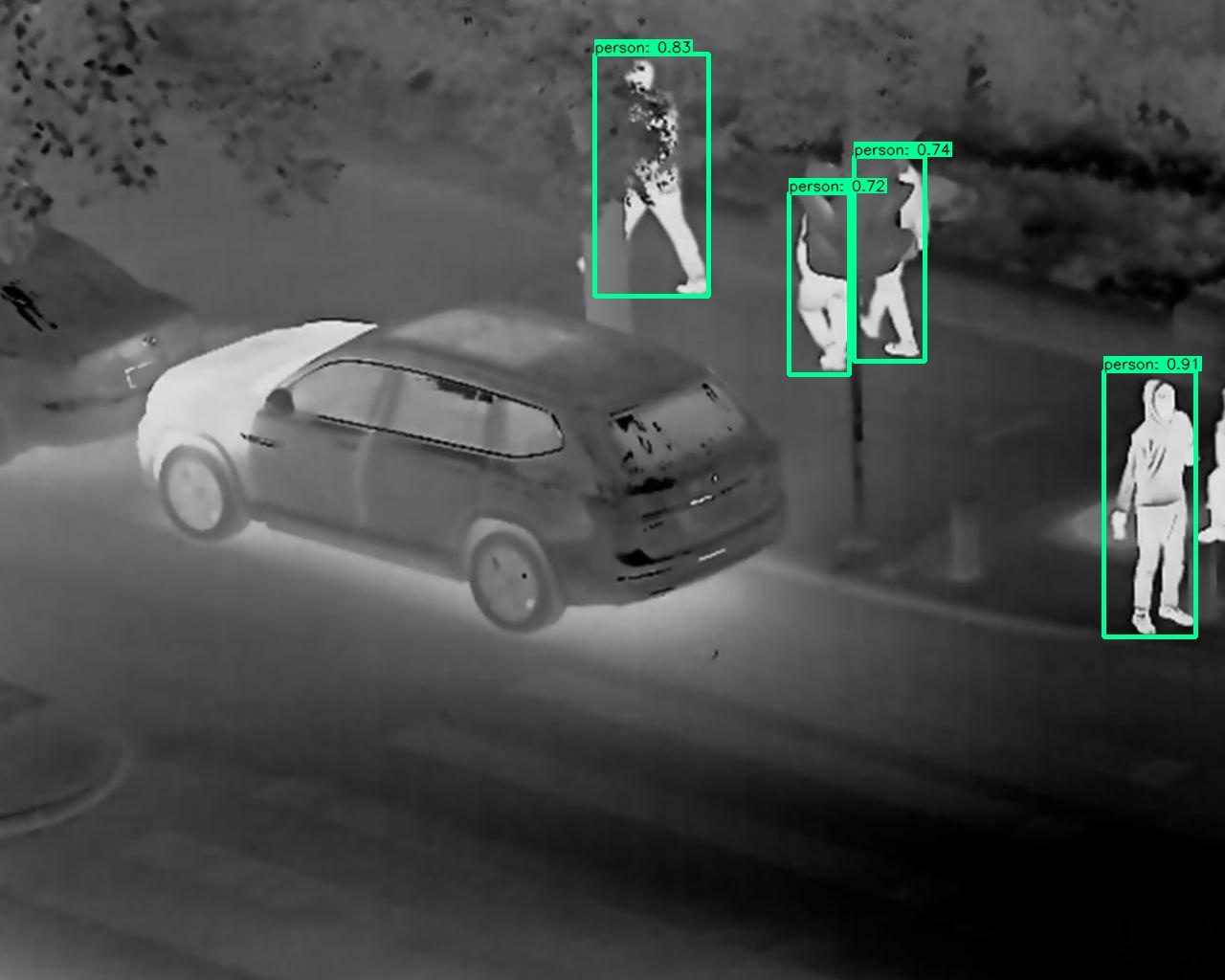}
\includegraphics[height=0.18\linewidth]{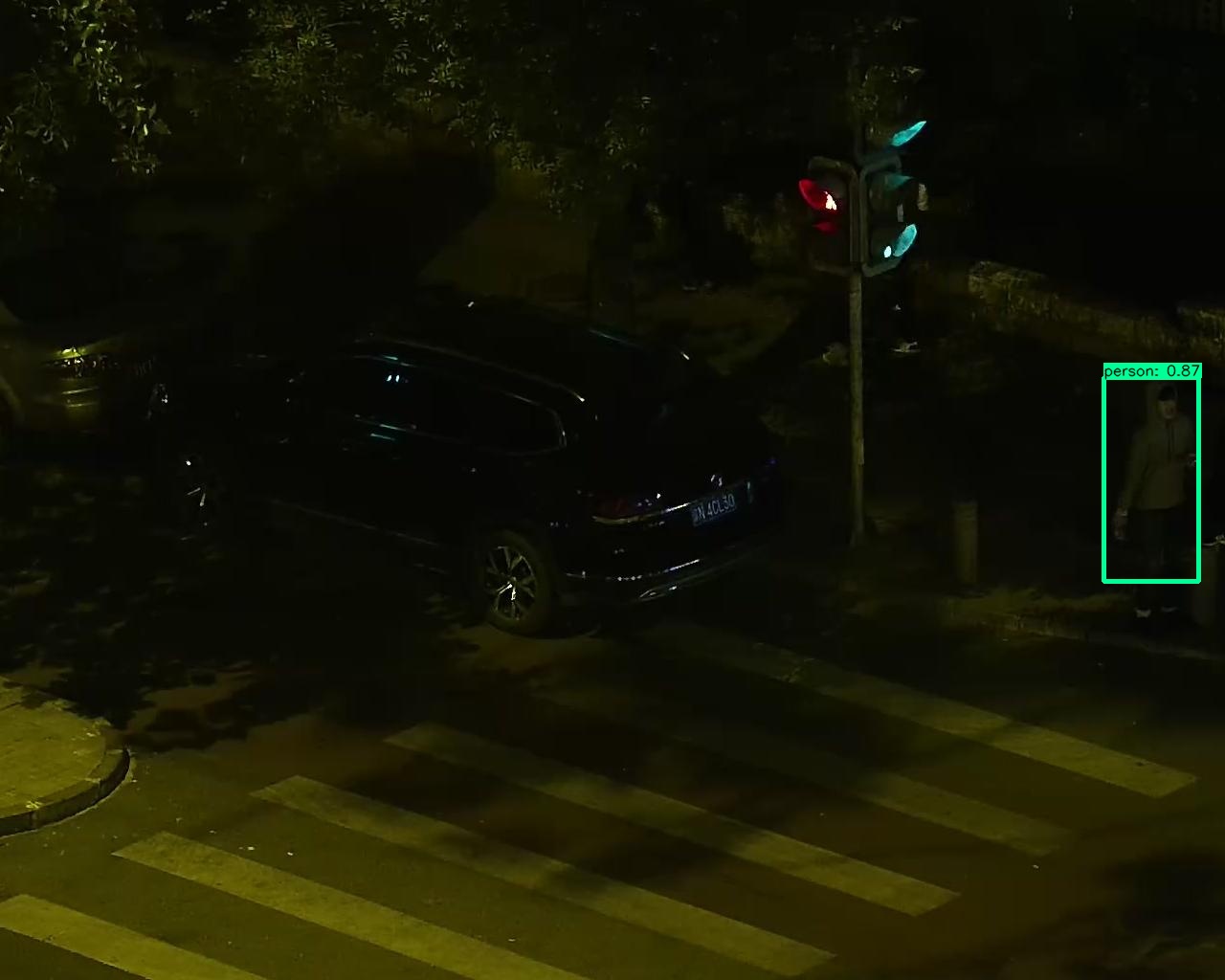}
\hspace{4pt}\vspace{4pt}
\includegraphics[height=0.18\linewidth]{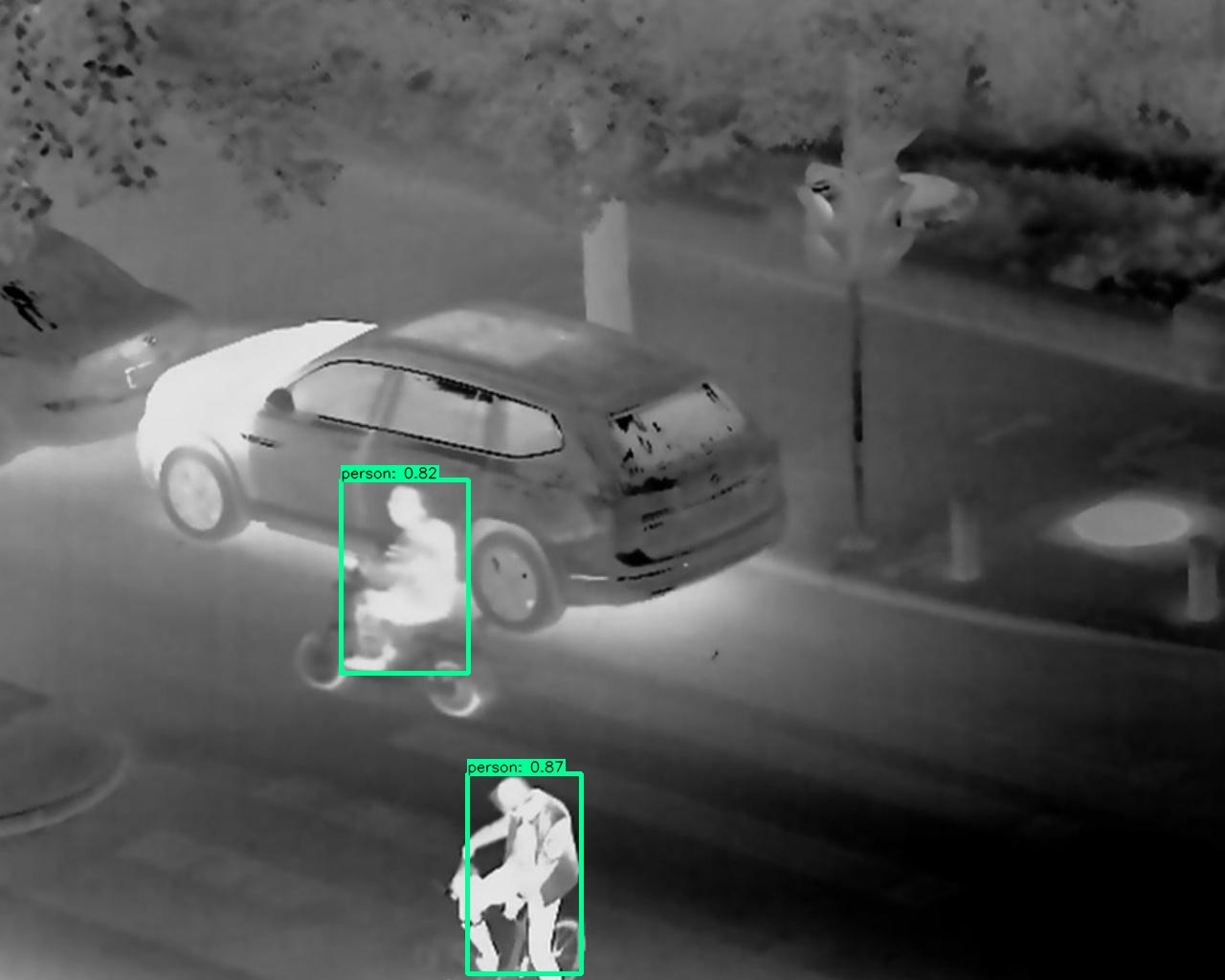}
\includegraphics[height=0.18\linewidth]{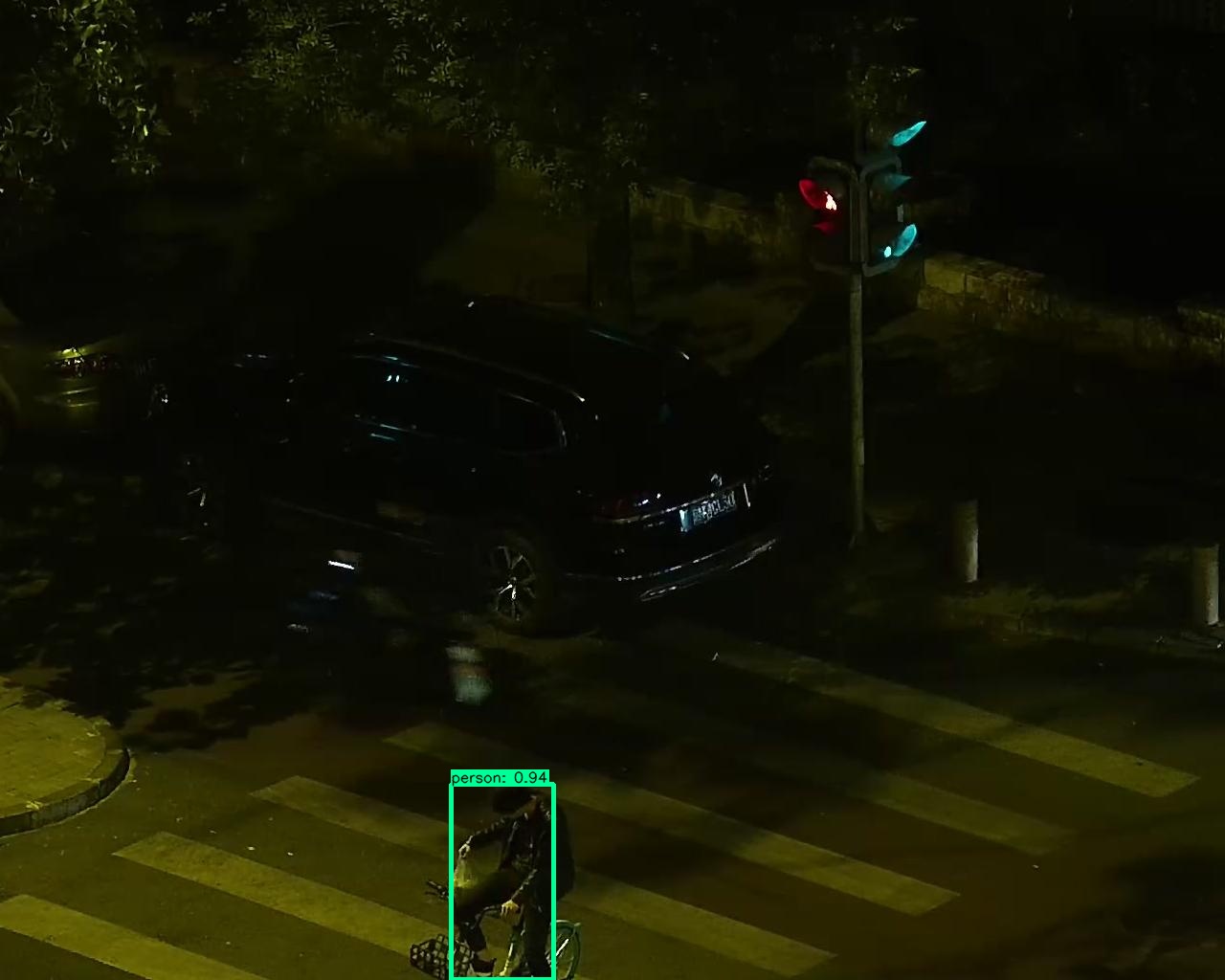}
\end{center}
\caption{Examples of the results of detection experiments.} 
\label{fig-detection} 
\end{figure}

After training and testing, experiment results on visible images and infrared images are shown in Table~\ref{table-mAP}, Table~\ref{table-missrate} and Figure~\ref{fig-mr_fppi}. Examples of the results of the experiments are shown in Fig.~\ref{fig-detection}. There are many missed detection phenomena in visible images. The infrared image highlights pedestrians, and achieves a better effect in the detection task, which not only proves the necessity of infrared images but also indicates that the performance of pedestrian detection algorithm is not good enough under low-light conditions. There is at least some discrepancy between the results of visible and infrared images. This dataset can then be used to study and improve the performance of pedestrian detection algorithms at night.

\subsection{Image-to-image Translation}

For image-to-image translation, pix2pixGAN~\cite{isola2017image} is used to experiment. The structure of generator is unet256, and the structure of discriminator is the basic PatchGAN as default. We first resize images to 320$\times$256, and then crop them to 256$\times$256 in the data preprocessing stage. The batch size is set to 8 with the same GPU mentioned before. We train the model in 100 epochs with the initial learning rate 0.0002 and then in 100 epochs to linearly decay learning rate to zero. 

\begin{figure}[htb]
  \begin{center}
  \begin{minipage}[b]{0.96\linewidth}
  \subfigure[visible]{
    \begin{minipage}[b]{0.28\linewidth}
      \centering
      \includegraphics[width=\linewidth]{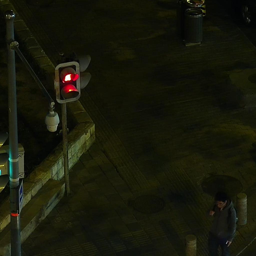}\vspace{4pt}
      \includegraphics[width=\linewidth]{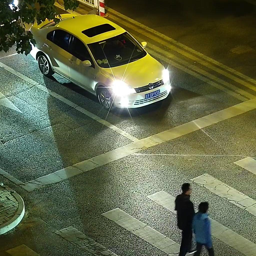}\vspace{4pt}
    \end{minipage}
  }
  \subfigure[generated]{
    \begin{minipage}[b]{0.28\linewidth}  
      \centering
      \includegraphics[width=\linewidth]{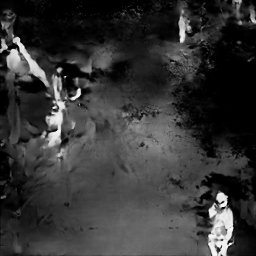}\vspace{4pt}
      \includegraphics[width=\linewidth]{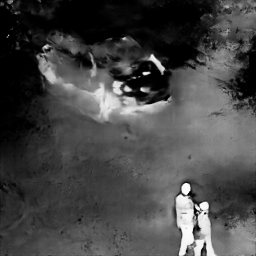}\vspace{4pt}
    \end{minipage}
  }
  \subfigure[GT]{
    \begin{minipage}[b]{0.28\linewidth}  
      \centering
      \includegraphics[width=\linewidth]{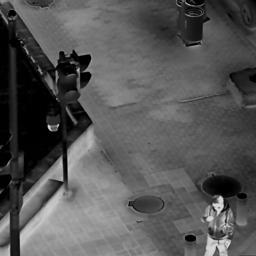}\vspace{4pt}
      \includegraphics[width=\linewidth]{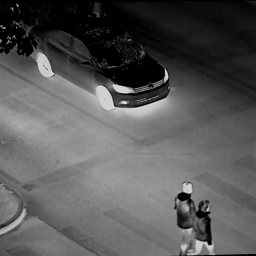}\vspace{4pt}
    \end{minipage}
  }
  \end{minipage}
  \vfill
  \end{center}
  \caption{Examples of image-to-image translation results of pix2pixGAN on the LLVIP dataset. From left to right: (a) the original visible images, (b) the generated infrared images, (c) the ground truth infrared images.}
\label{fig-translation} 
\end{figure}

\begin{table}[htb]
\begin{center}
\resizebox{0.5\linewidth}{!}{
\begin{tabular}{ccc}
\hline
Dataset  & SSIM   & PSNR    \\ \hline 
KAIST    & 0.6918 & 28.9935 \\ 
LLVIP    & 0.1757 & 10.7688  \\ \hline
\end{tabular}}
\end{center}
\caption{Experiment results of pix2pixGAN on KAIST dataset and our LLVIP dataset. } 
\label{table-pix2pix}
\end{table}

The popular pix2pixGAN has shown very poor performance on our LLVIP. Qualitatively, we show two examples of image-to-image translation results in Fig.~\ref{fig-translation}. It can be seen that both the quality of the generated image and the similarity to the real image are not satisfactory. Specifically, the background in the generated image is messy, the contours of pedestrian and the car is not clear and the details are wrong, and there are many artifacts on the image.

Quantitatively, it shows extremely low SSIM and PSNR as shown in Table~\ref{table-pix2pix}. We compare the experimental results of pix2pixGAN presented by Qian \etal~\cite{qian2020sparse} on the KAIST multi-spectral pedestrian dataset. Obviously, the performance of the image-to-image translation algorithm on LLVIP is much worse than on KAIST. The reasons for this gap are probably: 1) The pix2pixGAN  has poor generalization ability. The scenarios of KAIST dataset have little change, while the scenarios of LLVIP training set and test set are different. 2) The performance of pix2pixGAN decreases significantly in low light conditions. The lighting conditions of dark night images in KAIST are still good, unlike the images in LLVIP. Therefore, there is still a lot of room for improvement in image-to-image translation algorithms under low light conditions, and a visible-infrared paired dataset for low-light vision is desperately needed.

\section{Conclusion}

In this paper, we present LLVIP, a visible-infrared paired dataset for low-light vision. The dataset is strictly aligned in time and space, containing a large number of pedestrians, containing a large number of images with low-light conditions, containing annotations for pedestrian detection. Experiments on the dataset indicate that the performance of visible and infrared image fusion, low-light pedestrian detection and image-to-image translation all need to be improved.

We provide LLVIP dataset for use in, but not limited to, the following studies: 1) Visible and infrared image fusion. Images are aligned in the dataset. 2) Low-light pedestrian detection. Low-light visible images are accurately labeled. 3) Image-to-image translation. 4) Others, such as multi-modal image registration and domain adaptation.

\section*{Acknowledgments}

This work was supported in part by 111 Project of China (B17007), and in part by the National Natural Science Foundation of China (61602011).


{\small
\bibliographystyle{ieee_fullname}
\bibliography{egpaper_arXiv}
}

\end{document}